\documentclass[10pt]{article} 
\usepackage[accepted]{tmlr}

\usepackage{amsmath,amsfonts,bm}

\def\eqref#1{equation~\ref{#1}}

\def\1{\bm{1}}

\DeclareMathAlphabet{\mathsfit}{\encodingdefault}{\sfdefault}{m}{sl}
\SetMathAlphabet{\mathsfit}{bold}{\encodingdefault}{\sfdefault}{bx}{n}

\usepackage[utf8]{inputenc} 
\usepackage{url}            
\usepackage{booktabs}       
\usepackage{amsfonts}       
\usepackage{nicefrac}       
\usepackage{microtype}      
\usepackage{diagbox}        
\usepackage{makecell}
\usepackage{xspace}
\usepackage{url}
\usepackage{graphicx}
\usepackage{bm}
\usepackage{amssymb}
\usepackage{amsmath}
\usepackage{amsthm}
\usepackage{algorithm}
\usepackage{algpseudocode}
\usepackage[table,xcdraw,dvipsnames]{xcolor}
\usepackage[title]{appendix}
\usepackage{wrapfig}
\usepackage{hyperref}
\definecolor{customblue}{rgb}{0.36, 0.55, 0.75}
\definecolor{myred}{RGB}{220,0,0}
\definecolor{myblue}{RGB}{0,80,255}
\definecolor{mygreen}{RGB}{0,150,0}
\usepackage{comment}
\usepackage{bbm}
\usepackage{multirow}
\usepackage{float}
\usepackage{subcaption}
\usepackage{makecell}

\newcommand{\cut}[1]{}
\newcommand{\std}[1]{\scriptsize{$\pm$#1}}

\newcommand{\myname}{LEAD\xspace}

\usepackage{listings}
\usepackage{graphicx}
\usepackage{amsmath}
\usepackage{setspace}
\usepackage{pifont}
\usepackage{tikz}

\title{
    LEAD: An EEG Foundation Model for Alzheimer’s Disease Detection%
}

\author{\name Yihe Wang \email ywang145@charlotte.edu \\
      University of North Carolina at Charlotte
      \AND
      \name Nan Huang \email nhuang1@charlotte.edu \\
      University of North Carolina at Charlotte
      \AND
      \name Nadia Mammone \email nadia.mammone@unirc.it\\
      University of Reggio Calabria
      \AND
      \name Marco Cecchi \email mcecchi@cognision.com\\
      Cognision
      \AND
      \name Xiang Zhang \email xiang.zhang@charlotte.edu\\
      University of North Carolina at Charlotte
}

\def\month{09}  
\def\year{2026} 
\def\openreview{\url{https://openreview.net/pdf?id=AigNTyxcvH}} 

\begin{document}

\maketitle

\begin{abstract}

Electroencephalography (EEG) provides a non-invasive, highly accessible, and cost-effective approach for detecting Alzheimer’s disease (AD). However, existing methods, whether based on handcrafted feature engineering or standard deep learning, face three major challenges: 1) the lack of large-scale EEG-based AD datasets for robust representation learning and evaluation; 2) limited cross-subject generalizability; and 3) difficulty in adapting to highly heterogeneous data. To address these challenges, we curate the world’s largest EEG-AD corpus to date, comprising 2,238 subjects. Leveraging this unique resource, we propose \myname, the first foundation model for EEG-based AD detection. Specifically, we design a gated temporal-spatial Transformer that can adapt to EEG recordings with diverse lengths, channel configurations, and sampling rates. In addition, we introduce a subject-regularized training strategy to enhance end-to-end subject-level detection. We further employ medical contrastive learning to pre-train on 13 datasets, including 4 AD datasets and 9 non-AD neurological disorder datasets, and fine-tune/test the model on the other 5 AD datasets. \myname achieves the best average ranking across all 20 evaluations on 5 downstream datasets, substantially outperforming existing approaches, including state-of-the-art (SOTA) EEG foundation models. These results strongly demonstrate the effectiveness of our proposed method and significant progress for EEG-based AD detection. Source code: \url{https://github.com/DL4mHealth/LEAD}

\end{abstract}

\section{Introduction}
\label{sec:intro}

\begin{table}[h]
    \centering
    \scriptsize
    \vspace{-2mm}
    \caption{\textbf{Comparison of Different AD Detection Modalities.} Different modalities present distinct trade-offs with respect to accessibility, safety, affordability, and detection accuracy. 
    }
    \vspace{-2mm}
    \label{tab:ad_method_comparison}
    \resizebox{\columnwidth}{!}{
    \begin{tabular}{lccccc}
    \toprule
    \textbf{Technologies} & \textbf{Accessibility$\uparrow$} & \textbf{Safety$\uparrow$} & \textbf{Affordability$\uparrow$} & \textbf{Accuracy$\uparrow$} \\
    \midrule
    \textbf{Cerebrospinal Fluid~~\citep{salvado2024disease}} & Low & Low & Low & High \\
    \textbf{Neuroimaging~~\citep{ottoy2025recent}} & High & High & Low & High \\
    \textbf{Blood Test~~\citep{grande2025blood}} & Low & Moderate & Moderate & Moderate \\
    \textbf{Neurological Exam~~\citep{voglein2023pattern}} & High & High & Moderate & Moderate \\
    \rowcolor[HTML]{D9EAF7}
    \textbf{EEG~~\citep{wang2024adformer}} & High & High & High & Low \\
    \bottomrule
    \end{tabular}
    }
    \vspace{-5mm}
\end{table}

Alzheimer's Disease is the most common neurodegenerative disorder in the elderly, affecting
38 million individuals with \$1.3 trillion annual financial cost~~\citep{masters2015alzheimer,breijyeh2020comprehensive,national2021reducing}. 
Early detection of AD is a significant step to slow symptom progression and increase patients' life expectancy~\citep{chu2012alzheimer,nelson2015slowing}. Compared with other AD detection modalities, EEG offers appealingly unique advantages (Table~\ref{tab:ad_method_comparison}). These include high accessibility through portable devices, strong safety due to its non-invasive nature, and high affordability with low hardware and operational costs~\citep{ieracitano2019convolutional}. In addition, EEG enables long-term monitoring of disease progression with high temporal resolution and real-time capability~\citep{ahmed2025stick}. Existing EEG-based AD detection methods generally fall into two main research directions. The first relies on handcrafted biomarkers, such as phase shift~\citep{wang2017enhanced}, power spectral density~\citep{fahimi2017index}, and Shannon entropy~\citep{azami2019multiscale}. The second direction leverages deep learning models, including convolutional neural networks~\citep{roncero2024inter}, graph neural networks~\citep{klepl2023adaptive}, and Transformer-based architectures~\citep{miltiadous2023dice,wang2024adformer}. 

However, existing EEG-based AD detection methods, whether using manual feature extraction or deep learning, still suffer from several limitations. \textbf{First,} large and high-quality datasets remain scarce. The cost and complexity of collecting EEG data from patients with AD result in most studies being conducted on relatively small cohorts, typically ranging from 20 to 100 subjects~\citep{aviles2024machine}. The limited diversity and scale of existing EEG datasets pose a fundamental challenge to learn robust models. \textbf{Second,} model generalizability across subjects remains a critical bottleneck. The core challenge for real-world application is generalization to unseen subjects (subject-independent evaluation). Current approaches, while often strong in controlled, at-lab settings (subject-dependent), suffer a substantial performance drop when moving to subject-independent evaluation~\citep{wang2024medformer}. This degradation is largely caused by inter-subject variability and shortcut learning between diagnostic labels and subject-specific characteristics~\citep{wang2024causes}. Although subjects diagnosed with AD are expected to exhibit consistent disease-related patterns, confounding factors such as age, gender, and other personal attributes may obscure these patterns. \textbf{Third,} existing studies are difficult to adapt to heterogeneous data. Most prior works train models on a single dataset or on datasets with identical acquisition configurations~\citep{nayana2025eeg}. In practice, EEG data are highly heterogeneous, not only due to demographic differences among subjects but also because of variations in recording environments, acquisition devices, channel montages, sampling rates, and recording durations. This heterogeneity makes it challenging to transfer the detection capability of existing methods across datasets, which is critical for real-world applicability.

\begin{figure*}
    \centering
    \includegraphics[width=1.0\linewidth]{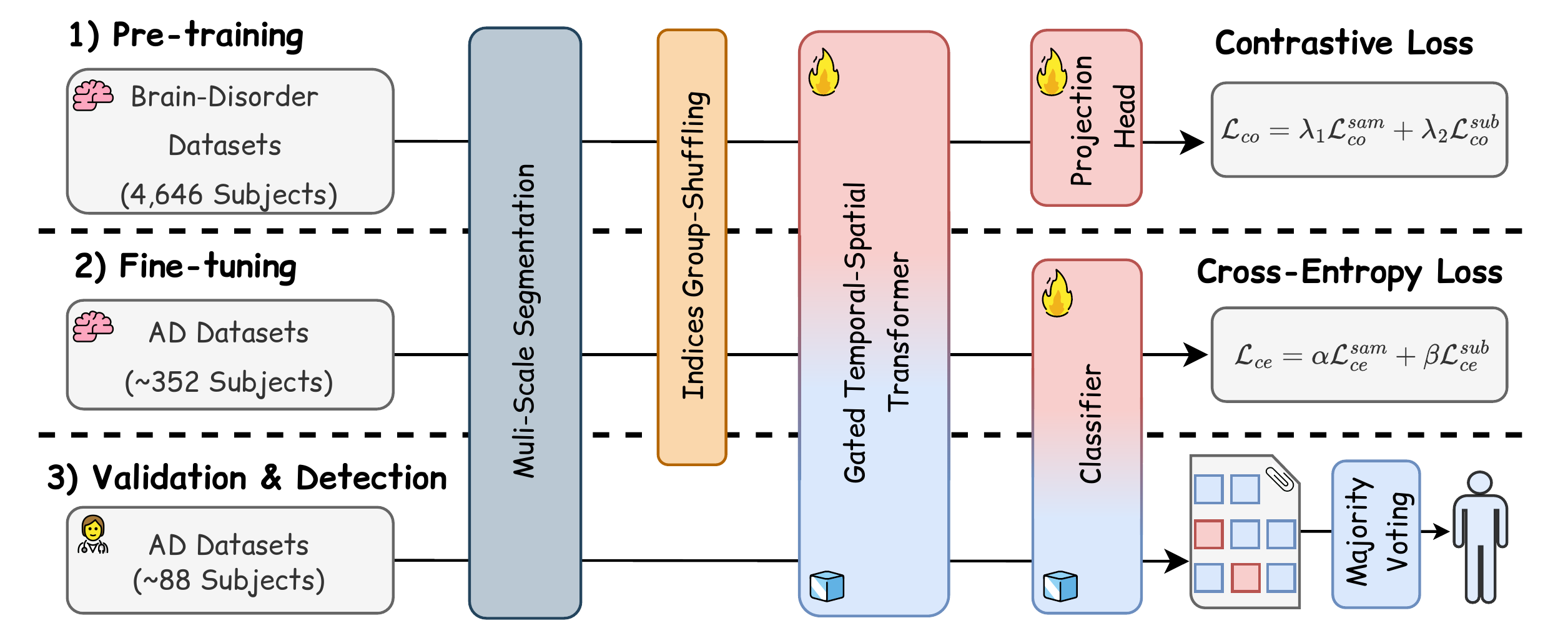}
    \caption{\textbf{LEAD overview.} We pre-train \myname on 13 neurological disorder datasets using medical contrastive learning, and subsequently fine-tune it on 5 AD datasets under a \textbf{subject-independent cross-validation} protocol, with training, validation, and test splits of 80\%, 10\%, and 10\%, respectively. A gated temporal-spatial Transformer is employed to jointly capture temporal and spatial features. In addition, we propose a subject-level cross-entropy loss $\mathcal{L}_{\mathrm{ce}}^{\mathrm{sub}}$ to enhance subject-level representation learning. Engineering tricks, including multi-scale segmentation and index group shuffling, are further introduced to facilitate subject-level training. Finally, majority voting is applied to obtain subject-level detection results.
    }
    \label{fig:lead_pipeline}
    \vspace{-5mm}
\end{figure*}

To address the aforementioned limitations, we propose \textbf{\myname}, the first AD detection EEG foundation model. We conduct a comprehensive review of relevant publications and public EEG repositories (by 2026), including OpenNeuro, Dryad, and Figshare, and identify 9 highly heterogeneous EEG datasets for AD and dementia detection. These datasets comprise \textbf{2,238 subjects and 427.81 hours} of recordings. To the best of our knowledge, this constitutes the world’s \textbf{largest} EEG dataset for AD detection, which is approximately \textbf{20 times larger} than those commonly used in prior publications. Building on this unique resource, we train an EEG foundation model designed for \textbf{subject-level} Alzheimer’s detection that can handle EEG signals of \textbf{diverse lengths, channel montages, and commonly used sampling rates.} To this end, we design a gated temporal-spatial Transformer as the backbone. Univariate patch embedding is used to adapt to downstream tasks with varying input lengths and channel counts. In addition, 3D channel embeddings and sampling rate embeddings allow the model to capture spatial relationships and to distinguish EEG recorded at different sampling rates. A parallel temporal-spatial self-attention mechanism is employed to extract both temporal and spatial features, with a learnable gated fusion module to adaptively control their relative importance. Furthermore, we introduce a subject-regularized training strategy. A subject-level cross-entropy loss is proposed to propagate subject-level errors back to the model during training. Useful engineering tricks, such as multi-scale segmentation and index group shuffling, are used to further enhance subject-level learning. Finally, we employ a domain-inspired self-supervised medical contrastive learning strategy for pre-training on 4 of the 9 AD datasets, along with 9 additional neurological disorder datasets, comprising a total of 4,646 subjects and 1,185.84 hours of EEG recordings. The remaining 5 AD datasets are used for fine-tuning and evaluation under a realistic yet challenging subject-independent cross-validation protocol. Figure~\ref{fig:lead_pipeline} illustrates the overall pipeline of our proposed method. We compare our method against 16 baseline methods, including manual feature extraction, supervised deep learning methods, and SOTA EEG foundation models. \myname achieves the best average ranking across all 20 evaluations on 5 datasets, far surpassing the second-best model LaBraM while using significantly fewer pre-training resources.

The key contributions of this work are summarized as follows: 1) The largest EEG-based AD detection dataset. We curate the largest dataset for EEG-based AD detection to date, far surpassing prior work. 2) First EEG Foundation Model for AD detection. Built on this unique dataset, we develop the first foundation model for end-to-end subject-level EEG-based AD detection, adaptable to diverse recording lengths, channel topologies, and sampling rates. 3) Superior performance. \myname achieves SOTA subject-level results on 5 downstream AD datasets under challenging subject-independent cross-validation.

\section{Related Work}
\label{sec:related}

\subsection{EEG-Based AD Detection}
\label{sub:eeg_ad_detection}

In the last two decades, EEG-based AD detection has followed two main research directions: manual biomarker extraction and deep learning methods. 

\textbf{Manual Biomarker Extraction:} This research direction aims to identify potential biomarkers in EEG signals of AD patients and use simple classifiers, such as Multi-Layer Perceptrons (MLP) and Support Vector Machines (SVM), to differentiate these features from normal healthy subjects. Different types of EEG features are used, including statistical features like Mean, Skewness, Kurtosis, and Standard Deviation~\citep{tzimourta2019eeg, tzimourta2019analysis, kulkarni2017extracting, kanda2014clinician, waser2013eeg, tylova2013predictive, mora2019scale}, spectral features like Phase Shift, Phase Coherence, Bispectrum, and Bicoherence~\citep{wang2017enhanced, cassani2014effects, wang2015multiple, fraga2013characterizing, tait2019network, waser2016quantifying, trambaiolli2011improving}, power features like Power Spectrum Density, Relative Band Power, Ratio of EEG Rhythm, and Energy~\citep{fahimi2017index, schmidt2013index, liu2016multiple, kanda2014clinician}, as well as complexity features like Shannon Entropy, Tsallis Entropy, and Permutation Entropy~\citep{garn2015quantitative, azami2019multiscale, tylova2018unbiased, coronel2017quantitative, al2018complexity}. The main advantage of this approach is its interpretability, but it suffers from limited performance.

\textbf{Deep Learning:} Compared to manual biomarker extraction, deep learning offers an alternative approach by automatically extracting useful representations for AD detection. Models such as Convolutional Neural Networks (CNNs)~\citep{li2022predictive, cura2022deep}, Graph Neural Networks (GNNs)~\citep{shan2022spatial, klepl2023adaptive}, and Transformers~\citep{wang2024adformer} are widely used for representation learning. Some researchers still perform manual feature extraction or transform the data before applying deep learning models. For example, the method in~\citep{ieracitano2019convolutional} converts 5-second EEG intervals into Power Spectral Density (PSD) images and uses 2D convolutional layers on the images for feature extraction. DICE-net~\citep{miltiadous2023dice} extracts relative band power and spectral coherence connectivity across five frequency bands and applies convolutional layers followed by transformers. In contrast, some studies apply deep learning methods directly to EEG data. For instance, the method in~\citep{gallego2024alzheimer} uses semi-supervised spatiotemporal representation learning with deep learning models for AD detection based on different sleep-stage EEG data. STEADYNet~\citep{kachare2024steadynet} designs low-complexity convolutional models for AD and dementia detection, focusing on fast inference times. Research in ~\citep{watanabe2024deep} using MNet that applies convolutional networks for feature extraction and concatenates with the relative power spectrum for AD and other dementia detection. ADformer~\citep{wang2024adformer} uses a multi-granularity transformer for AD detection and widely tests on 4 AD datasets. Research in \citep{nayana2025eeg} compares biomarker extraction and deep learning methods for the detection of neurodegenerative diseases like AD.

\subsection{Self-Supervised and Foundation Model in EEG}
\label{sub:self_supervised_pretrain}

Two main strategies are widely used for self-supervised representation learning and foundation model training in EEG: contrastive learning and masked autoencoding. 

\textbf{Contrastive Learning:} BENDR~\citep{kostas2021bendr} follows a similar contrastive learning pipeline as Wav2Vec~\citep{baevski2020wav2vec}, but it is trained on EEG data. EEG2Vec~\citep{zhu2023eeg2vec} explores both contrastive learning and mask-reconstruction for self-supervised pre-training on EEG data. BIOT~\citep{yang2024biot} designs a transformer architecture for biomedical signal embedding and applies a self-supervised contrastive framework similar to BYOL~\citep{grill2020bootstrap}. COMET~\citep{wang2023contrast} utilizes various data levels in biomedical time series to define positive and negative pairs. 

\textbf{Masked Autoencoding:} Neuro-BERT~\citep{wu2024neuro} employs masked autoencoding to predict missing amplitude and phase of EEG signals during pre-training. EEG2Rep~\citep{mohammadi2024eeg2rep} combines a context encoder with a momentum target encoder to reconstruct context-level representations rather than raw data in self-supervised pre-training. LaBraM~\citep{jiang2024large}, the first large foundation model in the EEG domain, uses a neural tokenizer to reconstruct the Fourier spectrum during self-supervised pre-training. EEGPT~\citep{wangeegpt} is a foundation model for EEG representation learning that integrates reconstruction loss with an alignment loss between the encoder and momentum encoder. CBraMod~\citep{wang2024cbramod} designs a criss-cross transformer to leverage both spatial and temporal features of EEG. LUNA~\citep{donerluna} introduces a cross-attention mechanism for a topology-agnostic foundation model capable of handling heterogeneous channels. CSBrain~\citep{zhou2025csbrain} utilizes regions of the brain to design cross-window and cross-region dependencies for diverse decoding EEG tasks. NeurIPT~\citep{fang2025neuript} leverages a progressive mixture-of-experts architecture to build backbones and introduces 3D electrode embedding for EEG channel embedding. REVE~\citep{ouahidi2025reve} pretrains on over 60,000 hours of EEG data from 92 datasets spanning 25,000 subjects, demonstrating the largest pre-training resources so far. ST-EEGFormer~\citep{yang2026eeg} uses a simple Vision Transformer-based structure with raw masked autoencoder training to achieve SOTA performance on EEG tasks.

\textbf{Other Strategies.} Recent work has begun exploring autoregressive pre-training for EEG, such as NeuroGPT~\citep{cui2024neuro} and a study also named EEGPT~\citep{yue2024eegpt}. Research bridging language and EEG has also emerged, for instance, NeuroLM~\citep{jiang2024neurolm}, which treats EEG as a foreign language for representation learning. Laya applies the JEPA~\citep{assran2023self} approach on EEG representation learning~\citep{panchavati2026laya}. Research~\citep{tang2026eeg} on interpreting EEG foundation models is also emerging to explore what exact features EEG foundation models learn.

\textbf{Application-Oriented Approaches.} Several studies focus on application-specific foundation models. Brant series~\citep{zhang2023brant,zhang2024brant} train models tailored to intracranial neural signals; PPi~\citep{yuan2023ppi} develops a self-supervised framework for subject-independent seizure detection; MIRepNet~\citep{liu2025mirepnet} designs a foundation model for motor imagery classification. FAPEX~\citep{zheng2025fapex} utilizes the fractional amplitude-phase expressor for cross-subject seizure prediction. CogMoE~\citep{shah2026cogmoe} uses a signal quality-aware mixture-of-experts to train a cross-modal foundation model with EEG, ECG, EDA, and Gaze modalities together.

\section{Datasets}
\label{sec:preliminary}

\begin{table*}[h]
    \centering
    \caption{\textbf{Dataset Statistics.} The \textbf{h} and \textbf{m} in the total time denote hours and minutes. Abbreviations: \textbf{ASSR}: Auditory Steady-State Response; \textbf{RS}: Resting State; \textbf{HV}: Hyperventilation; \textbf{PS}: Photic Stimulation. \textbf{HC} - Healthy Controls; \textbf{AD}: Alzheimer’s Disease; \textbf{FTD}: Frontotemporal Dementia; \textbf{PD}: Parkinson’s disease; \textbf{MS}: Multiple Sclerosis; \textbf{MCI}: Mild Cognitive Impairment; \textbf{SCZ}: Schizophrenia; \textbf{DEP}: Depression; \textbf{DEM}: Dementia; \textbf{FEP}: First Episode Psychosis; \textbf{MDD}: Major Depressive Disorder; \textbf{ADHD}: Attention Deficit Hyperactivity Disorder; \textbf{SMC}: Subjective Memory Complaints; \textbf{OCD}: Obsessive-Compulsive Disorder.
    } 
    \vspace{-2mm}
    \label{tab:processed_data}
    \resizebox{\textwidth}{!}{%
    \begin{tabular}{@{}cl|ccccccccc@{}}
    \toprule
    \multicolumn{1}{c}{\textbf{Category}} &  \multicolumn{1}{l|}{\textbf{Datasets}}  & \textbf{Paradigm} & \textbf{\#Subjects} & \textbf{Sampling Rate} & \textbf{\#Channels} & \textbf{Total Time} & \textbf{\#Timestamps} & \textbf{\#Samples} & \textbf{Classes} & \textbf{Tasks}  \\ 
    \midrule

    \multicolumn{1}{c}{\textbf{}} & \multicolumn{1}{l|}{\textbf{BACA-RS}} & RS  & 608 &  1000-->200,100,50 Hz  & 19 &  169.80h  & 400 & 1,057,775 &  HC  &  Pre-train \\
    \multicolumn{1}{c}{\textbf{}} & \multicolumn{1}{l|}{\textbf{Depression}} & RS  & 122 &  500-->200,100,50 Hz  & 19 &  23.36h  & 400 & 146,062 &  HC, DEP  &  Pre-train \\
    \multicolumn{1}{c}{\textbf{}} & \multicolumn{1}{l|}{\textbf{FEPCR}} & RS  & 143 &  1000-->200,100,50 Hz  & 19 &  12.27h  & 400 & 76,694 &  HC, FEP  &  Pre-train \\
    \multicolumn{1}{c}{\textbf{}} & \multicolumn{1}{l|}{\textbf{MCEF-RS}} & RS  & 165 &   512-->200,100,50 Hz  & 19 &  14.14h  & 400 & 88,411 &  HC  &  Pre-train \\
    \multicolumn{1}{c}{\textbf{Non-AD}} & \multicolumn{1}{l|}{\textbf{PD-RS}} & RS  & 149 &  500-->200,100,50 Hz  & 19 &  6.64h  & 400 & 41,174 &  HC, PD  &  Pre-train \\
    \multicolumn{1}{c}{\textbf{}} & \multicolumn{1}{l|}{\textbf{PEARL-Neuro}} & RS  & 79 &  1000-->200,100,50 Hz  & 19 &  14.36h  & 400 & 90,147  &  HC  &  Pre-train \\
    \multicolumn{1}{c}{\textbf{}} & \multicolumn{1}{l|}{\textbf{SRM-RS}} & RS  & 109 &  1024-->200,100,50 Hz  & 19 &  9.10h  & 400 & 56,880 &  HC  &  Pre-train \\
    \multicolumn{1}{c}{\textbf{}} &\multicolumn{1}{l|}{\textbf{TDBrain}} & RS  & 1273 &  500-->200,100,50 Hz  & 19 &  89.71h  & 400 & 555,455 &  \makecell{MDD, ADHD, \\ SMC, OCD, etc.}  &  Pre-train \\
    \multicolumn{1}{c}{\textbf{}} & \multicolumn{1}{l|}{\textbf{TUEP}} & RS  & 200 &  256-->200,100,50 Hz  & 19 &  466.24h  & 400 & 2,930,241 &  HC, Epilepsy  &  Pre-train \\

    \midrule

    \multicolumn{1}{c}{\textbf{}} & \multicolumn{1}{l|}{\textbf{AD-Auditory}} & ASSR  & 35 &  250-->200,100,50 Hz & 19 & 5.20h & 400 & 32,655 & HC, AD, MCI, &  Pre-train \\
    \multicolumn{1}{c}{\textbf{}} & \multicolumn{1}{l|}{\textbf{BrainLat}} & RS  & 135 &  512-->200,100,50 Hz & 19 & 17.26h & 400 &  108,206 & HC, AD, FTD, PD, MS &  Pre-train \\
    \multicolumn{1}{c}{\textbf{}} & \multicolumn{1}{l|}{\textbf{P-ADIC}} & RS  & 249 &  500-->200,100,50 Hz  & 19 &  75.39h  & 400 & 473,970 &  \makecell{HC, AD, MCI, \\ SCZ, DEP}  & Pre-train \\
    \multicolumn{1}{c}{\textbf{AD}} & \multicolumn{1}{l|}{\textbf{CAUEEG}} & \makecell{RS, PS, \\ HV, etc.} & 1379 &  200-->200,100,50 Hz  & 19 &  282.37h  & 400 & 1,773,814  &  \makecell{HC, MCI, \\ DEM, etc.}  & Pre-train \\
    \cmidrule{2-11}
    \multicolumn{1}{c}{\textbf{}} & \multicolumn{1}{l|}{\textbf{ADFSU}} & RS & 92 &  128-->100,50 Hz  & 19 & 24.53m & 100  & 4048 &  HC, AD & Fine-tune \\
    \multicolumn{1}{c}{\textbf{}} & \multicolumn{1}{l|}{\textbf{ADFTD}} & RS, PS  & 88 &  500-->200,100,50 Hz  & 19 &  26.64h  & 400 & 167,083 &  HC, AD, FTD  &  Fine-tune \\
    \multicolumn{1}{c}{\textbf{}} & \multicolumn{1}{l|}{\textbf{ADSZ}} & RS & 48 &  128-->100,50 Hz  & 19 & 6.80m & 100  & 1128 &  HC, AD & Fine-tune \\
    \multicolumn{1}{c}{\textbf{}} & \multicolumn{1}{l|}{\textbf{APAVA}} & RS & 23 &  256-->200,100,50 Hz  & 16 & 0.92h & 200   & 9282 &  HC, AD & Fine-tune \\
    \multicolumn{1}{c}{\textbf{}} & \multicolumn{1}{l|}{\textbf{CNBPM}} & RS  & 189 &  256-->200,100,50 Hz  & 19 &  19.51h  & 400 & 122,029 &  HC, MCI, AD  &  Fine-tune \\

    \bottomrule
    \end{tabular}
    } 
    \vspace{-3mm}
\end{table*}

\subsection{Datasets Selection}
\label{sub:data_curation}

We categorize all EEG datasets utilized in this paper into two groups: those containing Alzheimer's Disease subjects (\textbf{AD} datasets) and those without (\textbf{Non-AD} datasets).

\textbf{AD Datasets.}
Despite the promise of EEG for AD detection, the field remains critically limited by the scarcity of accessible datasets. To address this, we conducted a comprehensive review of relevant publications and public EEG repositories (e.g., OpenNeuro, Dryad, figshare) published before 2026. This effort identified 8 public datasets. In total, we utilize 8 public and 1 private AD datasets, most of which consist of resting-state recordings, with minor evoked steady-state response: \textbf{AD-Auditory}~\citep{lahijanian2024auditory}, \textbf{BrainLat}~\citet{prado2023brainlat}, \textbf{P-ADIC}~\citep{shor2021eeg}, \textbf{CAUEEG}~\citep{kim2023deep}, \textbf{ADFSU}~\citep{vicchietti2023computational}, \textbf{ADFTD}~\citep{miltiadous2023dataset}, \textbf{ADSZ}~\citep{alves2022eeg}, \textbf{APAVA}~\citep{escudero2006analysis}, and \textbf{CNBPM}~\citep{amezquita2019novel}. Together, these datasets comprise \textbf{2,692,215 samples, 2,238 subjects, and 427.81 hours} of EEG, forming the \textbf{world's largest} EEG-AD corpus reported so far.

\textbf{Non-AD Datasets.}
While the inclusion of AD datasets represents a major step forward, their scale is still limited for large-scale pre-training, and a portion must be reserved for downstream evaluation. To further expand our training resources, we select 9 \textbf{domain-relevant} non-AD datasets. Our selected non-AD datasets contain recordings from both healthy subjects and patients with other brain disorders (e.g., PD, DEP, ADHD). Specifically, we include: \textbf{BACA-RS}~\citep{getzmann2024resting}, \textbf{Depression}~\citep{cavanagh2019multiple}, \textbf{FEPCR}~\citep{phalen2020non}, \textbf{MCEF-RS}~\citep{chenot2024investigating}, \textbf{PD-RS}~\citep{singh2023evoked}, \textbf{PEARL-Neuro}~\citep{dzianok2024pearl}, \textbf{SRM-RS}~\citep{hatlestad2022bids}, \textbf{TDBrain}~\citep{van2022two}, and \textbf{TUEP}~\citep{veloso2017big}. Collectively, these datasets comprise \textbf{5,042,839 samples, 2,848 subjects, and 805.62 hours} of EEG. Most of them were acquired under the resting-state paradigm.

\subsection{Unified Data Preprocessing}
\label{sub:data_preprocessing}

The statistics of processed datasets are provided in Table~\ref{tab:processed_data}. The datasets are highly heterogeneous, with substantial variability in channel numbers, sampling rates, and recording lengths. To utilize them for training, we apply the following preprocessing pipeline sequentially: 

\textbf{1) Removal of non-EEG channels:} All non-EEG channels are removed, such as EOG, ECG, or coordinate information. \textbf{2) Notch and Band-Pass Filtering:} Each trial undergoes a notch filter at 50 Hz or 60 Hz, followed by a band-pass filter between 0.5 Hz and 45 Hz. This step suppresses line noise, slow drifts, and high-frequency noise, which are generally outside the frequency range of scalp-recorded brain activity. \textbf{3) Average Re-referencing:} Average re-referencing is applied to reduce global noise and potential baseline shifts. \textbf{4) Artifact Removal:} For datasets lacking prior artifact rejection, we utilize independent component analysis (ICA) combined with the ICLabel algorithm~\citep{pion2019iclabel} to automatically identify and remove components associated with artifacts like eye blinks or muscle activity. \textbf{5) Channel Alignment:} We align all pre-training datasets (downstream datasets could be an arbitrary number of channels and montage) to the standard 19-channel montage based on the international 10-20 system: Fp1, Fp2, F7, F3, Fz, F4, F8, T3/T7, C3, Cz, C4, T4/T8, T5/P7, P3, Pz, P4, T6/P8, O1, and O2~\citep{homan1987cerebral}. If a dataset has more channels, we select only the 19 channels with these names and discard the rest. For datasets using different montages (e.g., Biosemi~\citep{biosemi}), we select the 19 closest channels in 3D coordinates. All the fine-tuning datasets keep the raw channels. \textbf{6) Multi-Scale Segmentation:} We employ a multi-scale segmentation strategy to downsample EEG recordings to a commonly used sampling rate and segment them into fixed-length training samples (see para.~\ref{para:multi_scale_segmentation} for details). \textbf{7) Metadata Construction:} A metadata JSON file is created for each dataset to store information such as the channel montage, channel names, and sampling rates for subsequent model design. \textbf{8) Z-Score Normalization:} Z-score normalization is applied to each segmented sample independently for each channel. This step is applied during data loading.

\section{Method}
\label{sec:method}

\subsection{Preliminaries}
\label{sub:preliminaries}

\textbf{Neurodegenerative Disease Detection.}
\label{para:disease_detection}
Neurodegenerative disease detection using EEG typically involves assigning a single label to each subject, reflecting their physiological brain state at the time of data acquisition, such as the presence or absence of AD. This formulation assumes that a subject’s state remains stable over the relatively short recording period (e.g., several days to several months), so that all EEG data collected from the same subject share the same label. Exceptions arise with longitudinal datasets that track disease progression (e.g., Healthy → Mild Cognitive Impairment → AD~\citep{vecchio2018sustainable,ge2025eeg}); in such cases, trials/sessions reflecting a change in diagnosis for the same subject are treated as different subjects during training. Besides, comorbid diseases are often reported in neurodegenerative diseases. For instance, a patient may present with both AD and Vascular Dementia (VD)~\citep{kim2023deep}. This introduces an inherently multi-label scenario, in which each disease should be predicted independently via a dedicated classifier formulated as a binary classification task. In this work, consistent with the datasets used in our experiments, \textbf{we simplify the problem formulation by assuming each subject is assigned a single label, with no label transitions during data collection and no comorbid conditions.}

\textbf{Problem Formulation.}
\label{para:problem_formulation}
EEG-based AD detection aims to identify AD and related cognitive conditions from EEG recordings. While most existing deep learning methods focus on classifying individual EEG samples (sample-level classification), our approach emphasizes end-to-end subject-level detection, which is more clinically relevant. Consider an input EEG sample $\bm{X} \in \mathbb{R}^{C \times T}$, where $C$ denotes the number of channels (electrodes) and $T$ is the number of timestamps. Let the corresponding class label be $y \in \{1,\dots,K\}$, where $K$ is the number of categories, such as AD, Healthy Controls (HC), Mild Cognitive Impairment (MCI), Frontotemporal Dementia (FTD), or other dementia-related conditions. We further assign each sample a subject ID $s$ and a sampling rate marker $r$ for subsequent use. We formulate two classification tasks: \textbf{(1) Sample-Level Classification:} The objective is to learn a model $f(\cdot)$ that predicts the class label $y$ for each EEG sample $\bm{X}$. \textbf{(2) Subject-Level Detection:} Given multiple EEG samples belonging to the same subject $s$, we aggregate their sample-level predictions using majority voting (Sec.~\ref{para:majority_vote}). The final subject-level prediction is determined by the class receiving the highest number of votes across all samples from subject $s$.

\begin{figure*}[t]
    \centering    \includegraphics[width=1.0\linewidth]{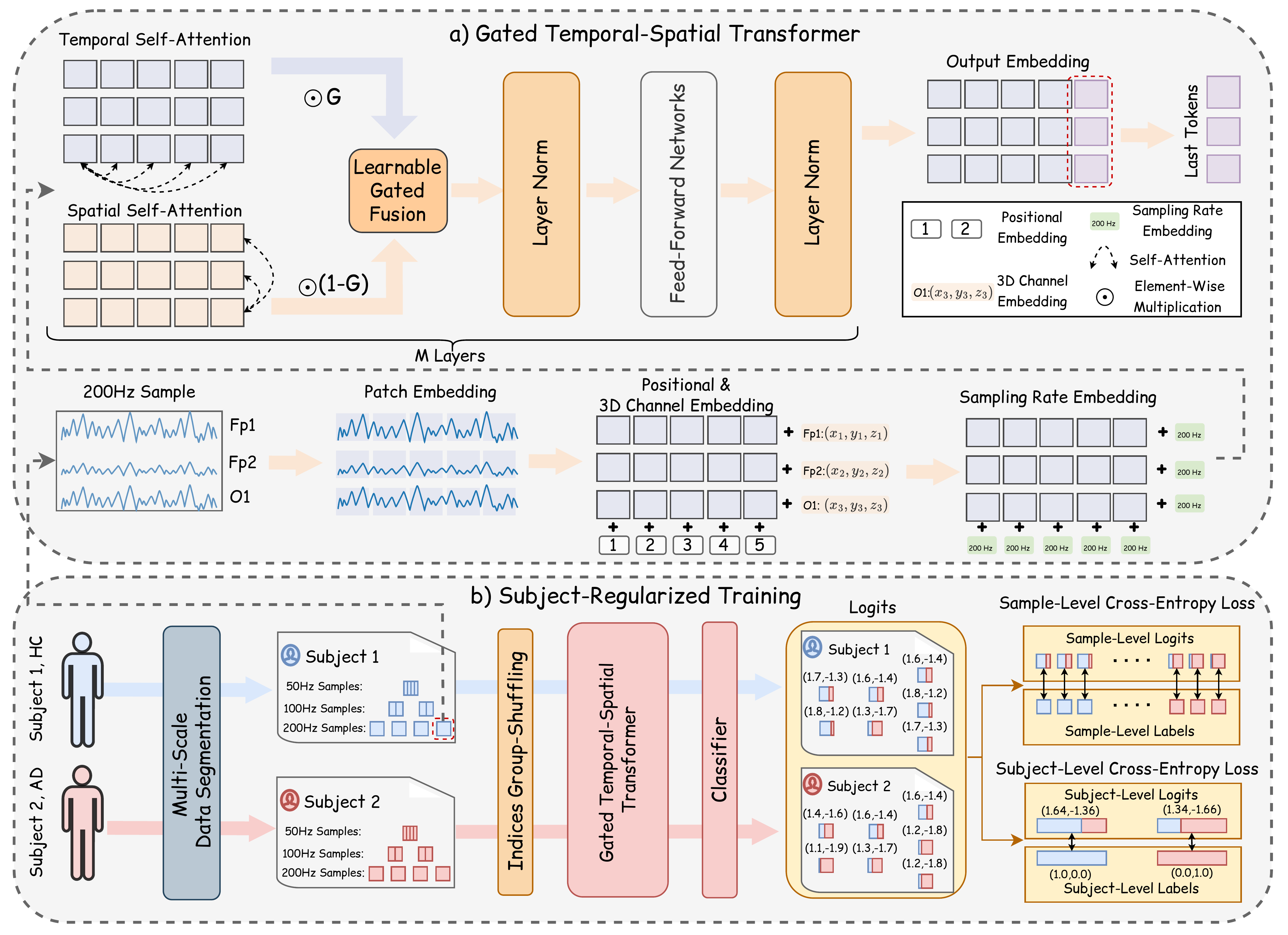}
    \caption{ \textbf{a) Gated Temporal-Spatial Transformer.} Input EEG samples are first divided into univariate patches. These patches are projected into patch embeddings, to which temporal positional embeddings, 3D channel embeddings, and sampling rate embeddings are added to form the final token representations. A parallel temporal-spatial self-attention mechanism is then applied along both the temporal and channel dimensions. The resulting features from the two branches are adaptively combined through a learnable gated fusion module. Standard Transformer components, including layer norm and feedforward networks, follow the gated fusion stage. Finally, the last token of each channel is used as the downstream representation. \textbf{b) Subject-Regularized Training.} Each subject's EEG recordings are segmented into samples with multiple sampling rates (e.g., 200, 100, and 50 Hz) using the proposed multi-scale segmentation strategy. Index group shuffling ensures that each training batch contains sufficient samples from the same subject, thereby facilitating subject-level learning. The classifier produces sample-level logits, which are further aggregated into subject-level logits. We then compute cross-entropy losses at both the sample and subject levels.
    }
    \label{fig:lead_details}
    \vspace{-5mm}
\end{figure*}

\subsection{Gated Temporal-Spatial Transformer}
\label{sub:temporal_spatial_transformer}

\textbf{Multi-Scale Segmentation.}
\label{para:multi_scale_segmentation}
Instead of resampling EEG signals to a single fixed sampling rate, we downsample each recording to multiple commonly used sampling rates. Specifically, signals with original rates above 200 Hz are downsampled to 200, 100, and 50 Hz. We then segment the signals into fixed-length samples with 50\% overlap, using 100, 200, or 400 timestamps depending on the dataset. If the original sampling rate is below 200 Hz, such as 128 Hz, we use only 100 and 50 Hz. The sample length is fixed in terms of timestamps; for example, a 400-timestamp sample at 200 Hz corresponds to a 2-s EEG segment. In addition to the class label $y$ and subject ID $s$, each segmented sample is assigned a sampling rate label $r$ corresponding to its downsampled sampling rate. This segmentation strategy naturally enables multi-scale feature extraction, allowing the model to capture both fine-grained and coarse temporal patterns. Moreover, since our ultimate goal is subject-level detection, this approach increases the total number of samples per subject, which benefits both model training and post-hoc majority voting.

\textbf{Patch Embedding.}
\label{para:patch_embedding}
To handle EEG samples with arbitrary channel configurations and variable recording lengths, we implement a patch embedding strategy as follows. Consider an input EEG sample $\bm{X} \in \mathbb{R}^{C \times T}$, for a patch length $L$, we slice each channel into $N$ univariate patches. 
This yields a patch tensor $\bm{X}_{p} \in \mathbb{R}^{C \times N \times L}$. Zero padding is applied when necessary to ensure that $T$ is divisible by $L$, resulting in $N = \left\lceil T / L \right\rceil$. 
Each patch is then projected into a $D$-dimensional embedding space through a linear mapping $\bm{W} \in \mathbb{R}^{L \times D}$, producing patch embeddings $\bm{E} \in \mathbb{R}^{C \times N \times D}$. A fixed temporal positional embedding $\bm{TE} \in \mathbb{R}^{N \times D}$ is added and broadcast across all $C$ channels. The final patch embeddings are obtained as:
\begin{equation}
    \bm{E} \leftarrow \bm{X}_{p} \bm{W} + \operatorname{Broadcast}_{C}{(\bm{TE})}, 
    \quad \bm{E} \in \mathbb{R}^{C \times N \times D}
\end{equation}

\textbf{3D Channel Embedding.}
\label{para:3d_channel_embedding}
To enable the model to flexibly adapt to arbitrary channel configurations and learn spatial relationships among channels, we employ the 3D EEG channel embedding introduced in NeurIPT~\citep{fang2025neuript}, with minor modifications. During preprocessing, each dataset stores its channel montage and channel names in a metadata JSON file. Given the montage type, such as the international 10-20 system~\citep{homan1987cerebral}, and channel names, such as Fp1 and Fp2, we retrieve the corresponding 3D electrode coordinates using the MNE-Python package~\citep{gramfort2013meg}. Let $\bm{P} \in \mathbb{R}^{C \times 3}$ denote the 3D coordinate matrix of the $C$ EEG channels, where each channel is represented by its spatial coordinates $(x_c, y_c, z_c)$ under the corresponding EEG montage. To encode spatial information, we apply sinusoidal positional encoding independently to the $x$, $y$, and $z$ coordinates. The embedding dimension $D$ is approximately evenly divided among the three spatial axes, and the resulting axis-wise embeddings are concatenated to obtain a $D$-dimensional representation for each channel:
\begin{equation}
    \bm{e}_c =
    \mathrm{PE}(x_c)
    \,\Vert\,
    \mathrm{PE}(y_c)
    \,\Vert\,
    \mathrm{PE}(z_c)
    \in \mathbb{R}^{D}, \quad c = 1, \dots, C.
\end{equation}
where $\mathrm{PE}(\cdot)$ denotes the standard sinusoidal positional encoding and $\Vert$ denotes concatenation. Stacking the embeddings of all channels $\{\bm{e}_c\}_{c=1}^{C}$ gives the initial 3D channel embedding $\bm{CE} \in \mathbb{R}^{C \times D}$. 
The channel embeddings are then broadcast along the temporal dimension and added to the patch embeddings:
\begin{equation}
    \bm{E}
    \leftarrow
    \bm{E} + \operatorname{Broadcast}_{N}(\bm{CE}),
    \quad
    \bm{E} \in \mathbb{R}^{C \times N \times D}.
\end{equation}

\textbf{Sampling Rate Embedding.}
\label{para:sampling_rate_embedding}
Since the sampling rate critically affects the temporal characteristics of EEG signals, even for recordings of identical duration, we introduce a sampling rate embedding module to explicitly encode the temporal scale of each input. This design enables the model to flexibly generalize across multiple sampling rates in both pre-training and downstream tasks, while naturally supporting the multi-scale segmentation strategy described in the previous paragraph. Given that each EEG sample is associated with a sampling rate label $r$, we define a learnable embedding lookup table that maps each sampling rate to a $D$-dimensional embedding vector $\bm{e}_r \in \mathbb{R}^{D}$. The sampling rate embedding is broadcast along both the channel and temporal dimensions and added to all $C \times N$ patch embeddings:
\begin{equation}
    \bm{E} \leftarrow \bm{E} + \operatorname{Broadcast}(\bm{e}_r), 
    \quad \bm{E} \in \mathbb{R}^{C \times N \times D}.
\end{equation}

\textbf{Parallel Temporal-Spatial Self-Attention.}
\label{para:parallel_temporal_spatial_attention}
Given the input representation $\bm{E}\in\mathbb{R}^{C\times N\times D}$, where $C$ denotes the number of channels, $N$ the number of temporal patch embeddings per channel, and $D$ the embedding dimension, we apply temporal and spatial self-attention in parallel to jointly model temporal dynamics and spatial dependencies. \textbf{a) Temporal Self-Attention.} We use $\bm{E}_c \in \mathbb{R}^{N \times D}$ to denote the sequence of patch embeddings for the $c$-th channel. Self-attention is applied independently to each channel along the temporal dimension:
\begin{equation}
    \bm{E}_c \leftarrow \operatorname{Attn}(\bm{E}_c,\bm{E}_c,\bm{E}_c)
    \in \mathbb{R}^{N \times D},
    \quad c = 1, \dots, C.
\end{equation}
This allows the model to capture temporal dependencies within each channel $c$. After updating all channels, we stack the outputs $\bm{E}_c$ along the channel dimension to obtain $\bm{E}^{T} \in \mathbb{R}^{C \times N \times D}$. \textbf{b) Spatial Self-Attention.} We use $\bm{E}_n \in \mathbb{R}^{C \times D}$ to denote the set of channel embeddings at the $n$-th temporal patch.  Self-attention is applied independently across channels at each temporal patch index:
\begin{equation}
    \bm{E}_n \leftarrow \operatorname{Attn}(\bm{E}_n,\bm{E}_n,\bm{E}_n)
    \in \mathbb{R}^{C \times D},
    \quad n = 1, \dots, N.
\end{equation}
This enables the model to capture spatial correlations among channels at each temporal patch index $n$. After updating all patch indices, we stack the outputs $\bm{E}_n$ along the temporal dimension to obtain $\bm{E}^{S} \in \mathbb{R}^{C \times N \times D}$. \textbf{c) Gated Fusion.} In real-world applications, EEG data collected from different devices, protocols, and tasks may exhibit varying relative importance of temporal and spatial patterns. We therefore design a learnable gated fusion module that allows the backbone model to adaptively determine the contribution of temporal and spatial features based on their respective outputs. Specifically, an element-wise gate is computed as:
\begin{equation}
    \bm{G} = \sigma\!\left( \big[ \bm{E}^{T} \,\Vert\, \bm{E}^{S} \big]\bm{W}_{f} \right),
    \quad \bm{G} \in \mathbb{R}^{C \times N \times D},
\end{equation}
where $\sigma(\cdot)$ denotes the sigmoid function, $\Vert$ denotes concatenation, and $\bm{W}_{f} \in \mathbb{R}^{2D \times D}$ is a learnable projection matrix. The final representation is then obtained through element-wise gated fusion:
\begin{equation}
    \bm{E} = \bm{G} \odot \bm{E}^{T} + (1 - \bm{G}) \odot \bm{E}^{S},
    \quad \bm{E} \in \mathbb{R}^{C \times N \times D},
\end{equation}
where $\odot$ denotes element-wise multiplication. This adaptive fusion mechanism enables the model to automatically learn the relative importance of temporal and spatial features for different datasets and tasks.

\textbf{Classifier.}
\label{para:classifier}
After $M$ stacked layers of self-attention, LayerNorm, and a feed-forward network with SwiGLU activation~\citep{shazeer2020glu}, the model outputs an embedding $\bm{E} \in \mathbb{R}^{C \times N \times D}$. The downstream representation $\bm{E}_{d} \in \mathbb{R}^{C \times D}$ is obtained by selecting the \textbf{last token} along the temporal dimension for each channel from $\bm{E}$. This design serves a role similar to using a per-channel \texttt{[CLS]} token, but avoids explicit token initialization and allows the backbone to adapt to downstream datasets with different numbers of channels. The representation $\bm{E}_{d}$ is then flattened and fed into a linear classifier $c(\cdot)$ to produce predictions.

\subsection{Subject-Regularized Training}
\label{sub:subject_regularized_training}

\textbf{Majority Voting.}
\label{para:majority_vote}
Since the model is trained on segmented fixed-length EEG samples, it naturally produces sample-level predictions during inference. Therefore, an additional post-processing step is required to obtain final subject-level detections. Following prior studies~\citep{ieracitano2019convolutional}, we adopt a majority voting strategy. Specifically, for a subject $s$ with $N_s$ EEG samples, we define the subject-level confidence score for class $k$ as the proportion of samples predicted as that class:
\begin{equation}
    q_{s,k} = \frac{1}{N_s}\sum_{i=1}^{N_s}\mathbb{I}(\hat{y}_{s,i}=k),
    \quad k \in \{1,\dots,K\},
\end{equation}
where $\hat{y}_{s,i}$ denotes the predicted class of the $i$-th sample in subject $s$ and $\mathbb{I}(\cdot)$ is the indicator function. The final subject-level prediction is then obtained by majority voting:
\begin{equation}
    \hat{y}_{s} = \arg\max_{k} q_{s,k}.
\end{equation}

For metrics based on discrete predictions, including Accuracy, Precision, Recall, Specificity, and F1, we directly use the majority-voted subject-level label $\hat{y}_{s}$. In contrast, AUROC and AUPRC require continuous confidence scores rather than hard class labels. We therefore use $q_{s,k}$ as the subject-level confidence score for each class for the calculation of AUROC and AUPRC. For example, if 80 out of 100 EEG samples from a subject are predicted as AD, the corresponding subject-level AD confidence score is 0.8.

\begin{algorithm}[h]
\caption{Subject-Level Cross-Entropy Loss}
\label{alg:subject_ce}
\begin{algorithmic}[1]
\Require Batch logits $\{\mathbf{z}_i\}_{i=1}^{B}$, labels $\{y_i\}_{i=1}^{B}$, and subject IDs $\{s_i\}_{i=1}^{B}$ \Comment{$B$ is the batch size}
\Ensure Subject-level cross-entropy loss $\mathcal{L}_{\mathrm{ce}}^{\mathrm{sub}}$

\State $\mathcal{U} \gets \mathrm{Unique}(\{s_i\}_{i=1}^{B})$ \Comment{Unique subject IDs in the current batch}
\State $\mathcal{Z} \gets [\,]$, $\mathcal{Y} \gets [\,]$ \Comment{Subject-level logits and labels list}

\For{each subject $u \in \mathcal{U}$}
    \State $\mathcal{M}_u \gets \{i \mid s_i=u\}$ \Comment{Samples from subject $u$}
    \State $\bar{\mathbf{z}}_u \gets \frac{1}{|\mathcal{M}_u|}\sum_{i\in\mathcal{M}_u}\mathbf{z}_i$ \Comment{Average subject-level logit}
    \State $y_u \gets y_i$ for any $i \in \mathcal{M}_u$ \Comment{All samples share the same subject label}
    \State $\mathcal{Z} \gets \mathcal{Z} \cup \{\bar{\mathbf{z}}_u\}$
    \State $\mathcal{Y} \gets \mathcal{Y} \cup \{y_u\}$
\EndFor

\State \Return $\mathrm{CrossEntropy}(\mathcal{Z},\mathcal{Y})$
\end{algorithmic}
\end{algorithm}

\textbf{Subject-Level Cross-Entropy Loss.}
\label{para:subject_level_cross_entropy}
Previous studies have shown that high sample-level prediction accuracy does not necessarily translate into strong subject-level detection performance~\citep{wang2024adformer}, since subject-level predictions are typically computed post hoc via majority voting and do not directly participate in training (See examples in Appendix~\ref{sec:subject_performance_example}). To better align model training with the final subject-level detection objective, we introduce a subject-level cross-entropy loss that provides additional supervision by aggregating predictions from samples belonging to the same subject. To distinguish it from the conventional sample-level cross-entropy loss $\mathcal{L}_{\mathrm{ce}}^{\mathrm{sam}}$, we denote the proposed subject-level loss as $\mathcal{L}_{\mathrm{ce}}^{\mathrm{sub}}$. The pseudocode is provided in Algorithm~\ref{alg:subject_ce}. For each unique subject $s$ appearing in a training batch, we compute a subject-level logit by averaging the logits of all samples from that subject. Since majority voting is non-differentiable, this logit aggregation serves as a differentiable surrogate for subject-level prediction during training. The subject-level loss $\mathcal{L}_{\mathrm{ce}}^{\mathrm{sub}}$ is then computed as the cross-entropy between the aggregated logits and the corresponding subject labels. In this way, samples from the same subject are jointly optimized to support the correct subject-level prediction. The final training objective combines sample-level and subject-level supervision:
\begin{equation}
\mathcal{L}_{\mathrm{ce}}
=
\alpha \mathcal{L}_{\mathrm{ce}}^{\mathrm{sam}}
+
\beta \mathcal{L}_{\mathrm{ce}}^{\mathrm{sub}},
\quad
\alpha + \beta = 1,\quad
\alpha,\beta \in [0,1],
\end{equation}
where $\alpha$ and $\beta$ are balancing coefficients.

\subsection{Self-Supervised Pre-training}
\label{sub:self_supervised_pretraining}

\textbf{Medical Contrastive Learning.}
\label{para:medical_contrastive_learning}
We employ \textbf{sample-level} and \textbf{subject-level} contrastive learning for self-supervised pre-training defined in COMET~\citep{wang2023contrast}. The sample-level contrastive learning module performs instance discrimination, treating different augmented views of the same sample as positive pairs and views from other samples as negative pairs. The augmentation methods are described in Appendix~\ref{sec:data_augmentation_banks}. The subject-level contrastive learning module leverages subject IDs as guidance: samples from the same subject are treated as positive pairs, while those from different subjects are treated as negative pairs. 
Compared with masked autoencoding (MAE)~\citep{he2022masked}, subject-level contrastive learning is more suitable for our goal of \textbf{capturing subject-level representations} for downstream AD detection, under the assumption that all samples from a subject collected in a short period share the same label. The overall self-supervised pre-training objective is a weighted sum of the two contrastive module losses: 

\begin{equation}
\mathcal{L}_{co}
=
{\lambda_1}\mathcal{L}^{sam}_{co}
+
{\lambda_2} \mathcal{L}^{sub}_{co},
\quad \lambda_1 + \lambda_2 = 1, \lambda_1, \lambda_2 \in [0,1]
\end{equation}
where $\lambda_1$ and $\lambda_2$ are balancing coefficients.

\begin{algorithm}[h]
\caption{Index Group Shuffling}
\label{alg:index_group_shuffling}
\begin{algorithmic}[1]
\Require Subject IDs $\{s_i\}_{i=1}^{N}$, group size $G$, batch size $B$, $G \leq B$
\Ensure A shuffled list of mini-batches $\mathcal{B}$

\State $\mathcal{U} \gets \mathrm{Unique}(\{s_i\}_{i=1}^{N})$ \Comment{Get unique subject IDs}
\State $\mathcal{G} \gets [\,]$ \Comment{Initialize the list of sample groups}

\For{each subject $u \in \mathcal{U}$}
    \State $\mathcal{I}_u \gets \{i \mid s_i = u\}$ \Comment{Indices of samples from subject $u$}
    \State $\mathrm{Shuffle}(\mathcal{I}_u)$
    \State Split $\mathcal{I}_u$ into groups of at most $G$ samples
    \State Append all groups to $\mathcal{G}$
\EndFor

\State $\mathrm{Shuffle}(\mathcal{G})$ \Comment{Shuffle group order without breaking individual groups}
\State $\mathcal{B} \gets [\,]$

\While{$\mathcal{G}$ is not empty}
    \State Initialize an empty mini-batch $\mathbf{b}$
    \State Add groups from $\mathcal{G}$ to $\mathbf{b}$ until $|\mathbf{b}| \approx B$
    \State $\mathrm{Shuffle}(\mathbf{b})$ \Comment{Shuffle sample order within the mini-batch}
    \State $\mathcal{B} \gets \mathcal{B} \cup \{\mathbf{b}\}$
\EndWhile

\State \Return $\mathcal{B}$
\end{algorithmic}
\end{algorithm}

\textbf{Index Group Shuffling.}
\label{para:index_group_shuffling}
In practical training with a large number of subjects, the probability that multiple samples from the same subject appear within the same mini-batch decreases, reducing the effectiveness of subject-level contrastive learning and potentially causing the subject-level objective to degenerate toward sample-level contrastive learning. To mitigate this issue, we introduce an index group shuffling algorithm as an engineering trick that dynamically reorganizes training indices at each epoch. The pseudocode is provided in Algorithm~\ref{alg:index_group_shuffling}. The computational overhead is negligible, as the algorithm only groups and shuffles sample indices. Specifically, samples are first grouped by subject ID. Within each subject, samples are randomly shuffled and divided into chunks of up to $G$ samples. These chunks are then shuffled and combined to form mini-batches, increasing the likelihood that multiple samples from the same subject co-occur within a batch while preserving randomness across batches.

\section{Experiments}
\label{sec:experiments}

\subsection{Setup}
\label{sub:setup}

\textbf{Pre-training and Downstream Datasets.}
For pre-training, we leverage all 9 Non-AD datasets together with 4 AD datasets, resulting in a total of \textbf{7,431,484 samples, 4,646 subjects, and 1,185.84 hours.} The remaining 5 AD datasets are reserved for downstream evaluation, resulting in a total of \textbf{303,570 samples, 440 subjects, and 47.59 hours.} More details are presented in Section~\ref{sec:preliminary}.

\textbf{Baselines.}
We compare our method against \textbf{16} baselines, including 1 manual-feature approach, 11 supervised deep learning methods, and 4 EEG foundation models. These selected baselines are SOTA methods or have shown strong performance in EEG classification tasks. The feature-based method extracts \textbf{Statistical}, \textbf{Spectral}, \textbf{Power}, and \textbf{Complexity} features (see Appendix~\ref{sec:implementation_details}) commonly used for EEG-based AD detection, followed by classification with a linear classifier. The 11 supervised learning methods include \textbf{EEGConformer}~\citep{song2022eeg}, \textbf{EEGInception}~\citep{zhang2021eeg}, \textbf{EEGNet}~\citep{lawhern2018eegnet}, \textbf{iTransformer}~\citep{liuitransformer}, \textbf{MedGNN}~\citep{fan2025towards}, \textbf{Medformer}~\citep{wang2024medformer}, \textbf{MNet}~\citep{watanabe2024deep}, \textbf{ModernTCN}~\citep{luo2024moderntcn}, \textbf{PatchTST}~\citep{nie2022time}, \textbf{TCN}~\citep{bai2018empirical}, \textbf{TimesNet}~\citep{wu2023timesnet}. The 4 EEG foundation models are \textbf{BIOT}~\citep{yang2024biot}, \textbf{LaBraM}~\citep{jiang2024large},  \textbf{CBraMod}~\citep{wang2024cbramod}, \textbf{CSBrain}~\citep{zhou2025csbrain}.

\textbf{Training \& Parameter Settings.}
Self-supervised pre-training is conducted for 30 epochs without early stopping, followed by fine-tuning for up to 200 epochs with early stopping (patience=15) based on the best sample-level F1. Batch sizes are 2048 for pre-training and 512 for supervised learning, linear probing, and fine-tuning. We use AdamW with learning rates of $2\!\times\!10^{-4}$ (pre-training) and $1\!\times\!10^{-4}$ (supervised learning, linear probing, fine-tuning), scheduled by CosineAnnealingLR. We employ 14 evaluation metrics, including sample-level accuracy, precision (macro-averaged), sensitivity (macro-averaged), specificity (macro-averaged), F1 score (macro-averaged), AUROC (macro-averaged), and AUPRC (macro-averaged), and their corresponding subject-level metrics computed via majority voting (Para.~\ref{para:majority_vote}). The contrastive loss coefficients in self-supervised pre-training are set to $\lambda_1=0.25$ and $\lambda_2=0.75$. Both sample-level and subject-level cross-entropy losses are set to $\alpha=\beta=0.5$ in the supervised learning, linear probing, and fine-tuning stages. 
The $G$ for index group shuffling is set to 32 for the pre-training and 8 for supervised learning, linear probing, and fine-tuning stages, respectively. For self-supervised pre-training, all subjects from the pre-training datasets are used. For fully supervised learning, linear probing, and fine-tuning, we adopt Monte Carlo cross-validation~\citep{xu2001monte} with a subject-independent 8:1:1 train/validation/test split (except for APAVA, which used a fixed split, the same as Medformer~\citep{wang2024medformer}), ensuring no subject overlap while preserving randomness across seeds (See Appendix~\ref{sec:evaluation_method} for more explanations). To enable cross-domain transfer, no pre-training subjects or datasets are used in fine-tuning. For 4 foundation model baselines, we fine-tune their officially released checkpoints, since the pre-training method and selection of pre-training resources are integral components of a foundation model. Each evaluation is repeated with five random seeds (41-45), reporting the mean and standard deviation. Experiments are run on 4 NVIDIA RTX A6000 GPUs with Python~3.10 and PyTorch~2.5.1+cu121. More details for each method are in Appendix~\ref{sec:implementation_details}.

\begin{figure*}[h]
    \centering
    \includegraphics[width=1.0\linewidth]{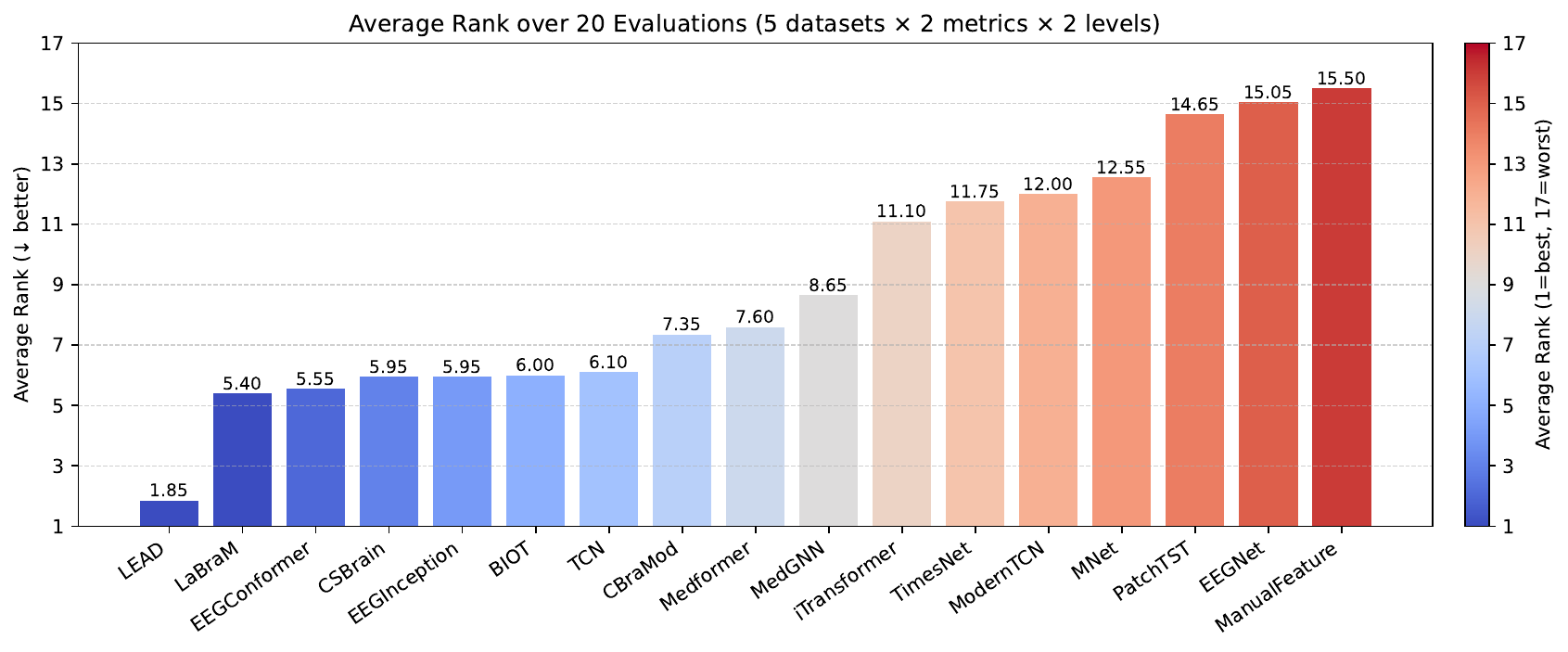}
    \caption{\textbf{Average Performance Rank.} Average performance rank of 17 methods across all 20 evaluations. For example, the value 1.85 of \myname indicates an average rank of 1.85. \textbf{Lower ranks and a deeper blue indicate better performance}.
    }
    \label{fig:avg_rank_bar_heatmap}
    \vspace{-2mm}
\end{figure*}

\begin{table*}[!t]
    \centering
    \def\arraystretch{1.0}
    \caption{\textbf{Method Comparison Results.} F1 score and AUROC comparison results of 16 baselines and our method. The full results for all 7 metrics are available in Appendix~\ref{sub:method_comparison_detail}.
    The \textcolor{myred}{\textbf{Top-1}}, \textcolor{myblue}{Top-2}, and \textcolor{mygreen}{Top-3} results are highlighted in red, blue, and green.
    }
    \vspace{-2mm}
    \label{tab:method_comparison_detail}
    \resizebox{\textwidth}{!}{
    \begin{tabular}{@{}ll|cc|cc|cc|cc|cc@{}}
    \toprule

    \multicolumn{2}{l|}{\textbf{Datasets}}
    & \multicolumn{2}{c|}{\makecell{\textbf{ADFSU} \\ \textit{(4,048 Samples)} \\ \textit{(92 Subjects)} \\ \textit{(HC vs AD)}}}
    & \multicolumn{2}{c|}{\makecell{\textbf{ADFTD} \\ \textit{(167,083 Samples)} \\ \textit{(88 Subjects)} \\ \textit{(HC vs AD vs FTD)}}}
    & \multicolumn{2}{c|}{\makecell{\textbf{ADSZ} \\ \textit{(1,128 Samples)} \\ \textit{(48 Subjects)} \\ \textit{(HC vs AD)}}}
    & \multicolumn{2}{c|}{\makecell{\textbf{APAVA} \\ \textit{(9,282 Samples)} \\ \textit{(23 Subjects)} \\ \textit{(HC vs AD)}}}
    & \multicolumn{2}{c}{\makecell{\textbf{CNBPM}  \\ \textit{(122,029 Samples)} \\ \textit{(189 Subjects)} \\ \textit{(HC vs MCI vs AD)}}}
    \\ \midrule

    \multicolumn{2}{l|}{\diagbox{\textbf{Methods}}{\textbf{Metrics}}} & \textbf{F1 Score} & \textbf{AUROC} & \textbf{F1 Score} & \textbf{AUROC}  & \textbf{F1 Score} & \textbf{AUROC}  & \textbf{F1 Score} & \textbf{AUROC}  & \textbf{F1 Score} & \textbf{AUROC} \\ 
    
    \midrule

    \multicolumn{2}{c}{} & \multicolumn{8}{c}{\textbf{Sample-Level Classification}}  \\
    \midrule

    \multicolumn{2}{l|}{\textbf{ManualFeature}}  & 75.67\std{11.41} & 80.90\std{8.87} & 44.73\std{3.88} & 62.89\std{4.31} & 58.42\std{21.26} & 67.20\std{15.34} & 63.22\std{6.51} & 66.49\std{5.78} & 43.00\std{6.36} & 61.29\std{6.44} \\
    \multicolumn{2}{l|}{\textbf{EEGConformer}}  & \textcolor{myblue}{94.22\std{3.28}} & \textcolor{mygreen}{98.87\std{0.95}} & 58.59\std{3.84} & 80.61\std{3.90} & \textcolor{myblue}{94.73\std{3.47}} & \textcolor{mygreen}{98.50\std{1.48}} & 73.05\std{4.93} & 83.07\std{3.70} & 59.17\std{8.85} & 81.76\std{8.20} \\
    \multicolumn{2}{l|}{\textbf{EEGInception}}   & 90.88\std{2.13} & 98.73\std{0.75} & 62.69\std{4.06} & 84.01\std{2.76} & 92.60\std{4.44} & 98.06\std{1.66}  & 71.87\std{2.22} & 80.17\std{3.54}  & 58.56\std{8.01} & 81.73\std{8.18} \\
    \multicolumn{2}{l|}{\textbf{EEGNet}}  & 74.48\std{9.91} & 85.88\std{7.95} & 40.31\std{6.74} & 60.93\std{4.42} & 64.90\std{12.88} & 77.18\std{19.10} & 67.84\std{4.60} & 78.04\std{4.19} & 38.57\std{6.52} & 63.51\std{7.95} \\
    \multicolumn{2}{l|}{\textbf{iTransformer}}  & 76.80\std{2.31} & 88.16\std{1.86} & 54.33\std{3.67} & 73.98\std{4.02} & 73.13\std{4.75} & 81.86\std{4.90} & 72.13\std{2.30} & 85.53\std{1.00} & 51.37\std{8.02} & 74.99\std{10.16} \\
    \multicolumn{2}{l|}{\textbf{MedGNN}}  & 90.09\std{2.37} & 98.06\std{1.42} & 67.67\std{4.59} & 85.52\std{2.57} & 88.59\std{8.01} & 95.90\std{4.10} & 67.54\std{7.95} & 81.39\std{3.83} & 58.68\std{7.50} & \textcolor{mygreen}{82.13\std{7.35}} \\
    \multicolumn{2}{l|}{\textbf{Medformer}}  & 87.15\std{1.91} & 97.48\std{1.32} & 64.83\std{6.72} & 84.13\std{3.27} & 88.73\std{5.76} & 94.18\std{3.72} & 67.42\std{2.44} & 78.38\std{2.75} & \textcolor{myred}{\textbf{60.39\std{7.68}}} & \textcolor{myblue}{82.17\std{7.87}} \\
    \multicolumn{2}{l|}{\textbf{MNet}}  & 76.19\std{8.72} & 92.75\std{5.08} & 59.91\std{9.76} & 82.30\std{5.04} & 79.21\std{7.35} & 88.87\std{5.92} & 53.76\std{10.86} & 81.13\std{4.02} & 47.04\std{9.76} & 77.03\std{11.79} \\
    \multicolumn{2}{l|}{\textbf{ModernTCN}}  & 75.09\std{3.18} & 91.94\std{1.84} & 58.53\std{3.71} & 77.25\std{3.33} & 75.73\std{4.05} & 84.42\std{4.30} & 58.42\std{0.76} & 75.88\std{2.41} & 55.95\std{8.11} & 77.85\std{9.30} \\
    \multicolumn{2}{l|}{\textbf{PatchTST}}  & 76.03\std{4.27} & 89.12\std{1.95} & 50.95\std{4.04} & 71.30\std{4.75} & 71.05\std{10.02} & 79.10\std{12.54} & 52.39\std{2.65} & 73.76\std{3.11} & 49.78\std{6.10} & 71.52\std{9.58} \\
    \multicolumn{2}{l|}{\textbf{TCN}}  & 90.10\std{3.13} & 98.82\std{0.73} & 64.26\std{2.40} & 82.94\std{1.89} & 88.93\std{9.23} & 94.02\std{6.85} & \textcolor{mygreen}{74.81\std{4.47}} & \textcolor{mygreen}{85.84\std{3.92}} & 58.72\std{5.02} & 82.01\std{6.03} \\
    \multicolumn{2}{l|}{\textbf{TimesNet}}  & 82.08\std{6.29} & 95.21\std{2.45} & 58.67\std{5.67} & 79.18\std{3.39} & 83.80\std{3.60} & 88.12\std{9.12} & 55.95\std{6.89} & 56.01\std{4.93} & 56.21\std{6.68} & 79.27\std{7.74} \\
    
    \midrule
    \multicolumn{2}{l|}{\textbf{BIOT}}  & 88.94\std{2.69} & 96.78\std{3.04} & \textcolor{mygreen}{69.79\std{5.90}} & 86.85\std{4.77} & 90.83\std{4.72} & 96.73\std{2.98} & \textcolor{myblue}{79.55\std{5.36}} & \textcolor{myred}{\textbf{92.92\std{2.77}}} & 54.79\std{9.52} & 77.95\std{11.46} \\
    \multicolumn{2}{l|}{\textbf{LaBraM}}  & 92.24\std{3.38} & \textcolor{myblue}{98.90\std{0.53}} & \textcolor{myblue}{74.37\std{4.31}} & \textcolor{myblue}{90.29\std{2.76}} & 91.12\std{4.50} & 97.54\std{1.88} & 71.65\std{3.35} & 85.61\std{1.20} & 52.37\std{7.21} & 78.27\std{9.83} \\
    \multicolumn{2}{l|}{\textbf{CBraMod}}  & \textcolor{mygreen}{92.32\std{4.27}} & 98.47\std{1.49} & 68.33\std{4.53} & \textcolor{mygreen}{86.95\std{2.89}} & 83.70\std{9.28} & 96.67\std{3.69} & 74.10\std{2.40} & 83.10\std{1.85} & 53.93\std{6.87} & 78.68\std{9.10} \\
    \multicolumn{2}{l|}{\textbf{CSBrain}}  & 91.11\std{2.34} & 98.23\std{0.51} & 69.40\std{2.63} & 86.82\std{1.28} & \textcolor{mygreen}{93.16\std{4.03}} & \textcolor{myblue}{98.87\std{1.14}} & 61.69\std{8.63} & 68.20\std{1.53} & \textcolor{myblue}{60.30\std{5.41}} & \textcolor{myred}{\textbf{83.38\std{5.73}}} \\

    \midrule
    \multicolumn{2}{l|}{\textbf{LEAD}}  & \textcolor{myred}{\textbf{97.06\std{1.02}}} & \textcolor{myred}{\textbf{99.92\std{0.07}}} & \textcolor{myred}{\textbf{81.01\std{5.02}}} & \textcolor{myred}{\textbf{94.03\std{1.33}}} & \textcolor{myred}{\textbf{97.42\std{2.74}}} & \textcolor{myred}{\textbf{100.00\std{0.00}}} & \textcolor{myred}{\textbf{83.70\std{5.30}}} & \textcolor{myblue}{92.51\std{4.33}} & \textcolor{mygreen}{59.70\std{6.64}} & 79.34\std{10.07} \\

    \midrule
    \multicolumn{2}{c}{} & \multicolumn{8}{c}{\textbf{Subject-Level Detection}}  \\
    \midrule

    \multicolumn{2}{l|}{\textbf{ManualFeature}}  & 81.05\std{20.29} & 100.00\std{0.00} & 54.26\std{7.80} & 86.15\std{11.29} & 63.05\std{30.68} & 75.56\std{22.39} & 57.33\std{19.60} & 72.50\std{22.91} & 42.90\std{8.74} & 69.86\std{8.21} \\
    \multicolumn{2}{l|}{\textbf{EEGConformer}}  & \textcolor{myblue}{100.00\std{0.00}} & \textcolor{mygreen}{100.00\std{0.00}} & 68.45\std{15.21} & 95.56\std{3.67} & \textcolor{myblue}{100.00\std{0.00}} & \textcolor{mygreen}{100.00\std{0.00}} & 65.33\std{16.00} & 100.00\std{0.00} & 58.42\std{8.60} & 81.63\std{4.71} \\
    \multicolumn{2}{l|}{\textbf{EEGInception}}  & 100.00\std{0.00} & 100.00\std{0.00} & 79.47\std{17.70} & 96.55\std{4.93} & 100.00\std{0.00} & 100.00\std{0.00} & 73.33\std{0.00} & 80.00\std{10.00} & 60.41\std{7.37} & \textcolor{myblue}{82.93\std{5.36}} \\
    \multicolumn{2}{l|}{\textbf{EEGNet}}  & 78.37\std{21.07} & 100.00\std{0.00} & 54.82\std{14.34} & 68.69\std{15.52} & 61.38\std{22.11} & 84.44\std{18.39} & 70.67\std{21.33} & 77.50\std{12.25} & 34.02\std{11.43} & 64.46\std{5.93} \\
    \multicolumn{2}{l|}{\textbf{iTransformer}}  & 77.12\std{18.02} & 100.00\std{0.00} & 67.17\std{9.15} & 90.28\std{6.75} & 100.00\std{0.00} & 100.00\std{0.00} & \textcolor{mygreen}{84.00\std{13.06}} & \textcolor{mygreen}{100.00\std{0.00}} & 52.85\std{8.58} & 78.30\std{7.84} \\
    \multicolumn{2}{l|}{\textbf{MedGNN}}  & 100.00\std{0.00} & 100.00\std{0.00} & 75.23\std{4.81} & 94.72\std{3.45} & 96.57\std{6.86} & 100.00\std{0.00} & 65.33\std{16.00} & 70.00\std{18.71} & 52.18\std{11.32} & 76.60\std{5.61} \\
    \multicolumn{2}{l|}{\textbf{Medformer}}  & 100.00\std{0.00} & 100.00\std{0.00} & 78.98\std{12.08} & 94.76\std{5.13} & 100.00\std{0.00} & 100.00\std{0.00} & 73.33\std{0.00} & 80.00\std{10.00} & 60.93\std{6.49} & 80.58\std{3.94} \\
    \multicolumn{2}{l|}{\textbf{MNet}}  & 58.82\std{17.61} & 100.00\std{0.00} & 68.63\std{17.53} & 92.38\std{7.59} & 89.90\std{13.38} & 95.56\std{5.44} & 41.33\std{16.00} & 87.50\std{15.81} & 39.97\std{4.38} & 74.12\std{8.65} \\
    \multicolumn{2}{l|}{\textbf{ModernTCN}}  & 58.82\std{17.61} & 100.00\std{0.00} & 74.39\std{11.35} & 93.77\std{4.71} & 89.71\std{8.40} & 100.00\std{0.00} & 33.33\std{0.00} & 72.50\std{5.00} & 61.96\std{9.90} & 82.28\std{5.41} \\
    \multicolumn{2}{l|}{\textbf{PatchTST}}  & 77.12\std{18.02} & 100.00\std{0.00} & 65.04\std{14.32} & 87.78\std{12.73} & 86.29\std{19.99} & 95.56\std{8.89} & 33.33\std{0.00} & 65.00\std{5.00} & 51.35\std{9.07} & 76.80\std{3.11} \\
    \multicolumn{2}{l|}{\textbf{TCN}}  & 96.08\std{7.84} & 100.00\std{0.00} & 82.80\std{6.10} & 96.43\std{1.96} & 96.57\std{6.86} & 97.78\std{4.44} & 73.33\std{0.00} & 95.00\std{10.00} & \textcolor{myblue}{62.66\std{3.41}} & \textcolor{mygreen}{82.69\std{3.36}} \\
    \multicolumn{2}{l|}{\textbf{TimesNet}}  & 84.97\std{21.64} & 100.00\std{0.00} & 64.23\std{17.73} & 94.72\std{3.70} & 96.57\std{6.86} & 100.00\std{0.00} & 33.33\std{0.00} & 70.00\std{18.71} & 55.52\std{4.93} & 77.82\std{4.56} \\
    
    \midrule
    \multicolumn{2}{l|}{\textbf{BIOT}}  & 100.00\std{0.00} & 100.00\std{0.00} & \textcolor{mygreen}{88.10\std{9.72}} & 97.10\std{2.78} & 100.00\std{0.00} & 100.00\std{0.00} & \textcolor{myblue}{94.67\std{10.67}} & \textcolor{myblue}{100.00\std{0.00}} & 55.74\std{8.58} & 78.10\std{8.27} \\
    \multicolumn{2}{l|}{\textbf{LaBraM}}  & 100.00\std{0.00} & \textcolor{myblue}{100.00\std{0.00}} & \textcolor{myblue}{91.81\std{7.56}} & \textcolor{myred}{\textbf{99.44\std{0.68}}} & 100.00\std{0.00} & 100.00\std{0.00} & 70.67\std{21.33} & 100.00\std{0.00} & 49.83\std{6.46} & 78.06\std{4.39} \\
    \multicolumn{2}{l|}{\textbf{CBraMod}}  & \textcolor{mygreen}{100.00\std{0.00}} & 100.00\std{0.00} & 82.21\std{6.30} & 97.30\std{2.05} & 89.07\std{14.85} & 100.00\std{0.00} & 73.33\std{0.00} & 75.00\std{15.81} & 50.60\std{10.89} & 75.14\std{4.75} \\
    \multicolumn{2}{l|}{\textbf{CSBrain}}  & 100.00\std{0.00} & 100.00\std{0.00} & 78.33\std{7.52} & \textcolor{mygreen}{97.74\std{2.65}} & \textcolor{mygreen}{100.00\std{0.00}} & \textcolor{myblue}{100.00\std{0.00}} & 57.33\std{19.60} & 50.00\std{0.00} & \textcolor{mygreen}{62.41\std{5.80}} & \textcolor{myred}{\textbf{83.50\std{5.90}}} \\

    \midrule
    \multicolumn{2}{l|}{\textbf{LEAD}}  & \textcolor{myred}{\textbf{100.00\std{0.00}}} & \textcolor{myred}{\textbf{100.00\std{0.00}}} & \textcolor{myred}{\textbf{93.95\std{4.95}}} & \textcolor{myblue}{99.17\std{1.11}} & \textcolor{myred}{\textbf{100.00\std{0.00}}} & \textcolor{myred}{\textbf{100.00\std{0.00}}} & \textcolor{myred}{\textbf{100.00\std{0.00}}} & \textcolor{myred}{\textbf{100.00\std{0.00}}} & \textcolor{myred}{\textbf{65.67\std{6.59}}} & 80.37\std{7.21} \\

    \bottomrule
    \end{tabular}
    }
\vspace{-3mm}
\end{table*}

\begin{figure*}[t]
    \centering
    \includegraphics[width=1.0\linewidth]{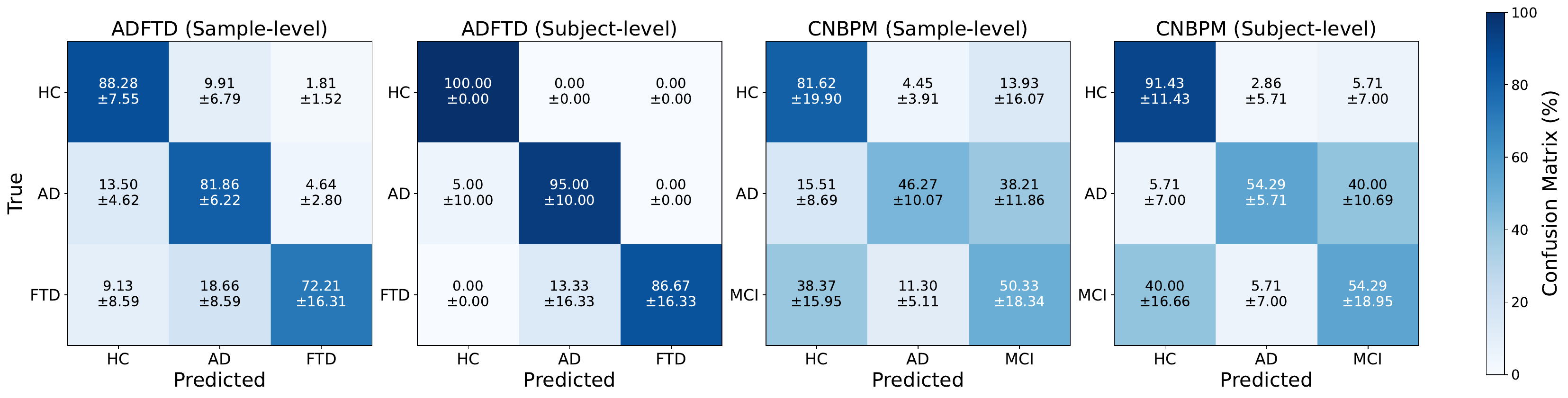}
    \caption{ The confusion matrix for multi-class classification results on ADFTD and CNBPM with \myname.
    }
    \label{fig:confusion_matrix}
    \vspace{-3mm}
\end{figure*}

\subsection{Results}

\textbf{Performance Comparison with Baselines.} 
\label{para:method_comparison}
We compare \myname with 16 baseline methods on 5 downstream datasets. The F1 score and AUROC results are reported in Table~\ref{tab:method_comparison_detail}, while the full results for all 7 evaluation metrics are provided in Appendix~\ref{sub:method_comparison_detail}. The confusion matrix of \myname for two multi-class classification datasets, ADFTD and CNBPM, is provided in Figure~\ref {fig:confusion_matrix}. Our method achieves top-1 performance in 15 out of 20 evaluations. Notably, \myname achieves a top-1 performance ranking among 17 methods across all 20 evaluations, as shown in figure~\ref{fig:avg_rank_bar_heatmap}. Furthermore, our method surpasses existing EEG foundation models despite being pre-trained on only 1,185.84 hours of EEG data, which is substantially less than the approximately 2,500 hours used by LaBraM and more than 9,000 hours used by CBraMod and CSBrain. These results highlight the effectiveness of the proposed backbone, subject-level training strategy, and domain-relevant pre-training method, as well as the careful selection of pre-training data.

\textbf{Subject-Level Performance Uncertainty.}
\label{para:subject_level_uncertainty}
First, the results in Table~\ref{tab:method_comparison_detail} confirm the phenomenon observed in~\cite{wang2024adformer} that strong sample-level classification performance does not necessarily translate into strong subject-level detection performance. For example, on CNBPM, Medformer achieves the highest sample-level F1 score of 60.39\%, but only 60.93\% at the subject level. In contrast, \myname achieves a slightly lower sample-level F1 score of 59.70\% but the highest subject-level F1 score of 65.67\%. We further provide two illustrative examples in Appendix~\ref{sec:subject_performance_example} to show how different distributions of sample-level predictions across subjects can lead to substantially different subject-level outcomes. Second, although \myname achieves 100\% subject-level F1 scores on ADFSU, ADSZ, and APAVA, these results should not be interpreted as evidence that EEG-based AD detection is already solved in real-world clinical applications. These datasets contain either relatively few samples per subject (ADFSU and ADSZ) or few subjects (APAVA), introducing considerable uncertainty in subject-level estimates. In addition, according to the original publications, the EEG recordings in these three datasets were collected under relatively controlled conditions, with high-quality recordings selected and recordings with severe artifacts manually rejected by EEG experts. The number of samples per subject in ADFSU and ADSZ is also relatively balanced, which may further stabilize majority voting, a condition that is less likely to hold in real-world settings. Therefore, these near-saturated results reflect strong performance under controlled in-lab settings rather than universally reliable clinical performance.

\begin{table*}[!t]
    \centering
    \def\arraystretch{1.0}
    \caption{
    \textbf{Pre-training Effectiveness.} Comparison of fully supervised learning, linear probing, and fine-tuning of \myname to evaluate the effectiveness of self-supervised pre-training for AD detection.
    }
    \vspace{-2mm}
    \label{tab:pretrain_effectiveness}
    \resizebox{\textwidth}{!}{
    \begin{tabular}{@{}ll|cc|cc|cc|cc|cc@{}}
    \toprule

    \multicolumn{2}{l|}{\textbf{Datasets}}
    & \multicolumn{2}{c|}{\makecell{\textbf{ADFSU} \\ \textit{(4,048 Samples)} \\ \textit{(92 Subjects)} \\ \textit{(HC vs AD)}}}
    & \multicolumn{2}{c|}{\makecell{\textbf{ADFTD} \\ \textit{(167,083 Samples)} \\ \textit{(88 Subjects)} \\ \textit{(HC vs AD vs FTD)}}}
    & \multicolumn{2}{c|}{\makecell{\textbf{ADSZ} \\ \textit{(1,128 Samples)} \\ \textit{(48 Subjects)} \\ \textit{(HC vs AD)}}}
    & \multicolumn{2}{c|}{\makecell{\textbf{APAVA} \\ \textit{(9,282 Samples)} \\ \textit{(23 Subjects)} \\ \textit{(HC vs AD)}}}
    & \multicolumn{2}{c}{\makecell{\textbf{CNBPM}  \\ \textit{(122,029 Samples)} \\ \textit{(189 Subjects)} \\ \textit{(HC vs MCI vs AD)}}}
    \\ \midrule

    \multicolumn{2}{l|}{\diagbox{\textbf{Methods}}{\textbf{Metrics}}} & \textbf{F1 Score} & \textbf{AUROC} & \textbf{F1 Score} & \textbf{AUROC}  & \textbf{F1 Score} & \textbf{AUROC}  & \textbf{F1 Score} & \textbf{AUROC}  & \textbf{F1 Score} & \textbf{AUROC} \\ 
    
    \midrule

    \multicolumn{2}{c}{} & \multicolumn{8}{c}{\textbf{Sample-Level Classification}}  \\
    \midrule

    \multicolumn{2}{l|}{\textbf{Fully Supervised}}  & 88.43\std{0.97} & 97.35\std{0.70} & 73.47\std{5.75} & 89.26\std{3.34} & 87.15\std{3.87} & 94.05\std{3.32} & 56.99\std{1.34} & 71.37\std{3.67} & 54.30\std{5.36} & 77.52\std{7.87} \\
    \multicolumn{2}{l|}{\textbf{Linear Probing}}  & 95.31\std{1.65} & \textbf{100.00\std{0.00}} & 69.43\std{5.06} & 85.67\std{2.99} & 93.09\std{5.66} & 99.57\std{0.46} & 80.70\std{0.24} & 88.45\std{1.15} & 55.76\std{6.93} & 78.98\std{8.80} \\
    \multicolumn{2}{l|}{\textbf{Fine-tuning (LEAD)}}  & \textbf{97.06\std{1.02}} & 99.92\std{0.07} & \textbf{81.01\std{5.02}} & \textbf{94.03\std{1.33}} & \textbf{97.42\std{2.74}} & \textbf{100.00\std{0.00}} & \textbf{83.70\std{5.30}} & \textbf{92.51\std{4.33}} & \textbf{59.70\std{6.64}} & \textbf{79.34\std{10.07}} \\

    \midrule
    \multicolumn{2}{c}{} & \multicolumn{8}{c}{\textbf{Subject-Level Detection}}  \\
    \midrule

    \multicolumn{2}{l|}{\textbf{Fully Supervised}}  & 100.00\std{0.00} & 100.00\std{0.00} & 81.13\std{14.83} & 98.21\std{2.38} & 100.00\std{0.00} & 100.00\std{0.00} & 36.67\std{6.67} & 75.00\std{0.00} & 60.36\std{4.05} & \textbf{80.92\std{5.43}} \\
    \multicolumn{2}{l|}{\textbf{Linear Probing}}  & 100.00\std{0.00} & 100.00\std{0.00} & 83.09\std{6.02} & 96.15\std{3.34} & 96.57\std{6.86} & 100.00\std{0.00} & 73.33\std{0.00} & 95.00\std{10.00} & 54.95\std{3.61} & 75.54\std{4.68} \\
    \multicolumn{2}{l|}{\textbf{Fine-tuning (LEAD)}}  & \textbf{100.00\std{0.00}} & \textbf{100.00\std{0.00}} & \textbf{93.95\std{4.95}} & \textbf{99.17\std{1.11}} & \textbf{100.00\std{0.00}} & \textbf{100.00\std{0.00}} & \textbf{100.00\std{0.00}} & \textbf{100.00\std{0.00}} & \textbf{65.67\std{6.59}} & 80.37\std{7.21} \\

    \bottomrule
    \end{tabular}
    }
\vspace{-4mm}
\end{table*}

\textbf{Pre-training Effectiveness.} 
\label{para:pretrain_effectiveness}
To demonstrate the effectiveness of self-supervised pre-training and evaluate the representation learning ability of \myname as a foundation model, we compare three variants: fully supervised learning from scratch, linear probing, and fine-tuning. The fully supervised setting retains all model components and training designs, such as multi-scale segmentation and subject-level cross-entropy loss, but does not use pre-training. Linear probing freezes the pre-trained backbone encoder and trains only a classifier for downstream classification. Fine-tuning corresponds to the full version of \myname used in this paper, where the pre-trained backbone is further optimized on each downstream dataset.

The results are presented in Table~\ref{tab:pretrain_effectiveness}. Overall, fine-tuning achieves the best performance in most cases, demonstrating the benefit of adapting the pre-trained representations to downstream AD detection tasks. Interestingly, on relatively large datasets such as ADFTD and CNBPM, linear probing does not consistently outperform fully supervised learning and can even perform worse in some cases. In contrast, on smaller datasets with limited numbers of subjects, such as APAVA with only 23 subjects, linear probing substantially outperforms the less stable fully supervised training. These results suggest that self-supervised pre-training is particularly beneficial when downstream training data are limited, a common situation in EEG research and especially in EEG-based AD detection.

\begin{table*}[!h]
    \centering
    \def\arraystretch{1.0}
    \caption{\textbf{EEG Paradigms Comparison.} Each ADFTD subject was recorded under two paradigms: resting-state eyes-closed and photic-stimulation eyes-open, with varying recording lengths. We evaluate model performance on each paradigm separately and on both.}
    \vspace{-2mm}
    \label{tab:eeg_paradigm_comparison}
    \resizebox{\textwidth}{!}{%
    \begin{tabular}{clccccccc}

    \toprule
    \textbf{Datasets} & \textbf{Paradigms} & \multicolumn{1}{c}{\textbf{Accuracy}} & \multicolumn{1}{c}{\textbf{Precision}} & \multicolumn{1}{c}{\textbf{Sensitivity}} & \multicolumn{1}{c}{\textbf{Specificity}} & \multicolumn{1}{c}{\textbf{F1 Score}} & \multicolumn{1}{c}{\textbf{AUROC}} & \multicolumn{1}{c}{\textbf{AUPRC}} \\
    \midrule
    \multirow{3}{*}{\begin{tabular}[c]{@{}l@{}}\;\makecell{\makecell{\textbf{ADFTD} \\ \textit{(Sample-Level)} \\ \textit{(HC vs AD vs FTD)}}} \end{tabular}} 
    & \textbf{Resting-State (19.40h)}  & 53.09\std{10.04} & 50.42\std{13.91} & 49.30\std{10.77} & 75.15\std{5.49} & 47.54\std{11.79} & 70.91\std{11.12} & 58.18\std{13.59} \\
    & \textbf{Photic-Stimulation (7.25h)}  & 58.23\std{9.57} & 58.13\std{12.16} & 56.24\std{10.83} & 78.83\std{4.93} & 55.60\std{11.70} & 72.08\std{11.15} & 60.17\std{13.52} \\
    & \textbf{Both (26.64h)}  & \textbf{81.70\std{4.74}} & \textbf{82.93\std{4.40}} & \textbf{80.78\std{4.95}} & \textbf{90.52\std{2.45}} & \textbf{81.01\std{5.02}} & \textbf{94.03\std{1.33}} & \textbf{90.37\std{2.14}} \\

    \midrule
    \multirow{3}{*}{\begin{tabular}[c]{@{}l@{}}\;\makecell{\makecell{\textbf{ADFTD} \\ \textit{(Subject-Level)} \\ \textit{(HC vs AD vs FTD)}}} \end{tabular}} 
    & \textbf{Resting-State (19.40h)}  & 60.00\std{20.98} & 52.87\std{29.78} & 57.22\std{21.77} & 78.89\std{11.06} & 52.27\std{24.98} & 76.39\std{15.92} & 74.34\std{17.42} \\
    & \textbf{Photic-Stimulation (7.25h)}  & 60.00\std{10.95} & 61.67\std{18.05} & 60.56\std{11.84} & 79.68\std{5.80} & 58.58\std{14.37} & 75.52\std{16.44} & 73.75\std{15.88} \\
    & \textbf{Both (26.64h)}  & \textbf{94.00\std{4.90}} & \textbf{95.67\std{3.59}} & \textbf{93.89\std{5.09}} & \textbf{96.83\std{2.61}} & \textbf{93.95\std{4.95}} & \textbf{99.17\std{1.11}} & \textbf{99.11\std{1.14}} \\

\bottomrule
\end{tabular}
}
\vspace{-3mm}
\end{table*}

\textbf{EEG Paradigms Comparison.} 
\label{para:paradigm_comparison}
We evaluate the ADFTD dataset across three paradigms: resting-state, photic-stimulation, and both. Results are shown in Table~\ref{tab:eeg_paradigm_comparison}. Using both paradigms substantially outperforms the individual paradigms, achieving a subject-level F1 of 93.95\% versus 52.27\% (resting-state) and 58.58\% (photic-stimulation). While it remains unclear whether this gain arises from longer recordings or complementary paradigm features, the evidence indicates that \textit{combining resting-state and photic-stimulation EEG paradigms} is non-detrimental and likely to be beneficial for AD detection. As shown in~\citep{wang2024causes}, EEG tasks such as AD detection are single-class-per-subject tasks, whose cross-subject performance can be strongly affected by shortcut learning based on subject-specific features. One possible explanation for the substantial performance improvement when combining the two paradigms is that resting-state and photic-stimulation EEG exhibit considerably different signal patterns. As a result, samples from the same subject can differ substantially across paradigms, potentially making it harder for the model to rely on subject-specific shortcuts and encouraging it to learn more disease-relevant features. However, this remains a hypothesis and requires further investigation.

\begin{figure*}[h]
    \centering
    \includegraphics[width=1.0\linewidth]{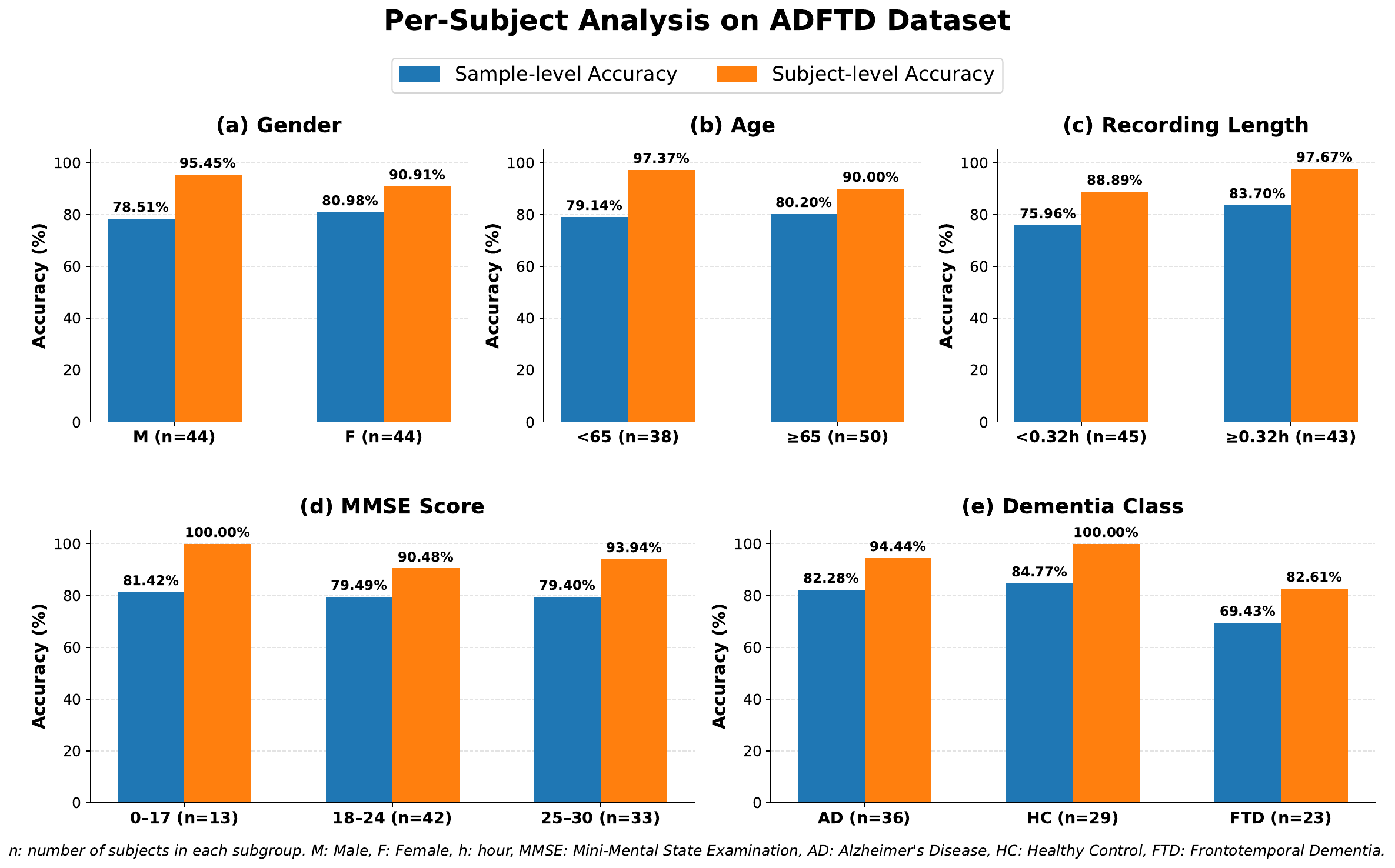}
    \caption{\textbf{Grouped Bar Plot of Per-Subject Analysis.} Sample-level and subject-level accuracy are reported for subgroups defined by gender, age, recording length, MMSE score, and dementia class.
    }
    \label{fig:per_subject_analysis}
    \vspace{-3mm}
\end{figure*}

\begin{table*}[!h]
    \centering
    \scriptsize
    \caption{\textbf{Per-Subject Analysis.} We report per-subject results on the ADFTD dataset under the LOSO setting and analyze the factors influencing model performance, including demographic characteristics, recording length, MMSE score, and dementia category.}
    \vspace{-2mm}
    \label{tab:per_subject_analysis}
    \resizebox{\textwidth}{!}{%
    \begin{tabular}{ll|cc|cc|cc|ccc|ccc}
    \toprule
    \multicolumn{2}{l|}{\textbf{Label}} & \multicolumn{2}{c|}{\textbf{Gender}} & \multicolumn{2}{c|}{\textbf{Age}} & \multicolumn{2}{c|}{\textbf{Length}} & \multicolumn{3}{c|}{\textbf{MMSE}}  & \multicolumn{3}{c}{\textbf{Dementia Class}} \\
    \cmidrule(lr){1-14}
    \multicolumn{2}{l|}{\textbf{Group}} & \multicolumn{1}{c|}{\textbf{M}} & \multicolumn{1}{c|}{\textbf{F}} & \multicolumn{1}{c|}{\textbf{$<$65}} & \multicolumn{1}{c|}{\textbf{$\geq$65}} & \multicolumn{1}{c|}{\textbf{<0.32h}} & \multicolumn{1}{c|}{\textbf{>=0.32h}} & \multicolumn{1}{c|}{\textbf{0-17}} & \multicolumn{1}{c|}{\textbf{18-24}} & \multicolumn{1}{c|}{\textbf{25-30}} & \multicolumn{1}{c|}{\textbf{AD}} & \multicolumn{1}{c|}{\textbf{HC}} & \multicolumn{1}{c}{\textbf{FTD}} \\
    \cmidrule(lr){1-14}
    \multicolumn{2}{l|}{\textbf{Subject Number}} & 44 & 44 & 38 & 50 & 45 & 43 & 13 & 42 & 33 & 36 & 29 & 23 \\
    \multicolumn{2}{l|}{\textbf{Sample-level Accuracy}} & 78.51\% & \textbf{80.98\%} & 79.14\% & \textbf{80.20\%} & 75.96\% & \textbf{83.70\%} & \textbf{81.42\%} & 79.49\% & 79.40\% & 82.28\% & \textbf{84.77\%} & 69.43\% \\
    \multicolumn{2}{l|}{\textbf{Subject-level Accuracy}} & \textbf{95.45\%} & 90.91\% & \textbf{97.37\%} & 90.00\% & 88.89\% & \textbf{97.67\%} & \textbf{100.00\%} & 90.48\% & 93.94\% & 94.44\% & \textbf{100.00\%} & 82.61\% \\
    \bottomrule
    \end{tabular}%
    }
\vspace{-3mm}
\end{table*}

\textbf{Per-Subject Analysis.}
\label{para:per_subject_analysis}
As ADFTD is a public dataset with detailed demographic information, we further explore per-subject performance across demographics, recording length, MMSE score, and clinical subgroups under the leave-one-subject-out (LOSO) setting (See Appendix ~\ref{sec:subject_independent}). The full results of each subject are in Appendix~\ref{sub:per_subject_analysis}. A comparison among factors is in Table~\ref{tab:per_subject_analysis} and plotted in Figure~\ref{fig:per_subject_analysis}. According to the results, longer recordings (>=0.32 h) consistently yield higher accuracy at both the sample and subject levels, highlighting the importance of sufficient EEG recording length. Subjects younger than 65 years show higher subject-level accuracy than older subjects, suggesting increased variability in elderly EEG recordings. Performance also varies with cognitive status: subjects with lower MMSE scores (0-17) are classified most reliably, whereas intermediate MMSE ranges (18-24) exhibit reduced accuracy. In terms of dementia category, the model achieves perfect subject-level accuracy in HC, strong performance in AD, and noticeably lower accuracy in FTD, indicating that FTD remains the most challenging to distinguish.

\begin{table*}[!t]
    \centering
    \def\arraystretch{1.0}
    \caption{
    \textbf{Module Ablation Study.} We investigate the effectiveness of individual modules and design choices through systematic ablation, removing each component separately.
    }
    \vspace{-2mm}
    \label{tab:module_ablation_study}
    \resizebox{\textwidth}{!}{
    \begin{tabular}{@{}ll|cc|cc|cc|cc|cc@{}}
    \toprule

    \multicolumn{2}{l|}{\textbf{Datasets}}
    & \multicolumn{2}{c|}{\makecell{\textbf{ADFSU} \\ \textit{(4,048 Samples)} \\ \textit{(92 Subjects)} \\ \textit{(HC vs AD)}}}
    & \multicolumn{2}{c|}{\makecell{\textbf{ADFTD} \\ \textit{(167,083 Samples)} \\ \textit{(88 Subjects)} \\ \textit{(HC vs AD vs FTD)}}}
    & \multicolumn{2}{c|}{\makecell{\textbf{ADSZ} \\ \textit{(1,128 Samples)} \\ \textit{(48 Subjects)} \\ \textit{(HC vs AD)}}}
    & \multicolumn{2}{c|}{\makecell{\textbf{APAVA} \\ \textit{(9,282 Samples)} \\ \textit{(23 Subjects)} \\ \textit{(HC vs AD)}}}
    & \multicolumn{2}{c}{\makecell{\textbf{CNBPM}  \\ \textit{(122,029 Samples)} \\ \textit{(189 Subjects)} \\ \textit{(HC vs MCI vs AD)}}}
    \\ \midrule

    \multicolumn{2}{l|}{\diagbox{\textbf{Methods}}{\textbf{Metrics}}} & \textbf{F1 Score} & \textbf{AUROC} & \textbf{F1 Score} & \textbf{AUROC}  & \textbf{F1 Score} & \textbf{AUROC}  & \textbf{F1 Score} & \textbf{AUROC}  & \textbf{F1 Score} & \textbf{AUROC} \\ 

    \midrule

    \multicolumn{2}{c}{} & \multicolumn{8}{c}{\textbf{Sample-Level Classification}}  \\
    \midrule

    \multicolumn{2}{l|}{\textbf{No Index Group Shuffling}}  & 97.16\std{0.51} & 99.88\std{0.07} & 78.73\std{5.80} & 92.36\std{2.50} & \textbf{98.97\std{0.76}} & 99.96\std{0.06} & 79.80\std{2.22} & 91.28\std{2.66} & 55.87\std{6.55} & 77.59\std{8.20} \\
    \multicolumn{2}{l|}{\textbf{No Subject Cross-Entropy Loss}}  & \textbf{97.35\std{1.29}} & 99.94\std{0.05} & 79.76\std{5.11} & 93.44\std{1.53} & 98.29\std{1.99} & 99.95\std{0.08} & 84.67\std{4.18} & 94.54\std{1.56} & 59.01\std{6.92} & \textbf{79.84\std{11.04}} \\
    \multicolumn{2}{l|}{\textbf{No Sampling Rate Embedding}}  & 96.74\std{3.09} & \textbf{99.95\std{0.05}} & 79.83\std{5.58} & 92.56\std{2.93} & 98.66\std{2.06} & 100.00\std{0.00} & 76.37\std{3.66} & 87.18\std{3.01} & 55.90\std{9.12} & 79.15\std{8.63} \\
    \multicolumn{2}{l|}{\textbf{No Multi-Sampling Segmentation}}  & 97.02\std{1.77} & 99.84\std{0.18} & 78.59\std{4.35} & 92.57\std{1.49} & 98.11\std{2.48} & 99.82\std{0.29} & \textbf{86.75\std{2.63}} & \textbf{95.88\std{1.33}} & 57.78\std{7.71} & 78.44\std{11.69} \\
    \multicolumn{2}{l|}{\textbf{LEAD (Full)}}  & 97.06\std{1.02} & 99.92\std{0.07} & \textbf{81.01\std{5.02}} & \textbf{94.03\std{1.33}} & 97.42\std{2.74} & \textbf{100.00\std{0.00}} & 83.70\std{5.30} & 92.51\std{4.33} & \textbf{59.70\std{6.64}} & 79.34\std{10.07} \\

    \midrule
    \multicolumn{2}{c}{} & \multicolumn{8}{c}{\textbf{Subject-Level Detection}}  \\
    \midrule

    \multicolumn{2}{l|}{\textbf{No Index Group Shuffling}}  & 100.00\std{0.00} & 100.00\std{0.00} & 89.31\std{9.70} & 99.44\std{0.68} & 100.00\std{0.00} & 100.00\std{0.00} & 73.33\std{0.00} & 100.00\std{0.00} & 64.88\std{7.15} & 79.46\std{4.72} \\
    \multicolumn{2}{l|}{\textbf{No Subject Cross-Entropy Loss}}  & 100.00\std{0.00} & 100.00\std{0.00} & 87.40\std{8.24} & 98.85\std{1.69} & 100.00\std{0.00} & 100.00\std{0.00} & 94.67\std{10.67} & 100.00\std{0.00} & 58.63\std{10.71} & 80.07\std{5.82} \\
    \multicolumn{2}{l|}{\textbf{No Sampling Rate Embedding}}  & 100.00\std{0.00} & 100.00\std{0.00} & 88.76\std{10.47} & \textbf{100.00\std{0.00}} & 100.00\std{0.00} & 100.00\std{0.00} & 73.33\std{0.00} & 80.00\std{10.00} & 59.29\std{10.90} & 78.40\std{3.92} \\
    \multicolumn{2}{l|}{\textbf{No Multi-Sampling Segmentation}}  & 100.00\std{0.00} & 100.00\std{0.00} & 89.31\std{9.70} & 99.13\std{1.75} & 100.00\std{0.00} & 100.00\std{0.00} & 84.00\std{13.06} & 100.00\std{0.00} & 63.91\std{8.63} & 79.25\std{7.50} \\
    \multicolumn{2}{l|}{\textbf{LEAD (Full)}}  & \textbf{100.00\std{0.00}} & \textbf{100.00\std{0.00}} & \textbf{93.95\std{4.95}} & 99.17\std{1.11} & \textbf{100.00\std{0.00}} & \textbf{100.00\std{0.00}} & \textbf{100.00\std{0.00}} & \textbf{100.00\std{0.00}} & \textbf{65.67\std{6.59}} & \textbf{80.37\std{7.21}} \\

    \bottomrule
    \end{tabular}
    }
\vspace{-4mm}
\end{table*}

\textbf{Module Ablation Study.}
\label{para:module_ablation_study}
We conduct an ablation study on the 5 downstream datasets to evaluate the contribution of each module in our proposed method. The results are reported in Table~\ref{tab:module_ablation_study}. Overall, the contribution of each component varies across datasets. On the relatively balanced ADFSU and ADSZ datasets, most variants achieve near-saturated performance, particularly at the subject level, making the effect of individual modules less pronounced. In contrast, the three imbalanced datasets, ADFTD, APAVA, and CNBPM, show greater sensitivity to different design choices. In particular, removing the subject-level cross-entropy loss consistently reduces subject-level F1 on these datasets, from 93.95 to 87.40 on ADFTD, from 100.00 to 94.67 on APAVA, and from 65.67 to 58.63 on CNBPM, demonstrating its importance for aligning sample-level learning with subject-level detection. Removing index group shuffling, sampling-rate embedding, or multi-sampling segmentation also degrades performance on several datasets, although occasional improvements are observed for individual metrics. These results suggest that the proposed components provide complementary benefits rather than uniformly improving every dataset and metric.

\begin{table*}[!t]
    \centering
    \def\arraystretch{1.0}
    \caption{
    \textbf{Channel Embedding Ablation Study.} We compare the 3D channel embedding to fixed and learnable channel embeddings.
    }
    \vspace{-2mm}
    \label{tab:channel_ablation_study}
    \resizebox{\textwidth}{!}{
    \begin{tabular}{@{}ll|cc|cc|cc|cc|cc@{}}
    \toprule

    \multicolumn{2}{l|}{\textbf{Datasets}}
    & \multicolumn{2}{c|}{\makecell{\textbf{ADFSU} \\ \textit{(4,048 Samples)} \\ \textit{(92 Subjects)} \\ \textit{(HC vs AD)}}}
    & \multicolumn{2}{c|}{\makecell{\textbf{ADFTD} \\ \textit{(167,083 Samples)} \\ \textit{(88 Subjects)} \\ \textit{(HC vs AD vs FTD)}}}
    & \multicolumn{2}{c|}{\makecell{\textbf{ADSZ} \\ \textit{(1,128 Samples)} \\ \textit{(48 Subjects)} \\ \textit{(HC vs AD)}}}
    & \multicolumn{2}{c|}{\makecell{\textbf{APAVA} \\ \textit{(9,282 Samples)} \\ \textit{(23 Subjects)} \\ \textit{(HC vs AD)}}}
    & \multicolumn{2}{c}{\makecell{\textbf{CNBPM}  \\ \textit{(122,029 Samples)} \\ \textit{(189 Subjects)} \\ \textit{(HC vs MCI vs AD)}}}
    \\ \midrule

    \multicolumn{2}{l|}{\diagbox{\textbf{Methods}}{\textbf{Metrics}}} & \textbf{F1 Score} & \textbf{AUROC} & \textbf{F1 Score} & \textbf{AUROC}  & \textbf{F1 Score} & \textbf{AUROC}  & \textbf{F1 Score} & \textbf{AUROC}  & \textbf{F1 Score} & \textbf{AUROC} \\ 
    
    \midrule

    \multicolumn{2}{c}{} & \multicolumn{8}{c}{\textbf{Sample-Level Classification}}  \\
    \midrule

    \multicolumn{2}{l|}{\textbf{Fixed Embedding}}  & \textbf{97.76\std{0.66}} & \textbf{99.94\std{0.03}} & 78.32\std{5.20} & 91.80\std{2.73} & 97.41\std{2.43} & 99.97\std{0.06} & 83.40\std{3.58} & \textbf{93.51\std{2.34}} & 55.94\std{8.80} & 77.30\std{10.79} \\
    \multicolumn{2}{l|}{\textbf{Learnable Embedding}}  & 97.71\std{0.33} & \textbf{99.94\std{0.03}} & 78.37\std{5.56} & 92.62\std{2.31} & 97.35\std{3.08} & 99.93\std{0.13} & 78.24\std{2.81} & 89.12\std{2.00} & 55.78\std{9.75} & 76.48\std{12.46} \\
    \multicolumn{2}{l|}{\textbf{LEAD (3D)}}  & 97.06\std{1.02} & 99.92\std{0.07} & \textbf{81.01\std{5.02}} & \textbf{94.03\std{1.33}} & \textbf{97.42\std{2.74}} & \textbf{100.00\std{0.00}} & \textbf{83.70\std{5.30}} & 92.51\std{4.33} & \textbf{59.70\std{6.64}} & \textbf{79.34\std{10.07}} \\

    \midrule
    \multicolumn{2}{c}{} & \multicolumn{8}{c}{\textbf{Subject-Level Detection}}  \\
    \midrule

    \multicolumn{2}{l|}{\textbf{Fixed Embedding}}  & 100.00\std{0.00} & 100.00\std{0.00} & 84.64\std{9.03} & 98.85\std{1.69} & 100.00\std{0.00} & 100.00\std{0.00} & 89.33\std{13.06} & 100.00\std{0.00} & 57.88\std{7.50} & 77.38\std{5.48} \\
    \multicolumn{2}{l|}{\textbf{Learnable Embedding}}  & 100.00\std{0.00} & 100.00\std{0.00} & 88.01\std{7.28} & 98.25\std{2.31} & 100.00\std{0.00} & 100.00\std{0.00} & 89.33\std{13.06} & 95.00\std{10.00} & 59.27\std{9.55} & 78.27\std{7.88} \\
    \multicolumn{2}{l|}{\textbf{LEAD (3D)}}  & \textbf{100.00\std{0.00}} & \textbf{100.00\std{0.00}} & \textbf{93.95\std{4.95}} & \textbf{99.17\std{1.11}} & \textbf{100.00\std{0.00}} & \textbf{100.00\std{0.00}} & \textbf{100.00\std{0.00}} & \textbf{100.00\std{0.00}} & \textbf{65.67\std{6.59}} & \textbf{80.37\std{7.21}} \\

    \bottomrule
    \end{tabular}
    }
\vspace{-4mm}
\end{table*}

\textbf{Channel Embedding Ablation Study.}
\label{para:channel_embedding_ablation_study}
We conduct an ablation study on the 5 downstream datasets to evaluate the effectiveness of the 3D channel embedding compared with fixed and learnable channel embeddings. All other pre-training and fine-tuning settings are kept identical across the three variants. The results are shown in Table~\ref{tab:channel_ablation_study}. Overall, the 3D channel embedding achieves the best performance on most datasets and metrics, with particularly clear advantages on ADFTD and CNBPM. For example, on ADFTD, the 3D embedding improves sample-level F1 from 78.32/78.37 to 81.01 and subject-level F1 from 84.64/88.01 to 93.95 compared with fixed/learnable embeddings. On CNBPM, it also achieves the highest sample-level and subject-level F1 and AUROC. However, the fixed embedding performs slightly better on several metrics, such as sample-level F1 on ADFSU and sample-level AUROC on APAVA, indicating that the benefit of explicit spatial information is dataset-dependent. Overall, these results suggest that incorporating electrode coordinates through the 3D channel embedding provides a more effective spatial representation across heterogeneous downstream datasets.

\section{Conclusion and Limitations}
\label{sec:conclusion}

\textbf{Conclusion.} This paper introduces \myname, the first foundation model for EEG-based Alzheimer’s disease detection, trained on the world’s largest EEG AD corpus. We design a gated temporal-spatial Transformer that adapts to EEG recordings with diverse lengths, channel configurations, and commonly used sampling rates. We further propose a subject-regularized training strategy, including subject-level cross-entropy, index-group shuffling, and multi-scale segmentation, to enhance subject-level feature learning. Medical contrastive pre-training is conducted on 13 datasets, including 4 AD datasets and 9 non-AD neurological disorder datasets. Extensive experiments on 5 heterogeneous AD downstream datasets demonstrate that \myname achieves the best overall ranking compared with 16 baseline methods across 20 evaluations. These results validate the effectiveness of our proposed backbone transformer, training strategies, proper pre-training corpus, and pre-training methods. We hope this work will advance research on EEG-based AD detection and support researchers who face challenges training models from scratch due to the limited number of AD subjects, as collecting EEG data from AD patients is costly and time-consuming.

\textbf{Limitations.} Although the performance of classifying AD vs HC and AD vs other dementias, such as FTD, is relatively strong, the performance for dementia stage classification that separates AD, MCI, and HC remains limited, with an F1 score of approximately 66\%. This highlights the continued challenge of AD stage detection. AD is a progressive neurodegenerative disease. Healthy older adults may transition into MCI, and some MCI cases eventually progress to AD. However, not all MCI cases advance; some remain stable (SMCI) or even revert to normal cognition, while others progress (PMCI)~\citep{vecchio2018sustainable,ge2025eeg}. Critically, many PMCI patients may already exhibit AD-related features at the time of data acquisition, although these are not identifiable without longitudinal follow-up, while some SMCI patients are closer to HC. This ambiguity makes certain MCI subjects difficult to distinguish from either AD or HC. Future work should therefore prioritize the collection of longitudinal EEG data to track disease progression.

\subsubsection*{Acknowledgments}
We sincerely thank the researchers and clinical teams who collected, curated, and generously made the EEG datasets used in this study publicly available. In particular, we are grateful to the researchers and clinical teams who collected and shared the AD datasets, including Drs. Hamid Aghajan and Zahra Vahabi (AD-Auditory), Agustina Legaz and Agustín Ibañez (BrainLat), Felix Benninger (P-ADIC), Joonki Paik and Young Chul Youn (CAUEEG), Dennis Duke and his collaborators at Florida State University (ADFSU and ADSZ), Alexandros T. Tzallas and his clinical collaborators (ADFTD), and Pedro Espino and Javier Escudero (APAVA). We also thank the researchers and clinical teams who collected and shared the non-AD datasets, including Drs. Stephan Getzmann (BACA-RS), James F. Cavanagh and John J. B. Allen (Depression), Dean F. Salisbury (FEPCR), Quentin Chenot (MCEF-RS), Arun Singh and Nandakumar S. Narayanan (PD-RS), Ewa Kublik and Patrycja Dzianok (PEARL-Neuro), Christoffer Hatlestad-Hall (SRM-RS), Hanneke van Dijk and Martijn Arns (TDBrain), and Iyad Obeid and Joseph Picone (TUEP). We further extend our gratitude to all other contributing researchers, clinical staff, and participants. Their efforts in data acquisition and open sharing have been invaluable in advancing EEG-AD research and have made this study possible.

\clearpage
\newpage

\bibliography{refs}
\bibliographystyle{tmlr}


\clearpage

\appendix
\setcounter{page}{1}
\begin{appendices}

\section{Subject-Level Performance Examples}
\label{sec:subject_performance_example}

In EEG-based AD and other neurodegenerative disease detection tasks, high sample-level prediction performance does not necessarily translate into strong subject-level disease detection performance after post-hoc majority voting. Although this phenomenon may appear counter-intuitive, it can occur because sample-level metrics evaluate each EEG sample independently, whereas subject-level performance depends on how prediction errors are distributed across subjects.

\textbf{Example 1.}
Consider 10 subjects, each containing 100 EEG samples. Suppose a model correctly classifies all 100 samples for eight subjects, but only 49 out of 100 samples for each of the remaining two subjects. The resulting performances are:
\begin{equation}
    \text{Sample-level Accuracy}=\frac{8\times100+2\times49}{1000}=89.8\%,
    \qquad
    \text{Subject-level Accuracy}=\frac{8}{10}=80\%.
\end{equation}
Although nearly 90\% of all EEG samples are correctly classified, the two difficult subjects are misclassified after majority voting because more than half of their samples receive incorrect predictions.

\textbf{Example 2.}
Now consider another model that correctly classifies 80 out of 100 samples for every subject. Its sample-level accuracy is lower than that in Example~1, but every subject is correctly classified after majority voting:
\begin{equation}
    \text{Sample-level Accuracy}=\frac{10\times80}{1000}=80\%,
    \qquad
    \text{Subject-level Accuracy}=\frac{10}{10}=100\%.
\end{equation}

These examples demonstrate that subject-level performance depends not only on the total number of sample-level errors, but also on how these errors are distributed across subjects. Conventional sample-level cross-entropy loss does not explicitly account for this distribution. Moreover, since post-hoc majority voting is performed only during evaluation and does not participate in model training, subject-level prediction errors cannot be directly backpropagated to optimize the model. This motivates our subject-level cross-entropy loss, which additionally supervises the aggregated predictions of samples belonging to the same subject and better aligns model training with the final subject-level evaluation.

\section{Evaluation Methods}
\label{sec:evaluation_method}
Evaluation methods play a critical role in EEG-based disease detection, as EEG samples collected from the same subject are usually highly correlated. Depending on whether subjects are shared across the training, validation, and test sets, evaluation protocols can be broadly divided into two categories: \textbf{subject-dependent} and \textbf{subject-independent} evaluation.

\subsection{Subject-Dependent Evaluation}
\label{sec:subject_dependent}
In subject-dependent evaluation, EEG samples are split into training, validation, and test sets at the sample level. The splitting strategy can be random, predefined, $K$-fold, or other variants; as long as samples from the same subject can appear in multiple splits, the evaluation is considered subject-dependent. However, as demonstrated in previous research~\citep{wang2024causes}, this evaluation setting can lead to strong shortcut learning in single-class-per-subject EEG tasks, where the model may exploit subject-specific features rather than disease-related patterns. This issue is particularly relevant to EEG-based neurodegenerative disease detection, where all samples from the same subject usually share the same disease label.

\subsection{Subject-Independent Evaluation}
\label{sec:subject_independent}
In subject-independent evaluation, the split is performed at the subject level, such that samples from the same subject are assigned exclusively to the training, validation, or test set. This setting prevents subject information leakage and directly evaluates model generalization to unseen subjects. Depending on how subjects are partitioned, 4 commonly used subcategory methods can be considered.

\textbf{Fixed Split.}
Subjects are divided once into non-overlapping training, validation, and test sets according to predefined proportions, and the same partition is used across all experimental runs. This protocol is commonly adopted when a dataset provides official training, validation, and test splits. However, for EEG classification, performance may be highly dependent on a particular subject partition because of substantial inter-subject variability. Therefore, a single fixed split may provide a biased estimate of model performance when the dataset is relatively small.

\textbf{$K$-Fold Cross-Validation.}
Subjects are partitioned into $K$ non-overlapping folds. Each fold is used as the held-out test set in turn, while the remaining subjects are used for model training and validation. Results are averaged across all $K$ folds, allowing every subject to contribute to the test evaluation. However, inappropriate fold construction can still introduce bias. For example, if subjects are ordered by demographic characteristics and folds are created directly according to this order, some folds may contain predominantly male or female subjects. Therefore, subjects should generally be randomly shuffled, and stratified partitioning should be used when important demographic or class distributions need to be balanced.

\textbf{Monte Carlo Cross-Validation.}
Monte Carlo cross-validation (MCCV) repeatedly generates independent random subject-level splits according to predefined training, validation, and test ratios. To maintain a balanced diagnostic class distribution across splits, the partitioning is performed separately within each diagnostic class at the subject level. For example, if a dataset contains 100 AD subjects and 100 healthy controls and the split ratio is 60\%/20\%/20\%, each class is independently divided into 60 training subjects, 20 validation subjects, and 20 test subjects before the corresponding subsets are combined. We train and evaluate a new model for each random split, and summarize the final performance across multiple repetitions. Compared with a single fixed split, MCCV reduces dependence on one particular subject partition while allowing flexible control over split ratios. With a sufficient number of repetitions, MCCV provides a more stable estimate of model performance across different possible subject partitions. We therefore adopt MCCV as the primary evaluation protocol for comparing \myname with baseline methods in this study.

\textbf{Leave-One-Subject-Out.}
Leave-One-Subject-Out (LOSO) evaluation uses one subject as the test set and all remaining subjects as the training set. This procedure is repeated until every subject has been used once for testing, and the final performance is averaged across all subjects. LOSO provides direct per-subject evaluation and is particularly useful for small EEG datasets, where reserving a large independent test set may substantially reduce the amount of available training data. However, LOSO can be computationally expensive because the model must be trained once for every subject, making it impractical for datasets containing a large number of subjects. In addition, when no separate validation set is used, early stopping and test-set-based hyperparameter selection should be avoided, since they may introduce information leakage from the held-out subject. Consequently, hyperparameter tuning under LOSO requires additional care.

\section{Data Augmentation Banks}
\label{sec:data_augmentation_banks}

For contrastive pre-training, we employ a bank of data augmentation methods to enhance the model's robustness and generalization capabilities. During the forward pass in the training of each iteration, one augmentation method will be picked from available augmentation options with equal probability. The data augmentation methods include temporal flipping, temporal masking, frequency masking, channel masking, jittering, and dropout, and can be further expanded to more choices.

\textbf{1) Temporal Flippling.} We reverse the EEG data along the temporal dimension. The probability of applying this augmentation is controlled by a parameter \textit{prob}, with a default value of 0.5. \textbf{2) Temporal Masking.}
We randomly mask timestamps across all channels. The proportion of timestamps masked is controlled by the parameter \textit{ratio}, with a default value of 0.1. \textbf{3) Frequency Masking.} This method involves converting the EEG data into the frequency domain, randomly masking some frequency bands, and then converting it back. The proportion of frequency bands masked is controlled by the parameter \textit{ratio}, with a default value of 0.1. \textbf{4) Channel Masking.} We randomly mask channels across all timestamps. The proportion of channel masked is controlled by the parameter \textit{ratio}, with a default value of 0.1. \textbf{5) Jittering.} Random noise, ranging from 0 to 1, is added to the raw data. The intensity of the noise is adjusted by the parameter \textit{scale}, which is set by default to 0.1. \textbf{6) Dropout.} Similar to the dropout layer in neural networks, this method randomly drops some values. The proportion of values dropped is controlled by the parameter \textit{ratio}, with a default value of 0.1.

\section{Reproducibility and Implementation Details}
\label{sec:implementation_details}

To ensure reproducibility, we implement and evaluate all experiments under a unified training and evaluation framework. The source code for \myname and all baseline implementations is publicly available through the project repository link at the beginning of this paper, together with the released checkpoints of our proposed model. All datasets used in this study are publicly accessible, except for CNBPM, which is a private dataset and can be obtained by contacting the original authors. Detailed implementation settings and hyperparameters for each baseline are provided below, while complete running scripts for \myname and all baseline methods are also included in our repository to facilitate faithful reproduction of the reported results.

The selected comparison baselines include commonly used or state-of-the-art methods for time-series classification, medical time-series classification, EEG classification, and EEG foundation models. We also include methods specifically designed for AD-related detection, such as MNet~\citep{watanabe2024deep}, as well as methods previously evaluated on AD datasets, such as Medformer~\citep{wang2024medformer}.

\textbf{Manual Feature} utilize  32 features, including mean, variance, skewness, kurtosis, std, iqr, max, min, mean, median, delta power, theta power, alpha power, beta power, total power, theta alpha ratio, alpha beta ratio, delta relative power, theta relative power, alpha relative power, beta relative power, phase coherence, spectral centroid, spectral rolloff, spectral peak, average magnitude, median frequency, amplitude modulation, spectral entropy, tsallis entropy, and shannon entropy~\citep{tzimourta2019eeg, tzimourta2019analysis, kulkarni2017extracting, kanda2014clinician, waser2013eeg, tylova2013predictive, mora2019scale,wang2017enhanced, cassani2014effects, wang2015multiple, fraga2013characterizing, tait2019network, waser2016quantifying, trambaiolli2011improving,fahimi2017index, schmidt2013index, liu2016multiple, kanda2014clinician,garn2015quantitative, tylova2018unbiased, coronel2017quantitative, al2018complexity}. A linear projection layer is then applied to these manually extracted features for final classification.

\textbf{EEGConformer}~\citep{song2022eeg} uses convolutional modules to learn low-level local features and embed the raw data into patches for self-attention. We set \textit{e\_layers} = 12, \textit{n\_heads} = 8, \textit{d\_model} = 128, and \textit{d\_ff} = 256.

\textbf{EEGInception}~\citep{zhang2021eeg} uses different scales of convolutional kernels, combined with a spatial block for feature extraction. We set \textit{n\_blocks} = 3, \textit{channels} = (96,192,384), \textit{kernel\_sizes} = (8,16,32), \textit{depth\_multiplier} = 2, \textit{bottleneck\_channels} = 32.

\textbf{EEGNet}~\citep{lawhern2018eegnet} is a classic deep learning method for EEG decoding. It uses depthwise and separable convolutions to capture spatial and temporal features. We keep the same structure when applying convolutions, normalization, and activations as described in the paper.

\textbf{iTransformer}~\citep{liu2023itransformer} proposes a novel data-embedding method based on the transformer architecture for multivariate time-series analysis. It reverses the conventional embedding strategy. Instead of using multi-channel samples at a single time point into a single temporal token, iTransformer treats each channel/variate as an independent variable token. We set \textit{e\_layers} = 12, \textit{n\_heads} = 8, \textit{d\_model} = 128, and \textit{d\_ff} = 256.

\textbf{MedGNN}~\citep{fan2025towards} is a graph-nerual-network-based method for medical time-series classification. It uses a multi-resolution graph transformer architecture to model the dynamic dependencies and fuse the information from different resolutions. We set \textit{--resolution\_list} = 2,4,6,8, \textit{--nodedim} = 10, \textit{e\_layers} = 12, \textit{n\_heads} = 8, \textit{d\_model} = 128, and \textit{d\_ff} = 256.

\textbf{Medformer}~\citep{wang2024medformer} is designed for biomedical time series classification, including EEG and ECG. Cross-channel multi-granularity patch embedding and intra-inter-granularity self-attention are utilized. We extended our work based on this method. We set \textit{e\_layers} = 12, \textit{d\_model} = 128, and \textit{d\_ff} = 256, \textit{patch\_len\_list} = [5, 10, 20].

\textbf{MNet}~\citep{watanabe2024deep} is a convolution-based deep learning method for EEG-based AD detection. It contains 4 Con2D blocks for feature extraction. We keep the same structure and order when applying convolutions, normalization, and activations as described in the paper. 

\textbf{ModernTCN}~\citep{luo2024moderntcn} is a convolutional architecture designed for general time-series analysis and demonstrates strong performance, particularly in classification. It combines depthwise convolutions with multiple pointwise convolutions and introduces channel-independent embedding mechanisms. We set \textit{patch\_len} = 20, \textit{stride} = 10, \textit{num\_blocks} = 1 1 1 1,  \textit{large\_size} = 9 9 9 9, \textit{small\_size} = 5 5 5 5, \textit{dims} = 32 64 128 128. 

\textbf{PatchTST}~\citep{nie2022time} is a transformer-based model for time series prediction and representation learning, with its core innovation lying in a dual design of patching and channel independence. The model decomposes multi-channel data into multiple one-channel patches, treating them as semantically informative sub-sequence fragments for input. We set \textit{e\_layers} = 12, \textit{n\_heads} = 8, \textit{d\_model} = 128, and \textit{d\_ff} = 256, \textit{patch\_len} = 25. 

\textbf{TCN}~\citep{bai2018empirical} is a convolutional architecture designed explicitly for time-series modeling. They extend residual networks by incorporating causal and dilated convolutions, enabling effective learning of long-range temporal dependencies. We set \textit{e\_layers} = 12, \textit{n\_heads} = 8, \textit{d\_model} = 128, and \textit{d\_ff} = 256. 

\textbf{TimesNet}~\citep{wu2022timesnet} is a convolutional model for time-series analysis, especially on classification tasks. It transforms one-dimensional time-series data into two-dimensional tensors across multiple periodicities, thereby leveraging the strengths of 2D convolutional networks for time-series analysis. We set \textit{e\_layers} = 2, \textit{top\_k} = 3, \textit{d\_model} = 32, and \textit{d\_ff} = 64. 

\textbf{BIOT}~\citep{yang2024biot} is the first foundation model for biomedical time-series data, including EEG. It employs single-channel patch embedding to handle biosignals with varying channel counts. Each patch is mapped to tokens, with segment, channel, and positional embeddings added to make the tokens distinguishable. To align the channel configuration of our datasets with their checkpoints, we use a 1D convolution (Conv1D) for channel mapping. We retain the other default architectural parameters from the checkpoint. The pre-trained checkpoint is available at \url{https://github.com/ycq091044/BIOT/blob/main/pretrained-models/EEG-PREST-16-channels.ckpt}. 

\textbf{LaBraM}~\citep{jiang2024large} is the first EEG foundation model, trained on 2,500 hours of EEG recordings collected from multiple datasets. Its pre-training follows a two-step strategy that combines vector quantization with mask-based reconstruction. In our experiments, we use their released pre-trained checkpoint with default parameters and fine-tune it on 5 AD downstream datasets. To align the channel configuration of our datasets with their checkpoints, we use a 1D convolution (Conv1D) for channel mapping. We retain the other default architectural parameters from the checkpoint. The pre-trained checkpoint is available at \url{https://github.com/935963004/LaBraM/blob/main/checkpoints/labram-base.pth}. 

\textbf{CBraMod}~\citep{wang2024cbramod} is an SOTA EEG foundation model trained on the 9,000 hours TUEG dataset with 19 standard channels of the 10-20 system. They use a criss-cross transformer to perform spatial and temporal attention to capture features along two dimensions. We keep their default parameters to load their pre-trained checkpoint and fine-tune on 5 AD downstream datasets. The pre-trained checkpoint is available at \url{https://huggingface.co/weighting666/CBraMod/blob/main/pretrained_weights.pth}.

\textbf{CSBrain}~\citep{zhou2025csbrain} is an SOTA EEG foundation model that leverages brain regions to model cross-window and cross-region dependencies for diverse EEG decoding tasks. This model is also trained on the 9,000 hours TUEG dataset with 19 standard channels of the 10-20 system. We keep their default parameters to load their pre-trained checkpoint and fine-tune on 5 AD downstream datasets. The pre-trained checkpoint is available at \url{https://drive.google.com/drive/folders/1je-1TtdHv6klcd-kTlPNkiA1wLrxybva}.

\textbf{LEAD (Ours).}
We pre-train on 13 datasets (4 AD and 9 Non-AD) and then fine-tune on 5 downstream AD datasets. We set the patch length $L$ to 50 and the stride to 50. The sample- and subject-level cross-entropy loss weights $\alpha$ and $\beta$ are both set to 0.5. The contrastive loss coefficients $\lambda_1$ and $\lambda_2$ are set to 0.25 and 0.75. We set \textit{e\_layers} = 12, \textit{n\_heads} = 8, \textit{d\_model} = 128, \textit{d\_ff} = 256, and $G$ = 32, 8, 8, and 8 for pretraining, linear probing, fine-tuning, and fully supervised learning. The indices group shuffling, sampling rate embedding, use subject loss, and multi-sampling segmentation are all enabled, with \textit{multi\_sampling\_rate} set to 200, 100, and 50 Hz.

\section{Detailed Results}
\label{sec:detail_results}

This section reports detailed experimental results for different tasks, including baseline method comparisons and per-subject analysis. The evaluation is conducted on five downstream datasets: ADFSU, ADFTD, ADSZ, APAVA, and CNBPM. We report all 14 evaluation metrics, including sample-level accuracy, precision (macro-averaged), sensitivity/recall (macro-averaged), specificity (macro-averaged), F1 score (macro-averaged), AUROC (macro-averaged), and AUPRC (macro-averaged), and their corresponding subject-level metrics computed via majority voting (See para.~\ref{para:majority_vote}).

\subsection{Method Comparison Results}
\label{sub:method_comparison_detail}

\begin{table*}[t]
\centering
\def\arraystretch{1.0}
\caption{\textbf{Method Comparison Detailed Results on ADFSU.} The detailed results comparison between our method and baseline methods on the dataset ADFSU includes all 7 evaluation metrics. The \textcolor{myred}{\textbf{Top-1}}, \textcolor{myblue}{Top-2}, and \textcolor{mygreen}{Top-3} results are highlighted in red, blue, and green. 
}
\label{tab:adfsu_detail_results}
\resizebox{\textwidth}{!}{%
\begin{tabular}{clccccccc}

    \toprule
    \textbf{Datasets} & \textbf{Models} & \multicolumn{1}{c}{\textbf{Accuracy}} & \multicolumn{1}{c}{\textbf{Precision}} & \multicolumn{1}{c}{\textbf{Sensitivity}} & \multicolumn{1}{c}{\textbf{Specificity}} & \multicolumn{1}{c}{\textbf{F1 Score}} & \multicolumn{1}{c}{\textbf{AUROC}} & \multicolumn{1}{c}{\textbf{AUPRC}} \\
    \midrule
    \multirow{17}{*}{\begin{tabular}[c]{@{}l@{}}\;\makecell{\makecell{\textbf{ADFSU} \\ \textit{(Sample-Level)} \\ \textit{(4,048 Samples)} \\ \textit{(HC vs AD)}}} \end{tabular}} 
    & \textbf{ManualFeature} & 87.40\std{3.88} & 82.55\std{9.01} & 73.87\std{10.80} & 73.87\std{10.80} & 75.67\std{11.41} & 80.90\std{8.87} & 77.35\std{9.05} \\
    & \textbf{EEGConformer} & \textcolor{myblue}{96.53\std{1.93}} & \textcolor{myblue}{96.75\std{2.12}} & \textcolor{myblue}{92.33\std{4.46}} & \textcolor{myblue}{92.33\std{4.46}} & \textcolor{myblue}{94.22\std{3.28}} & \textcolor{mygreen}{98.87\std{0.95}} & \textcolor{myblue}{98.36\std{1.24}} \\
    & \textbf{EEGInception} & 94.60\std{0.98} & 94.49\std{1.15} & 88.63\std{4.21} & 88.63\std{4.21} & 90.88\std{2.13} & 98.73\std{0.75} & 97.99\std{1.19} \\
    & \textbf{EEGNet} & 82.27\std{10.27} & 77.52\std{10.71} & 74.92\std{6.79} & 74.92\std{6.79} & 74.48\std{9.91} & 85.88\std{7.95} & 82.41\std{10.23} \\
    & \textbf{iTransformer} & 87.80\std{1.07} & 86.93\std{3.61} & 72.75\std{2.50} & 72.75\std{2.50} & 76.80\std{2.31} & 88.16\std{1.86} & 85.59\std{1.34} \\
    & \textbf{MedGNN}  & 94.13\std{1.50} & 94.34\std{3.59} & 87.08\std{1.86} & 87.08\std{1.86} & 90.09\std{2.37} & 98.06\std{1.42} & 97.08\std{2.13} \\
    & \textbf{Medformer}  & 92.73\std{0.80} & 93.58\std{0.78} & 83.33\std{2.71} & 83.33\std{2.71} & 87.15\std{1.91} & 97.48\std{1.32} & 96.30\std{1.18} \\
    & \textbf{MNet}  & 88.40\std{3.38} & 90.34\std{3.43} & 72.12\std{8.24} & 72.12\std{8.24} & 76.19\std{8.72} & 92.75\std{5.08} & 90.30\std{5.71} \\
    & \textbf{ModernTCN}  & 87.80\std{1.19} & 90.98\std{2.03} & 70.38\std{2.93} & 70.38\std{2.93} & 75.09\std{3.18} & 91.94\std{1.84} & 89.57\std{2.09} \\
    & \textbf{PatchTST}  & 87.13\std{1.42} & 83.81\std{2.01} & 72.83\std{4.55} & 72.83\std{4.55} & 76.03\std{4.27} & 89.12\std{1.95} & 85.53\std{2.14} \\
    & \textbf{TCN}  & 94.27\std{1.70} & 95.07\std{2.54} & 86.79\std{3.71} & 86.79\std{3.71} & 90.10\std{3.13} & 98.82\std{0.73} & \textcolor{mygreen}{98.00\std{1.23}} \\
    & \textbf{TimesNet}  & 90.60\std{2.39} & 92.28\std{2.38} & 77.88\std{6.26} & 77.88\std{6.26} & 82.08\std{6.29} & 95.21\std{2.45} & 93.39\std{2.27} \\

    \cmidrule(lr){2-9}
    & \textbf{BIOT}  & 93.47\std{1.51} & 92.80\std{2.71} & 86.17\std{2.83} & 86.17\std{2.83} & 88.94\std{2.69} & 96.78\std{3.04} & 95.60\std{3.05} \\
    & \textbf{LaBraM}  & \textcolor{mygreen}{95.47\std{1.77}} & \textcolor{mygreen}{96.14\std{0.98}} & 89.67\std{4.97} & 89.67\std{4.97} & 92.24\std{3.38} & \textcolor{myblue}{98.90\std{0.53}} & 97.98\std{0.88} \\
    & \textbf{CBraMod}  & 95.40\std{2.28} & 94.56\std{2.40} & \textcolor{mygreen}{90.88\std{5.60}} & \textcolor{mygreen}{90.88\std{5.60}} & \textcolor{mygreen}{92.32\std{4.27}} & 98.47\std{1.49} & 97.59\std{2.27} \\
    & \textbf{CSBrain}  & 94.67\std{1.14} & 93.96\std{1.28} & 89.17\std{3.69} & 89.17\std{3.69} & 91.11\std{2.34} & 98.23\std{0.51} & 97.23\std{0.63} \\

    \cmidrule(lr){2-9}
    & \textbf{LEAD (Ours)}  & \textcolor{myred}{\textbf{98.18\std{0.61}}} & \textcolor{myred}{\textbf{98.70\std{0.44}}} & \textcolor{myred}{\textbf{95.62\std{1.54}}} & \textcolor{myred}{\textbf{95.62\std{1.54}}} & \textcolor{myred}{\textbf{97.06\std{1.02}}} & \textcolor{myred}{\textbf{99.92\std{0.07}}} & \textcolor{myred}{\textbf{99.85\std{0.13}}} \\

    \midrule
    \multirow{17}{*}{\begin{tabular}[c]{@{}l@{}}\;\makecell{\makecell{\textbf{ADFSU} \\ \textit{(Subject-Level)} \\ \textit{(92 Subjects)} \\ \textit{(HC vs AD)}}} \end{tabular}} 
    & \textbf{ManualFeature} & 92.00\std{7.48} & 85.78\std{23.02} & 80.00\std{18.71} & 80.00\std{18.71} & 81.05\std{20.29} & 100.00\std{0.00} & 100.00\std{0.00} \\
    & \textbf{EEGConformer} & \textcolor{myblue}{100.00\std{0.00}} & \textcolor{myblue}{100.00\std{0.00}} & \textcolor{myblue}{100.00\std{0.00}} & \textcolor{myblue}{100.00\std{0.00}} & \textcolor{myblue}{100.00\std{0.00}} & \textcolor{mygreen}{100.00\std{0.00}} & \textcolor{myblue}{100.00\std{0.00}} \\
    & \textbf{EEGInception} & 100.00\std{0.00} & 100.00\std{0.00} & 100.00\std{0.00} & 100.00\std{0.00} & 100.00\std{0.00} & 100.00\std{0.00} & \textcolor{mygreen}{100.00\std{0.00}} \\
    & \textbf{EEGNet} & 88.00\std{11.66} & 80.89\std{23.26} & 81.25\std{18.54} & 81.25\std{18.54} & 78.37\std{21.07} & 100.00\std{0.00} & 100.00\std{0.00} \\
    & \textbf{iTransformer} & 90.00\std{6.32} & 84.67\std{22.44} & 75.00\std{15.81} & 75.00\std{15.81} & 77.12\std{18.02} & 100.00\std{0.00} & 100.00\std{0.00} \\
    & \textbf{MedGNN}  & 100.00\std{0.00} & 100.00\std{0.00} & 100.00\std{0.00} & 100.00\std{0.00} & 100.00\std{0.00} & 100.00\std{0.00} & 100.00\std{0.00} \\
    & \textbf{Medformer}  & 100.00\std{0.00} & 100.00\std{0.00} & 100.00\std{0.00} & 100.00\std{0.00} & 100.00\std{0.00} & 100.00\std{0.00} & 100.00\std{0.00} \\
    & \textbf{MNet}  & 84.00\std{4.90} & 61.78\std{26.67} & 60.00\std{12.25} & 60.00\std{12.25} & 58.82\std{17.61} & 100.00\std{0.00} & 100.00\std{0.00} \\
    & \textbf{ModernTCN}  & 84.00\std{4.90} & 61.78\std{26.67} & 60.00\std{12.25} & 60.00\std{12.25} & 58.82\std{17.61} & 100.00\std{0.00} & 100.00\std{0.00} \\
    & \textbf{PatchTST}  & 90.00\std{6.32} & 84.67\std{22.44} & 75.00\std{15.81} & 75.00\std{15.81} & 77.12\std{18.02} & 100.00\std{0.00} & 100.00\std{0.00} \\
    & \textbf{TCN}  & 98.00\std{4.00} & 98.89\std{2.22} & 95.00\std{10.00} & 95.00\std{10.00} & 96.08\std{7.84} & 100.00\std{0.00} & 100.00\std{0.00} \\
    & \textbf{TimesNet}  & 94.00\std{8.00} & 86.89\std{23.54} & 85.00\std{20.00} & 85.00\std{20.00} & 84.97\std{21.64} & 100.00\std{0.00} & 100.00\std{0.00} \\

    \cmidrule(lr){2-9}
    & \textbf{BIOT}  & 100.00\std{0.00} & 100.00\std{0.00} & 100.00\std{0.00} & 100.00\std{0.00} & 100.00\std{0.00} & 100.00\std{0.00} & 100.00\std{0.00} \\

    & \textbf{LaBraM}  & \textcolor{mygreen}{100.00\std{0.00}} & \textcolor{mygreen}{100.00\std{0.00}} & 100.00\std{0.00} & 100.00\std{0.00} & 100.00\std{0.00} & \textcolor{myblue}{100.00\std{0.00}} & 100.00\std{0.00} \\
    & \textbf{CBraMod}  & 100.00\std{0.00} & 100.00\std{0.00} & \textcolor{mygreen}{100.00\std{0.00}} & \textcolor{mygreen}{100.00\std{0.00}} & \textcolor{mygreen}{100.00\std{0.00}} & 100.00\std{0.00} & 100.00\std{0.00} \\
    & \textbf{CSBrain}  & 100.00\std{0.00} & 100.00\std{0.00} & 100.00\std{0.00} & 100.00\std{0.00} & 100.00\std{0.00} & 100.00\std{0.00} & 100.00\std{0.00} \\

    \cmidrule(lr){2-9}
    & \textbf{LEAD (Ours)}  & \textcolor{myred}{\textbf{100.00\std{0.00}}} & \textcolor{myred}{\textbf{100.00\std{0.00}}} & \textcolor{myred}{\textbf{100.00\std{0.00}}} & \textcolor{myred}{\textbf{100.00\std{0.00}}} & \textcolor{myred}{\textbf{100.00\std{0.00}}} & \textcolor{myred}{\textbf{100.00\std{0.00}}} & \textcolor{myred}{\textbf{100.00\std{0.00}}} \\

\bottomrule
\end{tabular}
}
\end{table*}

\begin{table*}[t]
\centering
\def\arraystretch{1.0}
\caption{\textbf{Method Comparison Detailed Results on ADFTD.} The detailed results comparison between our method and baseline methods on the dataset ADFTD includes all 7 evaluation metrics. The \textcolor{myred}{\textbf{Top-1}}, \textcolor{myblue}{Top-2}, and \textcolor{mygreen}{Top-3} results are highlighted in red, blue, and green. 
}
\label{tab:adftd_detail_results}
\resizebox{\textwidth}{!}{%
\begin{tabular}{clccccccc}

    \toprule
    \textbf{Datasets} & \textbf{Models} & \multicolumn{1}{c}{\textbf{Accuracy}} & \multicolumn{1}{c}{\textbf{Precision}} & \multicolumn{1}{c}{\textbf{Sensitivity}} & \multicolumn{1}{c}{\textbf{Specificity}} & \multicolumn{1}{c}{\textbf{F1 Score}} & \multicolumn{1}{c}{\textbf{AUROC}} & \multicolumn{1}{c}{\textbf{AUPRC}} \\
    \midrule
    \multirow{17}{*}{\begin{tabular}[c]{@{}l@{}}\;\makecell{\makecell{\textbf{ADFTD} \\ \textit{(Sample-Level)} \\ \textit{(167,083 Samples)} \\ \textit{(HC vs AD vs FTD)}}} \end{tabular}} 
    & \textbf{ManualFeature} & 46.98\std{4.11} & 45.82\std{4.61} & 45.27\std{4.08} & 73.13\std{2.00} & 44.73\std{3.88} & 62.89\std{4.31} & 43.69\std{3.63} \\
    & \textbf{EEGConformer} & 61.28\std{2.83} & 62.39\std{4.23} & 58.82\std{3.53} & 79.87\std{1.51} & 58.59\std{3.84} & 80.61\std{3.90} & 69.14\std{5.37} \\
    & \textbf{EEGInception} & 64.24\std{3.85} & 66.61\std{2.85} & 64.03\std{3.62} & 82.16\std{1.76} & 62.69\std{4.06} & 84.01\std{2.76} & 73.68\std{2.35} \\
    & \textbf{EEGNet} & 45.22\std{3.93} & 40.83\std{8.66} & 42.91\std{4.03} & 71.60\std{1.93} & 40.31\std{6.74} & 60.93\std{4.42} & 43.59\std{5.12} \\
    & \textbf{iTransformer} & 56.27\std{3.61} & 54.65\std{3.94} & 54.71\std{3.54} & 77.93\std{1.73} & 54.33\std{3.67} & 73.98\std{4.02} & 59.26\std{5.30} \\
    & \textbf{MedGNN}  & 69.60\std{3.85} & 70.23\std{4.08} & 68.16\std{4.74} & 84.48\std{2.17} & 67.67\std{4.59} & 85.52\std{2.57} & 76.86\std{3.18} \\
    & \textbf{Medformer}  & 66.66\std{4.89} & 67.20\std{4.49} & 65.14\std{6.15} & 82.80\std{2.69} & 64.83\std{6.72} & 84.13\std{3.27} & 74.59\std{5.16} \\
    & \textbf{MNet}  & 63.91\std{6.04} & 63.84\std{7.99} & 61.89\std{7.28} & 81.46\std{3.11} & 59.91\std{9.76} & 82.30\std{5.04} & 71.32\std{8.51} \\
    & \textbf{ModernTCN}  & 60.38\std{3.42} & 59.21\std{3.91} & 58.59\std{3.67} & 79.82\std{1.71} & 58.53\std{3.71} & 77.25\std{3.33} & 63.44\std{4.55} \\
    & \textbf{PatchTST}  & 55.48\std{3.23} & 54.35\std{6.92} & 52.24\std{2.94} & 76.72\std{1.53} & 50.95\std{4.04} & 71.30\std{4.75} & 56.53\std{6.94} \\
    & \textbf{TCN}  & 66.04\std{2.46} & 64.89\std{2.47} & 64.44\std{2.32} & 82.68\std{1.19} & 64.26\std{2.40} & 82.94\std{1.89} & 71.43\std{3.32} \\
    & \textbf{TimesNet}  & 61.85\std{3.36} & 61.13\std{4.67} & 59.72\std{4.92} & 80.38\std{1.87} & 58.67\std{5.67} & 79.18\std{3.39} & 66.47\std{6.35} \\

    \cmidrule(lr){2-9}
    & \textbf{BIOT}  & \textcolor{mygreen}{70.94\std{5.66}} & \textcolor{mygreen}{70.86\std{5.53}} & \textcolor{mygreen}{69.86\std{5.91}} & \textcolor{mygreen}{85.14\std{2.96}} & \textcolor{mygreen}{69.79\std{5.90}} & 86.85\std{4.77} & 78.40\std{6.69} \\
    & \textbf{LaBraM}  & \textcolor{myblue}{75.71\std{3.34}} & \textcolor{myblue}{75.97\std{2.48}} & \textcolor{myblue}{74.51\std{4.20}} & \textcolor{myblue}{87.59\std{1.82}} & \textcolor{myblue}{74.37\std{4.31}} & \textcolor{myblue}{90.29\std{2.76}} & \textcolor{myblue}{83.16\std{4.48}} \\
    & \textbf{CBraMod}  & 69.92\std{3.95} & 69.52\std{4.41} & 68.58\std{3.97} & 84.60\std{1.94} & 68.33\std{4.53} & \textcolor{mygreen}{86.95\std{2.89}} & 78.06\std{4.26} \\
    & \textbf{CSBrain}  & 70.68\std{2.12} & 70.42\std{2.02} & 69.61\std{2.49} & 85.03\std{1.11} & 69.40\std{2.63} & 86.82\std{1.28} & \textcolor{mygreen}{78.66\std{1.93}} \\

    \cmidrule(lr){2-9}
    & \textbf{LEAD (Ours)}  & \textcolor{myred}{\textbf{81.70\std{4.74}}} & \textcolor{myred}{\textbf{82.93\std{4.40}}} & \textcolor{myred}{\textbf{80.78\std{4.95}}} & \textcolor{myred}{\textbf{90.52\std{2.45}}} & \textcolor{myred}{\textbf{81.01\std{5.02}}} & \textcolor{myred}{\textbf{94.03\std{1.33}}} & \textcolor{myred}{\textbf{90.37\std{2.14}}} \\

    \midrule
    \multirow{17}{*}{\begin{tabular}[c]{@{}l@{}}\;\makecell{\makecell{\textbf{ADFTD} \\ \textit{(Subject-Level)} \\ \textit{(88 Subjects)} \\ \textit{(HC vs AD vs FTD)}}} \end{tabular}} 
    & \textbf{ManualFeature} & 62.00\std{4.00} & 56.03\std{15.90} & 59.44\std{4.16} & 80.16\std{1.81} & 54.26\std{7.80} & 86.15\std{11.29} & 83.35\std{13.88} \\
    & \textbf{EEGConformer} & 74.00\std{10.20} & 70.95\std{20.86} & 72.22\std{10.97} & 86.03\std{5.35} & 68.45\std{15.21} & 95.56\std{3.67} & 94.74\std{3.38} \\
    & \textbf{EEGInception} & 82.00\std{13.27} & 80.67\std{21.12} & 81.67\std{13.68} & 90.79\std{6.78} & 79.47\std{17.70} & 96.55\std{4.93} & 97.00\std{3.92} \\
    & \textbf{EEGNet} & 64.00\std{10.20} & 58.81\std{22.32} & 60.56\std{11.17} & 80.63\std{5.60} & 54.82\std{14.34} & 68.69\std{15.52} & 63.73\std{17.89} \\
    & \textbf{iTransformer} & 72.00\std{7.48} & 72.56\std{14.91} & 70.56\std{7.16} & 85.87\std{3.56} & 67.17\std{9.15} & 90.28\std{6.75} & 87.86\std{7.97} \\
    & \textbf{MedGNN}  & 78.00\std{4.00} & 83.33\std{2.79} & 77.78\std{4.65} & 89.05\std{2.09} & 75.23\std{4.81} & 94.72\std{3.45} & 92.93\std{3.83} \\
    & \textbf{Medformer}  & 80.00\std{10.95} & 84.11\std{8.15} & 80.00\std{11.17} & 89.68\std{5.66} & 78.98\std{12.08} & 94.76\std{5.13} & 94.10\std{5.75} \\
    & \textbf{MNet}  & 74.00\std{12.00} & 70.30\std{21.93} & 73.33\std{13.10} & 86.83\std{5.89} & 68.63\std{17.53} & 92.38\std{7.59} & 88.19\std{11.83} \\
    & \textbf{ModernTCN}  & 78.00\std{7.48} & 78.37\std{13.91} & 77.22\std{8.85} & 88.41\std{4.16} & 74.39\std{11.35} & 93.77\std{4.71} & 92.62\std{5.61} \\
    & \textbf{PatchTST}  & 72.00\std{9.80} & 68.81\std{21.28} & 69.44\std{10.24} & 85.08\std{5.00} & 65.04\std{14.32} & 87.78\std{12.73} & 84.36\std{16.90} \\
    & \textbf{TCN}  & 84.00\std{4.90} & 88.67\std{4.40} & 83.33\std{5.83} & 91.43\std{2.81} & 82.80\std{6.10} & 96.43\std{1.96} & 95.31\std{2.36} \\
    & \textbf{TimesNet}  & 72.00\std{11.66} & 61.86\std{21.27} & 71.11\std{13.10} & 86.03\std{5.82} & 64.23\std{17.73} & 94.72\std{3.70} & 94.62\std{3.39} \\

    \cmidrule(lr){2-9}
    & \textbf{BIOT}  & \textcolor{mygreen}{88.00\std{9.80}} & \textcolor{mygreen}{91.11\std{7.54}} & \textcolor{mygreen}{87.78\std{10.03}} & \textcolor{mygreen}{93.65\std{5.19}} & \textcolor{mygreen}{88.10\std{9.72}} & 97.10\std{2.78} & 96.28\std{3.68} \\
    & \textbf{LaBraM}  & \textcolor{myblue}{92.00\std{7.48}} & \textcolor{myblue}{94.44\std{4.65}} & \textcolor{myblue}{91.67\std{8.24}} & \textcolor{myblue}{95.87\std{4.09}} & \textcolor{myblue}{91.81\std{7.56}} & \textcolor{myred}{\textbf{99.44\std{0.68}}} & \textcolor{myred}{\textbf{99.33\std{0.82}}} \\
    & \textbf{CBraMod}  & 84.00\std{4.90} & 88.78\std{4.13} & 82.78\std{5.09} & 91.43\std{2.48} & 82.21\std{6.30} & 97.30\std{2.05} & 96.44\std{2.50} \\
    & \textbf{CSBrain}  & 80.00\std{6.32} & 84.67\std{4.88} & 79.44\std{6.24} & 89.68\std{3.01} & 78.33\std{7.52} & \textcolor{mygreen}{97.74\std{2.65}} & \textcolor{mygreen}{97.72\std{2.56}} \\

    \cmidrule(lr){2-9}
    & \textbf{LEAD (Ours)}  & \textcolor{myred}{\textbf{94.00\std{4.90}}} & \textcolor{myred}{\textbf{95.67\std{3.59}}} & \textcolor{myred}{\textbf{93.89\std{5.09}}} & \textcolor{myred}{\textbf{96.83\std{2.61}}} & \textcolor{myred}{\textbf{93.95\std{4.95}}} & \textcolor{myblue}{99.17\std{1.11}} & \textcolor{myblue}{99.11\std{1.14}} \\

\bottomrule
\end{tabular}
}
\end{table*}

\begin{table*}[t]
\centering
\def\arraystretch{1.0}
\caption{\textbf{Method Comparison Detailed Results on ADSZ.} The detailed results comparison between our method and baseline methods on the dataset ADSZ includes all 7 evaluation metrics. The \textcolor{myred}{\textbf{Top-1}}, \textcolor{myblue}{Top-2}, and \textcolor{mygreen}{Top-3} results are highlighted in red, blue, and green. 
}
\label{tab:adsz_detail_results}
\resizebox{\textwidth}{!}{%
\begin{tabular}{clccccccc}

    \toprule
    \textbf{Datasets} & \textbf{Models} & \multicolumn{1}{c}{\textbf{Accuracy}} & \multicolumn{1}{c}{\textbf{Precision}} & \multicolumn{1}{c}{\textbf{Sensitivity}} & \multicolumn{1}{c}{\textbf{Specificity}} & \multicolumn{1}{c}{\textbf{F1 Score}} & \multicolumn{1}{c}{\textbf{AUROC}} & \multicolumn{1}{c}{\textbf{AUPRC}} \\
    \midrule
    \multirow{17}{*}{\begin{tabular}[c]{@{}l@{}}\;\makecell{\makecell{\textbf{ADSZ} \\ \textit{(Sample-Level)} \\ \textit{(1,128 Samples)} \\ \textit{(HC vs AD)}}} \end{tabular}} 
    & \textbf{ManualFeature} & 65.64\std{13.56} & 55.33\std{25.09} & 64.85\std{14.35} & 64.85\std{14.35} & 58.42\std{21.26} & 67.20\std{15.34} & 64.07\std{12.94} \\
    & \textbf{EEGConformer} & \textcolor{myblue}{94.73\std{3.47}} & \textcolor{myblue}{94.91\std{3.36}} & \textcolor{myblue}{94.91\std{3.32}} & \textcolor{myblue}{94.91\std{3.32}} & \textcolor{myblue}{94.73\std{3.47}} & \textcolor{mygreen}{98.50\std{1.48}} & \textcolor{mygreen}{98.52\std{1.49}} \\
    & \textbf{EEGInception} & 92.62\std{4.43} & 93.43\std{3.85} & 92.93\std{4.16} & 92.93\std{4.16} & 92.60\std{4.44} & 98.06\std{1.66} & 98.04\std{1.75} \\
    & \textbf{EEGNet} & 66.27\std{12.39} & 71.26\std{13.28} & 67.19\std{12.37} & 67.19\std{12.37} & 64.90\std{12.88} & 77.18\std{19.10} & 78.29\std{17.21} \\
    & \textbf{iTransformer} & 73.29\std{4.86} & 73.91\std{5.53} & 73.30\std{4.80} & 73.30\std{4.80} & 73.13\std{4.75} & 81.86\std{4.90} & 82.02\std{5.15} \\
    & \textbf{MedGNN}  & 88.62\std{8.00} & 89.11\std{7.79} & 88.84\std{7.86} & 88.84\std{7.86} & 88.59\std{8.01} & 95.90\std{4.10} & 96.05\std{3.98} \\
    & \textbf{Medformer}  & 88.78\std{5.71} & 88.81\std{5.76} & 88.73\std{5.77} & 88.73\std{5.77} & 88.73\std{5.76} & 94.18\std{3.72} & 94.22\std{3.91} \\
    & \textbf{MNet}  & 79.26\std{7.34} & 79.67\std{7.36} & 79.42\std{7.29} & 79.42\std{7.29} & 79.21\std{7.35} & 88.87\std{5.92} & 89.64\std{5.23} \\
    & \textbf{ModernTCN}  & 76.15\std{3.85} & 77.15\std{3.25} & 75.94\std{3.83} & 75.94\std{3.83} & 75.73\std{4.05} & 84.42\std{4.30} & 84.40\std{4.18} \\
    & \textbf{PatchTST}  & 71.13\std{10.02} & 71.93\std{10.56} & 71.48\std{10.34} & 71.48\std{10.34} & 71.05\std{10.02} & 79.10\std{12.54} & 79.27\std{11.47} \\
    & \textbf{TCN}  & 89.00\std{9.14} & 89.05\std{9.28} & 88.93\std{9.25} & 88.93\std{9.25} & 88.93\std{9.23} & 94.02\std{6.85} & 93.44\std{7.78} \\
    & \textbf{TimesNet}  & 83.88\std{3.58} & 84.16\std{3.47} & 83.87\std{3.57} & 83.87\std{3.57} & 83.80\std{3.60} & 88.12\std{9.12} & 87.78\std{9.75} \\

    \cmidrule(lr){2-9}
    & \textbf{BIOT}  & 90.88\std{4.66} & 91.44\std{4.13} & 91.04\std{4.75} & 91.04\std{4.75} & 90.83\std{4.72} & 96.73\std{2.98} & 96.52\std{3.55} \\
    & \textbf{LaBraM}  & 91.23\std{4.37} & 92.10\std{3.37} & 91.16\std{4.38} & 91.16\std{4.38} & 91.12\std{4.50} & 97.54\std{1.88} & 97.58\std{1.94} \\
    & \textbf{CBraMod}  & 84.46\std{8.37} & 88.51\std{4.55} & 84.62\std{8.54} & 84.62\std{8.54} & 83.70\std{9.28} & 96.67\std{3.69} & 96.77\std{3.44} \\
    & \textbf{CSBrain}  & \textcolor{mygreen}{93.18\std{4.03}} & \textcolor{mygreen}{93.66\std{3.56}} & \textcolor{mygreen}{93.36\std{3.78}} & \textcolor{mygreen}{93.36\std{3.78}} & \textcolor{mygreen}{93.16\std{4.03}} & \textcolor{myblue}{98.87\std{1.14}} & \textcolor{myblue}{98.92\std{1.10}} \\

    \cmidrule(lr){2-9}
    & \textbf{LEAD (Ours)}  & \textcolor{myred}{\textbf{97.42\std{2.74}}} & \textcolor{myred}{\textbf{97.60\std{2.59}}} & \textcolor{myred}{\textbf{97.51\std{2.57}}} & \textcolor{myred}{\textbf{97.51\std{2.57}}} & \textcolor{myred}{\textbf{97.42\std{2.74}}} & \textcolor{myred}{\textbf{100.00\std{0.00}}} & \textcolor{myred}{\textbf{100.00\std{0.00}}} \\

    \midrule
    \multirow{17}{*}{\begin{tabular}[c]{@{}l@{}}\;\makecell{\makecell{\textbf{ADSZ} \\ \textit{(Subject-Level)} \\ \textit{(48 Subjects)} \\ \textit{(HC vs AD)}}} \end{tabular}} 
    & \textbf{ManualFeature} & 70.00\std{24.49} & 60.00\std{33.91} & 70.00\std{24.49} & 70.00\std{24.49} & 63.05\std{30.68} & 75.56\std{22.39} & 77.33\std{22.84} \\
    & \textbf{EEGConformer} & \textcolor{myblue}{100.00\std{0.00}} & \textcolor{myblue}{100.00\std{0.00}} & \textcolor{myblue}{100.00\std{0.00}} & \textcolor{myblue}{100.00\std{0.00}} & \textcolor{myblue}{100.00\std{0.00}} & \textcolor{mygreen}{100.00\std{0.00}} & \textcolor{mygreen}{100.00\std{0.00}} \\
    & \textbf{EEGInception} & 100.00\std{0.00} & 100.00\std{0.00} & 100.00\std{0.00} & 100.00\std{0.00} & 100.00\std{0.00} & 100.00\std{0.00} & 100.00\std{0.00} \\
    & \textbf{EEGNet} & 66.67\std{18.26} & 67.00\std{26.38} & 66.67\std{18.26} & 66.67\std{18.26} & 61.38\std{22.11} & 84.44\std{18.39} & 86.33\std{17.27} \\
    & \textbf{iTransformer} & 100.00\std{0.00} & 100.00\std{0.00} & 100.00\std{0.00} & 100.00\std{0.00} & 100.00\std{0.00} & 100.00\std{0.00} & 100.00\std{0.00} \\
    & \textbf{MedGNN}  & 96.67\std{6.67} & 97.50\std{5.00} & 96.67\std{6.67} & 96.67\std{6.67} & 96.57\std{6.86} & 100.00\std{0.00} & 100.00\std{0.00} \\
    & \textbf{Medformer}  & 100.00\std{0.00} & 100.00\std{0.00} & 100.00\std{0.00} & 100.00\std{0.00} & 100.00\std{0.00} & 100.00\std{0.00} & 100.00\std{0.00} \\
    & \textbf{MNet}  & 90.00\std{13.33} & 90.83\std{13.02} & 90.00\std{13.33} & 90.00\std{13.33} & 89.90\std{13.38} & 95.56\std{5.44} & 96.67\std{4.08} \\
    & \textbf{ModernTCN}  & 90.00\std{8.16} & 92.50\std{6.12} & 90.00\std{8.16} & 90.00\std{8.16} & 89.71\std{8.40} & 100.00\std{0.00} & 100.00\std{0.00} \\
    & \textbf{PatchTST}  & 86.67\std{19.44} & 87.50\std{19.36} & 86.67\std{19.44} & 86.67\std{19.44} & 86.29\std{19.99} & 95.56\std{8.89} & 96.67\std{6.67} \\
    & \textbf{TCN}  & 96.67\std{6.67} & 97.50\std{5.00} & 96.67\std{6.67} & 96.67\std{6.67} & 96.57\std{6.86} & 97.78\std{4.44} & 98.33\std{3.33} \\
    & \textbf{TimesNet}  & 96.67\std{6.67} & 97.50\std{5.00} & 96.67\std{6.67} & 96.67\std{6.67} & 96.57\std{6.86} & 100.00\std{0.00} & 100.00\std{0.00} \\

    \cmidrule(lr){2-9}
    & \textbf{BIOT}  & 100.00\std{0.00} & 100.00\std{0.00} & 100.00\std{0.00} & 100.00\std{0.00} & 100.00\std{0.00} & 100.00\std{0.00} & 100.00\std{0.00} \\
    & \textbf{LaBraM}  & 100.00\std{0.00} & 100.00\std{0.00} & 100.00\std{0.00} & 100.00\std{0.00} & 100.00\std{0.00} & 100.00\std{0.00} & 100.00\std{0.00} \\
    & \textbf{CBraMod}  & 90.00\std{13.33} & 93.50\std{8.31} & 90.00\std{13.33} & 90.00\std{13.33} & 89.07\std{14.85} & 100.00\std{0.00} & 100.00\std{0.00} \\
    & \textbf{CSBrain}  & \textcolor{mygreen}{100.00\std{0.00}} & \textcolor{mygreen}{100.00\std{0.00}} & \textcolor{mygreen}{100.00\std{0.00}} & \textcolor{mygreen}{100.00\std{0.00}} & \textcolor{mygreen}{100.00\std{0.00}} & \textcolor{myblue}{100.00\std{0.00}} & \textcolor{myblue}{100.00\std{0.00}} \\

    \cmidrule(lr){2-9}
    & \textbf{LEAD (Ours)}  & \textcolor{myred}{\textbf{100.00\std{0.00}}} & \textcolor{myred}{\textbf{100.00\std{0.00}}} & \textcolor{myred}{\textbf{100.00\std{0.00}}} & \textcolor{myred}{\textbf{100.00\std{0.00}}} & \textcolor{myred}{\textbf{100.00\std{0.00}}} & \textcolor{myred}{\textbf{100.00\std{0.00}}} & \textcolor{myred}{\textbf{100.00\std{0.00}}} \\

\bottomrule
\end{tabular}
}
\end{table*}

\begin{table*}[t]
\centering
\def\arraystretch{1.0}
\caption{\textbf{Method Comparison Detailed Results on APAVA.} The detailed results comparison between our method and baseline methods on the dataset APAVA includes all 7 evaluation metrics. The \textcolor{myred}{\textbf{Top-1}}, \textcolor{myblue}{Top-2}, and \textcolor{mygreen}{Top-3} results are highlighted in red, blue, and green. 
}
\label{tab:apava_detail_results}
\resizebox{\textwidth}{!}{%
\begin{tabular}{clccccccc}

    \toprule
    \textbf{Datasets} & \textbf{Models} & \multicolumn{1}{c}{\textbf{Accuracy}} & \multicolumn{1}{c}{\textbf{Precision}} & \multicolumn{1}{c}{\textbf{Sensitivity}} & \multicolumn{1}{c}{\textbf{Specificity}} & \multicolumn{1}{c}{\textbf{F1 Score}} & \multicolumn{1}{c}{\textbf{AUROC}} & \multicolumn{1}{c}{\textbf{AUPRC}} \\
    \midrule
    \multirow{17}{*}{\begin{tabular}[c]{@{}l@{}}\;\makecell{\makecell{\textbf{APAVA} \\ \textit{(Sample-Level)} \\ \textit{(9,282 Samples)} \\ \textit{(HC vs AD)}}} \end{tabular}} 
    & \textbf{ManualFeature} & 70.40\std{4.92} & 77.53\std{8.45} & 64.55\std{5.25} & 64.55\std{5.25} & 63.22\std{6.51} & 66.49\std{5.78} & 64.17\std{5.79} \\
    & \textbf{EEGConformer} & 76.90\std{3.19} & 81.85\std{2.78} & 72.65\std{4.10} & 72.65\std{4.10} & 73.05\std{4.93} & 83.07\std{3.70} & 83.43\std{3.70} \\
    & \textbf{EEGInception} & 74.62\std{2.02} & 76.52\std{3.75} & 71.54\std{2.07} & 71.54\std{2.07} & 71.87\std{2.22} & 80.17\std{3.54} & 80.95\std{4.10} \\
    & \textbf{EEGNet} & 71.38\std{2.38} & 72.95\std{2.80} & 68.11\std{3.64} & 68.11\std{3.64} & 67.84\std{4.60} & 78.04\std{4.19} & 78.65\std{3.34} \\
    & \textbf{iTransformer} & 74.51\std{1.52} & 74.88\std{1.21} & 71.67\std{2.24} & 71.67\std{2.24} & 72.13\std{2.30} & 85.53\std{1.00} & 83.37\std{1.39} \\
    & \textbf{MedGNN}  & 73.71\std{4.64} & 81.38\std{2.00} & 68.40\std{5.99} & 68.40\std{5.99} & 67.54\std{7.95} & 81.39\std{3.83} & 82.69\std{3.50} \\
    & \textbf{Medformer}  & 71.99\std{1.67} & 74.97\std{2.64} & 67.47\std{2.03} & 67.47\std{2.03} & 67.42\std{2.44} & 78.38\std{2.75} & 78.33\std{2.96} \\
    & \textbf{MNet}  & 66.71\std{5.54} & 81.36\std{2.16} & 59.33\std{6.83} & 59.33\std{6.83} & 53.76\std{10.86} & 81.13\std{4.02} & 81.83\std{4.19} \\
    & \textbf{ModernTCN}  & 67.24\std{0.54} & 73.27\std{2.32} & 60.87\std{0.54} & 60.87\std{0.54} & 58.42\std{0.76} & 75.88\std{2.41} & 76.25\std{2.12} \\
    & \textbf{PatchTST}  & 64.54\std{0.37} & 75.24\std{6.16} & 57.34\std{0.79} & 57.34\std{0.79} & 52.39\std{2.65} & 73.76\std{3.11} & 72.85\std{3.27} \\
    & \textbf{TCN}  & \textcolor{mygreen}{78.16\std{3.07}} & \textcolor{mygreen}{82.81\std{2.51}} & \textcolor{mygreen}{74.26\std{4.16}} & \textcolor{mygreen}{74.26\std{4.16}} & \textcolor{mygreen}{74.81\std{4.47}} & \textcolor{mygreen}{85.84\std{3.92}} & \textcolor{mygreen}{86.33\std{3.81}} \\
    & \textbf{TimesNet}  & 66.71\std{3.77} & 74.47\std{6.25} & 59.80\std{4.51} & 59.80\std{4.51} & 55.95\std{6.89} & 56.01\std{4.93} & 59.13\std{5.35} \\

    \cmidrule(lr){2-9}
    & \textbf{BIOT}  & \textcolor{myblue}{81.62\std{3.83}} & \textcolor{myblue}{84.08\std{2.35}} & \textcolor{myblue}{78.92\std{5.17}} & \textcolor{myblue}{78.92\std{5.17}} & \textcolor{myblue}{79.55\std{5.36}} & \textcolor{myred}{\textbf{92.92\std{2.77}}} & \textcolor{myred}{\textbf{92.33\std{2.89}}} \\
    & \textbf{LaBraM}  & 74.24\std{1.74} & 75.28\std{1.25} & 71.52\std{3.22} & 71.52\std{3.22} & 71.65\std{3.35} & 85.61\std{1.20} & 83.91\std{1.44} \\
    & \textbf{CBraMod}  & 75.51\std{3.19} & 76.27\std{4.83} & 73.93\std{1.74} & 73.93\std{1.74} & 74.10\std{2.40} & 83.10\std{1.85} & 82.97\std{2.10} \\
    & \textbf{CSBrain}  & 70.16\std{5.00} & 78.21\std{4.83} & 63.95\std{6.04} & 63.95\std{6.04} & 61.69\std{8.63} & 68.20\std{1.53} & 68.80\std{2.51} \\

    \cmidrule(lr){2-9}
    & \textbf{LEAD (Ours)}  & \textcolor{myred}{\textbf{84.36\std{5.33}}} & \textcolor{myred}{\textbf{84.41\std{5.72}}} & \textcolor{myred}{\textbf{83.36\std{4.91}}} & \textcolor{myred}{\textbf{83.36\std{4.91}}} & \textcolor{myred}{\textbf{83.70\std{5.30}}} & \textcolor{myblue}{92.51\std{4.33}} & \textcolor{myblue}{92.23\std{4.80}} \\

    \midrule
    \multirow{17}{*}{\begin{tabular}[c]{@{}l@{}}\;\makecell{\makecell{\textbf{APAVA} \\ \textit{(Subject-Level)} \\ \textit{(23 Subjects)} \\ \textit{(HC vs AD)}}} \end{tabular}} 
    & \textbf{ManualFeature} & 65.00\std{12.25} & 60.00\std{28.58} & 65.00\std{12.25} & 65.00\std{12.25} & 57.33\std{19.60} & 72.50\std{22.91} & 75.00\std{20.41} \\
    & \textbf{EEGConformer} & 70.00\std{10.00} & 71.67\std{23.33} & 70.00\std{10.00} & 70.00\std{10.00} & 65.33\std{16.00} & 100.00\std{0.00} & 100.00\std{0.00} \\
    & \textbf{EEGInception} & 75.00\std{0.00} & 83.33\std{0.00} & 75.00\std{0.00} & 75.00\std{0.00} & 73.33\std{0.00} & 80.00\std{10.00} & 86.67\std{6.67} \\
    & \textbf{EEGNet} & 75.00\std{15.81} & 75.00\std{25.82} & 75.00\std{15.81} & 75.00\std{15.81} & 70.67\std{21.33} & 77.50\std{12.25} & 81.67\std{13.33} \\
    & \textbf{iTransformer} & \textcolor{mygreen}{85.00\std{12.25}} & \textcolor{mygreen}{90.00\std{8.16}} & \textcolor{mygreen}{85.00\std{12.25}} & \textcolor{mygreen}{85.00\std{12.25}} & \textcolor{mygreen}{84.00\std{13.06}} & \textcolor{mygreen}{100.00\std{0.00}} & \textcolor{mygreen}{100.00\std{0.00}} \\
    & \textbf{MedGNN}  & 70.00\std{10.00} & 71.67\std{23.33} & 70.00\std{10.00} & 70.00\std{10.00} & 65.33\std{16.00} & 70.00\std{18.71} & 76.67\std{16.16} \\
    & \textbf{Medformer}  & 75.00\std{0.00} & 83.33\std{0.00} & 75.00\std{0.00} & 75.00\std{0.00} & 73.33\std{0.00} & 80.00\std{10.00} & 86.67\std{6.67} \\
    & \textbf{MNet}  & 55.00\std{10.00} & 36.67\std{23.33} & 55.00\std{10.00} & 55.00\std{10.00} & 41.33\std{16.00} & 87.50\std{15.81} & 85.00\std{18.56} \\
    & \textbf{ModernTCN}  & 50.00\std{0.00} & 25.00\std{0.00} & 50.00\std{0.00} & 50.00\std{0.00} & 33.33\std{0.00} & 72.50\std{5.00} & 78.33\std{10.00} \\
    & \textbf{PatchTST}  & 50.00\std{0.00} & 25.00\std{0.00} & 50.00\std{0.00} & 50.00\std{0.00} & 33.33\std{0.00} & 65.00\std{5.00} & 63.33\std{10.00} \\
    & \textbf{TCN}  & 75.00\std{0.00} & 83.33\std{0.00} & 75.00\std{0.00} & 75.00\std{0.00} & 73.33\std{0.00} & 95.00\std{10.00} & 96.67\std{6.67} \\
    & \textbf{TimesNet}  & 50.00\std{0.00} & 25.00\std{0.00} & 50.00\std{0.00} & 50.00\std{0.00} & 33.33\std{0.00} & 70.00\std{18.71} & 76.67\std{16.16} \\

    \cmidrule(lr){2-9}
    & \textbf{BIOT}  & \textcolor{myblue}{95.00\std{10.00}} & \textcolor{myblue}{96.67\std{6.67}} & \textcolor{myblue}{95.00\std{10.00}} & \textcolor{myblue}{95.00\std{10.00}} & \textcolor{myblue}{94.67\std{10.67}} & \textcolor{myblue}{100.00\std{0.00}} & \textcolor{myblue}{100.00\std{0.00}} \\
    & \textbf{LaBraM}  & 75.00\std{15.81} & 75.00\std{25.82} & 75.00\std{15.81} & 75.00\std{15.81} & 70.67\std{21.33} & 100.00\std{0.00} & 100.00\std{0.00} \\
    & \textbf{CBraMod}  & 75.00\std{0.00} & 83.33\std{0.00} & 75.00\std{0.00} & 75.00\std{0.00} & 73.33\std{0.00} & 75.00\std{15.81} & 81.67\std{13.33} \\
    & \textbf{CSBrain}  & 65.00\std{12.25} & 60.00\std{28.58} & 65.00\std{12.25} & 65.00\std{12.25} & 57.33\std{19.60} & 50.00\std{0.00} & 58.33\std{0.00} \\

    \cmidrule(lr){2-9}
    & \textbf{LEAD (Ours)}  & \textcolor{myred}{\textbf{100.00\std{0.00}}} & \textcolor{myred}{\textbf{100.00\std{0.00}}} & \textcolor{myred}{\textbf{100.00\std{0.00}}} & \textcolor{myred}{\textbf{100.00\std{0.00}}} & \textcolor{myred}{\textbf{100.00\std{0.00}}} & \textcolor{myred}{\textbf{100.00\std{0.00}}} & \textcolor{myred}{\textbf{100.00\std{0.00}}} \\

\bottomrule
\end{tabular}
}
\end{table*}

\begin{table*}[t]
\centering
\def\arraystretch{1.0}
\caption{\textbf{Method Comparison Detailed Results on CNBPM.} The detailed results comparison between ours and baseline methods on the dataset CNBPM includes all 7 evaluation metrics. The \textcolor{myred}{\textbf{Top-1}}, \textcolor{myblue}{Top-2}, and \textcolor{mygreen}{Top-3} results are highlighted in red, blue, and green. 
}
\label{tab:cnbpm_detail_results}
\resizebox{\textwidth}{!}{%
\begin{tabular}{clccccccc}

    \toprule
    \textbf{Datasets} & \textbf{Models} & \multicolumn{1}{c}{\textbf{Accuracy}} & \multicolumn{1}{c}{\textbf{Precision}} & \multicolumn{1}{c}{\textbf{Sensitivity}} & \multicolumn{1}{c}{\textbf{Specificity}} & \multicolumn{1}{c}{\textbf{F1 Score}} & \multicolumn{1}{c}{\textbf{AUROC}} & \multicolumn{1}{c}{\textbf{AUPRC}} \\
    \midrule
    \multirow{17}{*}{\begin{tabular}[c]{@{}l@{}}\;\makecell{\makecell{\textbf{CNBPM} \\ \textit{(Sample-Level)} \\ \textit{(122,029 Samples)} \\ \textit{(HC vs MCI vs AD)}}} \end{tabular}} 
    & \textbf{ManualFeature} & 48.25\std{8.11} & 44.08\std{6.56} & 43.80\std{6.72} & 73.56\std{3.95} & 43.00\std{6.36} & 61.29\std{6.44} & 41.05\std{5.17} \\
    & \textbf{EEGConformer} & 65.30\std{11.18} & 59.85\std{8.02} & 59.20\std{9.36} & 82.13\std{5.87} & 59.17\std{8.85} & 81.76\std{8.20} & 64.85\std{9.17} \\
    & \textbf{EEGInception} & 63.46\std{10.07} & 59.59\std{8.13} & \textcolor{myblue}{60.66\std{7.84}} & 82.05\std{4.86} & 58.56\std{8.01} & 81.73\std{8.18} & 63.95\std{8.58} \\
    & \textbf{EEGNet} & 47.94\std{7.22} & 44.16\std{5.91} & 41.54\std{6.09} & 72.20\std{3.97} & 38.57\std{6.52} & 63.51\std{7.95} & 45.73\std{5.71} \\
    & \textbf{iTransformer} & 59.25\std{11.23} & 52.26\std{7.65} & 51.91\std{7.87} & 78.98\std{5.65} & 51.37\std{8.02} & 74.99\std{10.16} & 56.17\std{8.73} \\
    & \textbf{MedGNN}  & 65.34\std{9.04} & 59.36\std{7.76} & 59.06\std{7.98} & 81.88\std{4.88} & 58.68\std{7.50} & \textcolor{mygreen}{82.13\std{7.35}} & 65.80\std{7.03} \\
    & \textbf{Medformer}  & \textcolor{myblue}{66.31\std{10.00}} & \textcolor{myblue}{61.15\std{8.05}} & \textcolor{myred}{\textbf{60.75\std{8.15}}} & \textcolor{mygreen}{82.55\std{5.20}} & \textcolor{myred}{\textbf{60.39\std{7.68}}} & \textcolor{myblue}{82.17\std{7.87}} & \textcolor{myblue}{66.25\std{7.41}} \\
    & \textbf{MNet}  & 61.43\std{13.81} & 55.08\std{11.04} & 50.29\std{9.33} & 78.82\std{6.67} & 47.04\std{9.76} & 77.03\std{11.79} & 58.95\std{10.49} \\
    & \textbf{ModernTCN}  & 61.45\std{10.54} & 57.88\std{7.74} & 56.65\std{7.92} & 80.87\std{5.13} & 55.95\std{8.11} & 77.85\std{9.30} & 59.69\std{9.22} \\
    & \textbf{PatchTST}  & 57.43\std{9.39} & 51.20\std{6.10} & 50.79\std{5.75} & 77.75\std{4.22} & 49.78\std{6.10} & 71.52\std{9.58} & 53.69\std{8.24} \\
    & \textbf{TCN}  & \textcolor{mygreen}{66.05\std{7.32}} & 58.90\std{5.09} & 59.23\std{4.66} & \textcolor{myblue}{82.66\std{3.91}} & 58.72\std{5.02} & 82.01\std{6.03} & 63.39\std{5.42} \\
    & \textbf{TimesNet}  & 63.60\std{10.95} & 57.30\std{7.45} & 56.16\std{6.41} & 81.34\std{5.30} & 56.21\std{6.68} & 79.27\std{7.74} & 60.26\std{7.74} \\

    \cmidrule(lr){2-9}
    & \textbf{BIOT}  & 61.70\std{13.03} & 55.70\std{9.53} & 54.96\std{9.21} & 80.37\std{6.36} & 54.79\std{9.52} & 77.95\std{11.46} & 59.39\std{11.05} \\
    & \textbf{LaBraM}  & 60.83\std{12.07} & 53.08\std{7.43} & 53.34\std{6.32} & 79.76\std{5.95} & 52.37\std{7.21} & 78.27\std{9.83} & 58.82\std{8.45} \\
    & \textbf{CBraMod}  & 61.84\std{10.71} & 54.79\std{7.13} & 54.14\std{6.75} & 79.78\std{5.67} & 53.93\std{6.87} & 78.68\std{9.10} & 60.68\std{7.13} \\
    & \textbf{CSBrain}  & \textcolor{myred}{\textbf{67.11\std{7.61}}} & \textcolor{mygreen}{60.96\std{5.78}} & \textcolor{mygreen}{60.59\std{4.80}} & \textcolor{myred}{\textbf{82.69\std{3.85}}} & \textcolor{myblue}{60.30\std{5.41}} & \textcolor{myred}{\textbf{83.38\std{5.73}}} & \textcolor{myred}{\textbf{66.55\std{5.21}}} \\

    \cmidrule(lr){2-9}
    & \textbf{LEAD (Ours)}  & 65.49\std{11.26} & \textcolor{myred}{\textbf{62.36\std{7.14}}} & 59.41\std{6.36} & 81.09\std{5.76} & \textcolor{mygreen}{59.70\std{6.64}} & 79.34\std{10.07} & \textcolor{mygreen}{65.98\std{9.53}} \\

    \midrule
    \multirow{17}{*}{\begin{tabular}[c]{@{}l@{}}\;\makecell{\makecell{\textbf{CNBPM} \\ \textit{(Subject-Level)} \\ \textit{(189 Subjects)} \\ \textit{(HC vs MCI vs AD)}}} \end{tabular}} 
    & \textbf{ManualFeature} & 47.62\std{6.02} & 41.40\std{10.97} & 47.62\std{6.02} & 73.81\std{3.01} & 42.90\std{8.74} & 69.86\std{8.21} & 57.50\std{9.77} \\
    & \textbf{EEGConformer} & 60.00\std{7.74} & \textcolor{myblue}{68.74\std{8.17}} & 60.00\std{7.74} & 80.00\std{3.87} & 58.42\std{8.60} & 81.63\std{4.71} & 73.79\std{5.32} \\
    & \textbf{EEGInception} & 61.90\std{6.73} & 61.98\std{8.63} & 61.90\std{6.73} & 80.95\std{3.37} & 60.41\std{7.37} & \textcolor{myblue}{82.93\std{5.36}} & \textcolor{myblue}{75.18\std{6.41}} \\
    & \textbf{EEGNet} & 40.00\std{9.81} & 35.17\std{17.28} & 40.00\std{9.81} & 70.00\std{4.90} & 34.02\std{11.43} & 64.46\std{5.93} & 53.19\std{3.85} \\
    & \textbf{iTransformer} & 56.19\std{7.00} & 57.95\std{9.63} & 56.19\std{7.00} & 78.10\std{3.50} & 52.85\std{8.58} & 78.30\std{7.84} & 69.53\std{10.18} \\
    & \textbf{MedGNN}  & 54.29\std{10.26} & 56.81\std{15.83} & 54.29\std{10.26} & 77.14\std{5.13} & 52.18\std{11.32} & 76.60\std{5.61} & 67.65\std{5.94} \\
    & \textbf{Medformer}  & 62.86\std{5.55} & 64.22\std{7.00} & 62.86\std{5.55} & 81.43\std{2.78} & 60.93\std{6.49} & 80.58\std{3.94} & 74.24\std{5.16} \\
    & \textbf{MNet}  & 49.52\std{5.71} & 34.50\std{3.91} & 49.52\std{5.71} & 74.76\std{2.86} & 39.97\std{4.38} & 74.12\std{8.65} & 64.48\std{10.85} \\
    & \textbf{ModernTCN}  & 63.81\std{7.13} & \textcolor{mygreen}{66.42\std{10.20}} & 63.81\std{7.13} & 81.90\std{3.56} & 61.96\std{9.90} & 82.28\std{5.41} & \textcolor{mygreen}{74.81\std{7.33}} \\
    & \textbf{PatchTST}  & 56.19\std{5.55} & 56.83\std{16.05} & 56.19\std{5.55} & 78.10\std{2.78} & 51.35\std{9.07} & 76.80\std{3.11} & 67.53\std{6.35} \\
    & \textbf{TCN}  & \textcolor{myblue}{63.81\std{3.81}} & 63.66\std{3.19} & \textcolor{myblue}{63.81\std{3.81}} & \textcolor{myblue}{81.90\std{1.90}} & \textcolor{myblue}{62.66\std{3.41}} & \textcolor{mygreen}{82.69\std{3.36}} & 74.34\std{4.02} \\
    & \textbf{TimesNet}  & 57.14\std{5.22} & 59.11\std{9.17} & 57.14\std{5.22} & 78.57\std{2.61} & 55.52\std{4.93} & 77.82\std{4.56} & 68.87\std{6.72} \\

    \cmidrule(lr){2-9}
    & \textbf{BIOT}  & 57.14\std{7.97} & 58.87\std{8.30} & 57.14\std{7.97} & 78.57\std{3.98} & 55.74\std{8.58} & 78.10\std{8.27} & 68.06\std{10.28} \\
    & \textbf{LaBraM}  & 53.33\std{4.67} & 49.93\std{7.94} & 53.33\std{4.67} & 76.67\std{2.33} & 49.83\std{6.46} & 78.06\std{4.39} & 68.33\std{6.68} \\
    & \textbf{CBraMod}  & 54.29\std{9.33} & 52.97\std{13.37} & 54.29\std{9.33} & 77.14\std{4.67} & 50.60\std{10.89} & 75.14\std{4.75} & 66.13\std{5.18} \\
    & \textbf{CSBrain}  & \textcolor{mygreen}{63.81\std{4.86}} & 64.93\std{7.29} & \textcolor{mygreen}{63.81\std{4.86}} & \textcolor{mygreen}{81.90\std{2.43}} & \textcolor{mygreen}{62.41\std{5.80}} & \textcolor{myred}{\textbf{83.50\std{5.90}}} & \textcolor{myred}{\textbf{75.71\std{7.84}}} \\

    \cmidrule(lr){2-9}
    & \textbf{LEAD (Ours)}  & \textcolor{myred}{\textbf{66.67\std{6.73}}} & \textcolor{myred}{\textbf{69.85\std{8.25}}} & \textcolor{myred}{\textbf{66.67\std{6.73}}} & \textcolor{myred}{\textbf{83.33\std{3.37}}} & \textcolor{myred}{\textbf{65.67\std{6.59}}} & 80.37\std{7.21} & 74.19\std{9.73} \\

\bottomrule
\end{tabular}
}
\end{table*}

The detailed results for baseline method comparisons on five downstream datasets are presented in Table~\ref{tab:adfsu_detail_results} (ADFSU), Table~\ref{tab:adftd_detail_results} (ADFTD), Table~\ref{tab:adsz_detail_results} (ADSZ)
, Table~\ref{tab:apava_detail_results} (APAVA), and Table~\ref{tab:cnbpm_detail_results} (CNBPM), respectively.

\subsection{Per-Subject Analysis}
\label{sub:per_subject_analysis}

\begin{table*}
    \centering
    \scriptsize
    \caption{\textbf{First Part of per-subject results of ADFTD.} Demographic information, MMSE, recording length, and per-subject results using the LOSO setting of all subjects from Sub-001 to Sub-065.}
    \vspace{-2mm}
    \label{tab:per_subject_analysis_first}
    \resizebox{0.72\textwidth}{!}{
    \begin{tabular}{l|ccccccc}
    \toprule
    \textbf{Subject ID} & \textbf{Gender} & \textbf{Age} & \textbf{Class} & \textbf{MMSE} & \textbf{Length} & \textbf{\makecell{Sample-level \\ Accuracy}} & \textbf{\makecell{Subject-level \\ Accuracy}} \\
    \midrule
    \textbf{Sub-001} & F & 57 & AD & 16 & 0.25h & 77.13\% & \textcolor{ForestGreen}{\ding{51}} \\
    \textbf{Sub-002} & F & 78 & AD & 22 & 0.33h & 40.35\% & \textcolor{BrickRed}{\ding{55}} \\
    \textbf{Sub-003} & M & 70 & AD & 14 & 0.16h & 82.44\% & \textcolor{ForestGreen}{\ding{51}} \\
    \textbf{Sub-004} & F & 67 & AD & 20 & 0.29h & 95.04\% & \textcolor{ForestGreen}{\ding{51}} \\
    \textbf{Sub-005} & M & 70 & AD & 22 & 0.30h & 96.08\% & \textcolor{ForestGreen}{\ding{51}} \\
    \textbf{Sub-006} & F & 61 & AD & 14 & 0.28h & 58.93\% & \textcolor{ForestGreen}{\ding{51}} \\
    \textbf{Sub-007} & F & 79 & AD & 20 & 0.33h & 92.98\% & \textcolor{ForestGreen}{\ding{51}} \\
    \textbf{Sub-008} & M & 62 & AD & 16 & 0.33h & 79.14\% & \textcolor{ForestGreen}{\ding{51}} \\
    \textbf{Sub-009} & F & 77 & AD & 23 & 0.28h & 73.26\% & \textcolor{ForestGreen}{\ding{51}} \\
    \textbf{Sub-010} & M & 69 & AD & 20 & 0.44h & 54.76\% & \textcolor{ForestGreen}{\ding{51}} \\
    \textbf{Sub-011} & M & 71 & AD & 22 & 0.29h & 89.66\% & \textcolor{ForestGreen}{\ding{51}} \\
    \textbf{Sub-012} & M & 63 & AD & 18 & 0.34h & 74.04\% & \textcolor{ForestGreen}{\ding{51}} \\
    \textbf{Sub-013} & F & 64 & AD & 20 & 0.36h & 92.89\% & \textcolor{ForestGreen}{\ding{51}} \\
    \textbf{Sub-014} & M & 77 & AD & 14 & 0.40h & 85.86\% & \textcolor{ForestGreen}{\ding{51}} \\
    \textbf{Sub-015} & M & 61 & AD & 18 & 0.33h & 87.08\% & \textcolor{ForestGreen}{\ding{51}} \\
    \textbf{Sub-016} & F & 68 & AD & 14 & 0.39h & 94.99\% & \textcolor{ForestGreen}{\ding{51}} \\
    \textbf{Sub-017} & F & 61 & AD & 6 & 0.35h & 98.31\% & \textcolor{ForestGreen}{\ding{51}} \\
    \textbf{Sub-018} & F & 73 & AD & 23 & 0.34h & 98.53\% & \textcolor{ForestGreen}{\ding{51}} \\
    \textbf{Sub-019} & F & 62 & AD & 14 & 0.34h & 59.57\% & \textcolor{ForestGreen}{\ding{51}} \\
    \textbf{Sub-020} & M & 71 & AD & 4 & 0.35h & 94.01\% & \textcolor{ForestGreen}{\ding{51}} \\
    \textbf{Sub-021} & M & 79 & AD & 22 & 0.30h & 79.66\% & \textcolor{ForestGreen}{\ding{51}} \\
    \textbf{Sub-022} & F & 68 & AD & 20 & 0.30h & 94.78\% & \textcolor{ForestGreen}{\ding{51}} \\
    \textbf{Sub-023} & M & 60 & AD & 16 & 0.25h & 69.73\% & \textcolor{ForestGreen}{\ding{51}} \\
    \textbf{Sub-024} & F & 69 & AD & 20 & 0.23h & 98.96\% & \textcolor{ForestGreen}{\ding{51}} \\
    \textbf{Sub-025} & F & 79 & AD & 20 & 0.26h & 31.06\% & \textcolor{BrickRed}{\ding{55}} \\
    \textbf{Sub-026} & F & 61 & AD & 18 & 0.27h & 97.96\% & \textcolor{ForestGreen}{\ding{51}} \\
    \textbf{Sub-027} & F & 67 & AD & 16 & 0.26h & 99.50\% & \textcolor{ForestGreen}{\ding{51}} \\
    \textbf{Sub-028} & M & 49 & AD & 20 & 0.33h & 91.12\% & \textcolor{ForestGreen}{\ding{51}} \\
    \textbf{Sub-029} & F & 53 & AD & 16 & 0.23h & 96.75\% & \textcolor{ForestGreen}{\ding{51}} \\
    \textbf{Sub-030} & F & 56 & AD & 20 & 0.18h & 99.47\% & \textcolor{ForestGreen}{\ding{51}} \\
    \textbf{Sub-031} & F & 67 & AD & 22 & 0.34h & 75.28\% & \textcolor{ForestGreen}{\ding{51}} \\
    \textbf{Sub-032} & F & 59 & AD & 20 & 0.24h & 85.07\% & \textcolor{ForestGreen}{\ding{51}} \\
    \textbf{Sub-033} & F & 72 & AD & 20 & 0.22h & 70.55\% & \textcolor{ForestGreen}{\ding{51}} \\
    \textbf{Sub-034} & F & 75 & AD & 18 & 0.29h & 93.91\% & \textcolor{ForestGreen}{\ding{51}} \\
    \textbf{Sub-035} & F & 57 & AD & 22 & 0.32h & 91.30\% & \textcolor{ForestGreen}{\ding{51}} \\
    \textbf{Sub-036} & F & 58 & AD & 9 & 0.26h & 62.05\% & \textcolor{ForestGreen}{\ding{51}} \\
    \textbf{Sub-037} & M & 57 & HC & 30 & 0.37h & 75.66\% & \textcolor{ForestGreen}{\ding{51}} \\
    \textbf{Sub-038} & M & 62 & HC & 30 & 0.35h & 93.12\% & \textcolor{ForestGreen}{\ding{51}} \\
    \textbf{Sub-039} & M & 70 & HC & 30 & 0.34h & 88.15\% & \textcolor{ForestGreen}{\ding{51}} \\
    \textbf{Sub-040} & M & 61 & HC & 30 & 0.32h & 80.61\% & \textcolor{ForestGreen}{\ding{51}} \\
    \textbf{Sub-041} & F & 77 & HC & 30 & 0.34h & 81.21\% & \textcolor{ForestGreen}{\ding{51}} \\
    \textbf{Sub-042} & M & 74 & HC & 30 & 0.33h & 91.37\% & \textcolor{ForestGreen}{\ding{51}} \\
    \textbf{Sub-043} & M & 72 & HC & 30 & 0.31h & 92.28\% & \textcolor{ForestGreen}{\ding{51}} \\
    \textbf{Sub-044} & F & 64 & HC & 30 & 0.38h & 87.45\% & \textcolor{ForestGreen}{\ding{51}} \\
    \textbf{Sub-045} & F & 70 & HC & 30 & 0.34h & 77.45\% & \textcolor{ForestGreen}{\ding{51}} \\
    \textbf{Sub-046} & M & 63 & HC & 30 & 0.36h & 85.24\% & \textcolor{ForestGreen}{\ding{51}} \\
    \textbf{Sub-047} & F & 70 & HC & 30 & 0.34h & 90.48\% & \textcolor{ForestGreen}{\ding{51}} \\
    \textbf{Sub-048} & M & 65 & HC & 30 & 0.38h & 88.31\% & \textcolor{ForestGreen}{\ding{51}} \\
    \textbf{Sub-049} & F & 62 & HC & 30 & 0.34h & 76.68\% & \textcolor{ForestGreen}{\ding{51}} \\
    \textbf{Sub-050} & M & 68 & HC & 30 & 0.31h & 96.14\% & \textcolor{ForestGreen}{\ding{51}} \\
    \textbf{Sub-051} & F & 75 & HC & 30 & 0.33h & 94.02\% & \textcolor{ForestGreen}{\ding{51}} \\
    \textbf{Sub-052} & F & 73 & HC & 30 & 0.33h & 74.64\% & \textcolor{ForestGreen}{\ding{51}} \\
    \textbf{Sub-053} & M & 70 & HC & 30 & 0.31h & 94.20\% & \textcolor{ForestGreen}{\ding{51}} \\
    \textbf{Sub-054} & M & 78 & HC & 30 & 0.35h & 92.48\% & \textcolor{ForestGreen}{\ding{51}} \\
    \textbf{Sub-055} & M & 67 & HC & 30 & 0.33h & 82.26\% & \textcolor{ForestGreen}{\ding{51}} \\
    \textbf{Sub-056} & F & 64 & HC & 30 & 0.34h & 81.32\% & \textcolor{ForestGreen}{\ding{51}} \\
    \textbf{Sub-057} & M & 64 & HC & 30 & 0.32h & 92.14\% & \textcolor{ForestGreen}{\ding{51}} \\
    \textbf{Sub-058} & M & 62 & HC & 30 & 0.27h & 77.62\% & \textcolor{ForestGreen}{\ding{51}} \\
    \textbf{Sub-059} & M & 77 & HC & 30 & 0.30h & 67.08\% & \textcolor{ForestGreen}{\ding{51}} \\
    \textbf{Sub-060} & F & 71 & HC & 30 & 0.33h & 95.32\% & \textcolor{ForestGreen}{\ding{51}} \\
    \textbf{Sub-061} & F & 63 & HC & 30 & 0.33h & 77.05\% & \textcolor{ForestGreen}{\ding{51}} \\
    \textbf{Sub-062} & M & 67 & HC & 30 & 0.35h & 51.18\% & \textcolor{ForestGreen}{\ding{51}} \\
    \textbf{Sub-063} & M & 66 & HC & 30 & 0.34h & 84.08\% & \textcolor{ForestGreen}{\ding{51}} \\
    \textbf{Sub-064} & M & 66 & HC & 30 & 0.35h & 97.25\% & \textcolor{ForestGreen}{\ding{51}} \\
    \textbf{Sub-065} & F & 71 & HC & 30 & 0.30h & 93.43\% & \textcolor{ForestGreen}{\ding{51}} \\
    \bottomrule
    \end{tabular}
    }
\vspace{-2mm}
\end{table*}

\begin{table*}
    \centering
    \scriptsize
    \caption{\textbf{Second Part of per-subject results of ADFTD.} Demographic information, MMSE, recording length, and results using the LOSO setting of all subjects from Sub-066 to Sub-088. The average in the last line applies to all subjects in the two tables, from Sub-001 to Sub-088.}
    \vspace{-2mm}
    \label{tab:per_subject_analysis_second}
    \resizebox{0.72\textwidth}{!}{
    \begin{tabular}{l|ccccccc}
    \toprule
    \textbf{Subject ID} & \textbf{Gender} & \textbf{Age} & \textbf{Class} & \textbf{MMSE} & \textbf{Length} & \textbf{\makecell{Sample-level \\ Accuracy}} & \textbf{\makecell{Subject-level \\ Accuracy}} \\
    \midrule
    \textbf{Sub-066} & M & 73 & FTD & 20 & 0.22h & 67.17\% & \textcolor{ForestGreen}{\ding{51}} \\
    \textbf{Sub-067} & M & 66 & FTD & 24 & 0.28h & 97.35\% & \textcolor{ForestGreen}{\ding{51}} \\
    \textbf{Sub-068} & M & 78 & FTD & 25 & 0.22h & 26.06\% & \textcolor{BrickRed}{\ding{55}} \\
    \textbf{Sub-069} & M & 70 & FTD & 22 & 0.31h & 97.36\% & \textcolor{ForestGreen}{\ding{51}} \\
    \textbf{Sub-070} & F & 67 & FTD & 22 & 0.25h & 68.77\% & \textcolor{ForestGreen}{\ding{51}} \\
    \textbf{Sub-071} & M & 62 & FTD & 20 & 0.31h & 46.27\% & \textcolor{ForestGreen}{\ding{51}} \\
    \textbf{Sub-072} & M & 65 & FTD & 18 & 0.28h & 50.94\% & \textcolor{ForestGreen}{\ding{51}} \\
    \textbf{Sub-073} & F & 57 & FTD & 22 & 0.39h & 87.97\% & \textcolor{ForestGreen}{\ding{51}} \\
    \textbf{Sub-074} & F & 53 & FTD & 20 & 0.37h & 94.46\% & \textcolor{ForestGreen}{\ding{51}} \\
    \textbf{Sub-075} & F & 71 & FTD & 22 & 0.31h & 73.04\% & \textcolor{ForestGreen}{\ding{51}} \\
    \textbf{Sub-076} & M & 44 & FTD & 24 & 0.35h & 77.96\% & \textcolor{ForestGreen}{\ding{51}} \\
    \textbf{Sub-077} & M & 61 & FTD & 22 & 0.27h & 94.68\% & \textcolor{ForestGreen}{\ding{51}} \\
    \textbf{Sub-078} & M & 62 & FTD & 22 & 0.34h & 90.94\% & \textcolor{ForestGreen}{\ding{51}} \\
    \textbf{Sub-079} & F & 60 & FTD & 18 & 0.25h & 98.09\% & \textcolor{ForestGreen}{\ding{51}} \\
    \textbf{Sub-080} & F & 71 & FTD & 20 & 0.28h & 94.20\% & \textcolor{ForestGreen}{\ding{51}} \\
    \textbf{Sub-081} & F & 61 & FTD & 18 & 0.25h & 64.80\% & \textcolor{ForestGreen}{\ding{51}} \\
    \textbf{Sub-082} & M & 63 & FTD & 27 & 0.24h & 65.38\% & \textcolor{ForestGreen}{\ding{51}} \\
    \textbf{Sub-083} & F & 68 & FTD & 20 & 0.27h & 58.15\% & \textcolor{BrickRed}{\ding{55}} \\
    \textbf{Sub-084} & F & 71 & FTD & 24 & 0.20h & 15.91\% & \textcolor{BrickRed}{\ding{55}} \\
    \textbf{Sub-085} & M & 64 & FTD & 26 & 0.18h & 50.53\% & \textcolor{ForestGreen}{\ding{51}} \\
    \textbf{Sub-086} & M & 49 & FTD & 26 & 0.18h & 20.11\% & \textcolor{BrickRed}{\ding{55}} \\
    \textbf{Sub-087} & M & 73 & FTD & 24 & 0.19h & 88.01\% & \textcolor{ForestGreen}{\ding{51}} \\
    \textbf{Sub-088} & M & 55 & FTD & 24 & 0.25h & 68.81\% & \textcolor{ForestGreen}{\ding{51}} \\

    \midrule
    \textbf{Average} & -- & 66.17 & -- & 22.94 & 0.30h & 79.74\% & 93.18\% \\
    
    \bottomrule
    \end{tabular}
    }
\vspace{-2mm}
\end{table*}

The detailed results for the per-subject analysis under the leave-one-subject-out (LOSO) setting on the ADFTD dataset are presented in Table~\ref{tab:per_subject_analysis_first} and Table~\ref{tab:per_subject_analysis_second}.

\end{appendices}


\clearpage

\end{document}